%% file: ms.tex
\documentclass[journal]{IEEEtran}

\ifCLASSINFOpdf
  \usepackage[pdftex]{graphicx}
  \graphicspath{{../pdf/}{../jpeg/}}
  \DeclareGraphicsExtensions{.pdf,.jpeg,.png}
\else
\fi

\usepackage{amsmath}

\usepackage{amssymb}  %
\usepackage{amsthm}
\usepackage{leftidx}
\usepackage{multirow}
\usepackage{ulem} %
\usepackage{algpseudocode}
\usepackage{algorithm}
\usepackage{booktabs}

\makeatletter
\let\NAT@parse\undefined
\makeatother
\usepackage[pagebackref=true,bookmarks=false,hidelinks]{hyperref}

\input{custom_commands/latin_abbreviations.tex}

\DeclareMathAlphabet{\mbf}{OT1}{ptm}{b}{n}
\newcommand{\mbs}[1]{{\boldsymbol{#1}}}

\newtheorem{theorem}{Theorem}
\newtheorem{corollary}{Corollary}[theorem]
\newtheorem{lemma}{Lemma}

\newcommand*\numcircledmod[1]{\raisebox{.5pt}{\textcircled{\raisebox{-.9pt} {#1}}}}

\usepackage{xcolor}

\newcommand*{\tran}{^{\mkern-1.5mu\mathsf{T}}}
\newcommand{\ra}[1]{\renewcommand{\arraystretch}{#1}}

\begin{document}
\title{Observability Analysis and Keyframe-Based Filtering for
	Visual Inertial Odometry with Full Self-Calibration}

\author{Jianzhu~Huai$^{1}$,~\IEEEmembership{Member,~IEEE},
        Yukai~Lin$^{2}$,
        Yuan~Zhuang$^{1\dagger}$,~\IEEEmembership{Member,~IEEE}\\
        Charles~Toth$^{3}$,~\IEEEmembership{Senior Life Member,~IEEE},
        and~Dong~Chen$^{1}$,~\IEEEmembership{Member,~IEEE}
\thanks{$^{1}$J. Huai, Y. Zhuang, D. Chen are with the State Key Laboratory of Information Engineering in Surveying, Mapping, and Remote Sensing (LIESMARS), Wuhan University,
129 Luoyu Road, Wuhan, Hubei, 430079, China. Y. Zhuang is also with Wuhan Institute of Quantum Technology, Wuhan, Hubei, 430206, China.
Homepage of J. Huai: https://www.jianzhuhuai.com/.}%
\thanks{$^{2}$Y. Lin is with Department of Computer Science, ETH Zürich, Switzerland.}%
\thanks{$^{3}$Charles Toth is with the Department
	of Civil, Environmental, and Geodetic Engineering, The Ohio State University, Columbus,
	OH 43210, USA.}%
\thanks{$^{\dagger}$Corresponding author, yuan.zhuang@whu.edu.cn.}%
\thanks{Jianzhu Huai is partly funded by the National Natural Science Foundation of China under Grant 62003248 and the LIESMARS Special Research Funding.
This work was partly supported by Excellent Youth Foundation of Hubei Scientific Committee
(2021CFA040) and Guangdong Basic and Applied Basic Research Foundation (2021A1515110343).}
}

\maketitle

\begin{abstract}
Camera-IMU (Inertial Measurement Unit) sensor fusion has been extensively studied in recent decades.
Numerous observability analysis and fusion schemes for motion estimation with self-calibration have been presented.
However, it has been uncertain whether both camera and IMU intrinsic parameters are observable under general motion.
To answer this question, by using the Lie derivatives, we first prove that for a rolling shutter (RS) camera-IMU system, all intrinsic and extrinsic parameters, camera time offset, and readout time of the RS camera, are observable with an unknown landmark.
To our knowledge, we are the first to present such a proof.
Next, to validate this analysis and to solve the drift issue of a structureless filter during standstills, 
we develop a Keyframe-based Sliding Window Filter (KSWF) for odometry and self-calibration, which works with a monocular RS camera or stereo RS cameras.
Though the keyframe concept is widely used in vision-based sensor fusion, to our knowledge, KSWF is the first of its kind to support self-calibration.
Our simulation and real data tests have validated that it is possible to fully calibrate the camera-IMU system using observations of opportunistic landmarks under diverse motion.
Real data tests confirmed previous allusions that keeping landmarks in the state vector can remedy the drift in standstill, 
and showed that the keyframe-based scheme is an alternative solution.
\end{abstract}


\IEEEpeerreviewmaketitle

\section{Introduction}
\IEEEPARstart{V}{isual} Inertial Odometry (VIO) estimates the motion of an agent using data captured by the rigidly mounted cameras and Inertial Measurement Units (IMUs).
With great potential for augmented reality and robotics,
many VIO methods have been developed and deployed on smartphones \cite{flintVisualbased2016}, 
unmanned aerial vehicles \cite{mcclurePerimeterStructureUnmanned2017},
floor-cleaning robots \cite{packSimultaneousLocalizationMapping2015},
and delivery robots \cite{huaiSegwayDRIVEBenchmark2019}. 
However, these methods still have many weaknesses related to dynamic scenes \cite{bescosDynaSLAMIITightlyCoupled2021},
unstable initialization \cite{zuniga-noelAnalytical2021},
coarse sensor calibration \cite{schneiderObservabilityaware2019}, 
and degenerate motion \cite{guoEfficientVisualinertialNavigation2014}, to name a few.
In an effort to build low-cost and robust VIO systems, this paper looks into the last two problems.

For most VIO methods, proper camera-IMU system calibration is ineluctable in 
achieving accurate motion estimation \cite{schneiderObservabilityaware2019} and stable initialization \cite{zuniga-noelAnalytical2021}.
But millions of consumer products, \eg, smartphones, often have low-cost sensors that are inaccurately calibrated.
For such sensor systems, real-time self-calibration along with motion estimation has become a viable option, motivated by the ensuing benefits.
First, additional computation needed for self-calibration is marginal and 
usually improves motion estimation \cite{liHighfidelity2014, genevaOpenVINSResearchPlatform2019}.
Second, this capability can be disabled online to save power if needed.
Third, self-calibration also comes in handy when dealing with legacy data captured without sensor calibration.
Thus, current VIO methods usually have some ability to calibrate the sensor system, 
especially the IMU biases and camera extrinsic parameters, \eg, \cite{leuteneggerKeyframebased2015,bloeschIteratedExtendedKalman2017}.
One question underlying self-calibration is whether the calibration parameters are observable.
Several works have analyzed observability of camera extrinsics \cite{kellyVisualinertialSensorFusion2011}, camera intrinsics \cite{tsaoObservabilityAnalysisPerformance2019}, 
and IMU intrinsics \cite{jungObservability2020}.
There is empirical evidence \cite{liHighfidelity2014,huaiCollaborativeSLAMCrowdsourced2017} that 
the camera-related parameters and IMU intrinsic parameters could be jointly estimated in VIO,
but there lacks a formal proof for this simultaneous self-calibration.

Ideally, a VIO method should exhibit little drift when the system goes though degenerate motion, \eg, standstill.
Although filtering-based VIO systems have been taking the lead in research and commercial products \cite{flintVisualbased2016,roumeliotisVisionaided2017} 
because of its efficiency among the existing geometrical approaches,
quite a few of them suffer from drift during degenerate motion \cite{kottasDetectingDealingHovering2013}.
One influential type of filters stems from the multi-state constraint Kalman filter (MSCKF) \cite{mourikisMultistateConstraintKalman2007}.
In contrast to the EKF style filters which keep the latest motion variables and a set of landmarks in the state, \eg,
\cite{heschCameraIMUbasedLocalizationObservability2014,bloeschIteratedExtendedKalman2017},
MSCKFs \cite{liHighprecisionConsistentEKFbased2013, genevaOpenVINSResearchPlatform2019} keep in the state vector a sliding window 
of motion variables associated with images
but no landmarks (hence structureless or memoryless).
Despite the efficiency brought about by the disposal of landmarks, existing structureless filters for monocular VIO are prone to dramatic drift under
degenerate motion, \eg hovering or standstill \cite{kottasDetectingDealingHovering2013}.
Several attempts have been made to address this issue.
A state management strategy based on the last-in-first-out rule was presented in \cite{guoEfficientVisualinertialNavigation2014}, 
but it requires detecting hovering.
OpenVINS \cite{genevaOpenVINSResearchPlatform2019} deals with standstills by using zero velocity update (ZUPT) which again requires motion detection heuristics.

This paper tries to address the above two challenges to VIO, inaccurate sensor calibration and degenerate motion.
First, we advocate calibrating the visual inertial system by considering all the relevant sensor modeling parameters,
and then support its feasibility by a thorough observability analysis.
Second, we propose a Keyframe-based Sliding Window Filter (KSWF), leveraging keyframes to handle degenerate motion.

Specifically, first, we develop the observability analysis of a VIO system that self-calibrates
a variety of camera and IMU parameters in an unknown environment,
with the Observability Rank Condition (ORC)
\cite{hermannNonlinearControllabilityObservability1977, martinelliVisualinertialStructureMotion2013}
of Lie derivatives.
Under general motion, we prove that camera extrinsic and intrinsic parameters, IMU biases, scale and misalignment, camera time offset, and rolling shutter (RS) effect, are weakly observable.
To our knowledge, observability analysis for the full calibration has not been conducted before.

Then, the KSWF with self-calibration is presented for setups consisting of an IMU and one or two cameras.
Its frontend takes as input from the camera rig a sequence of frames or roughly synced frame pairs (called frame bundles for generality),
associates keypoints between frames,
and selects keyframe bundles based on view overlaps.
The backend filter keeps in the state vector opportunistic landmarks and a sliding window of motion state variables
which will be opportunely marginalized depending on
whether they are associated with keyframe bundles.
As keyframes typically do not change when the platform goes through degenerate motion, 
the keyframe-based scheme is resilient to standstill and hovering.
Granted that KSWF has similarities with recent VIO methods, yet it is unique in several ways.
For example, compared to the original MSCKF \cite{mourikisMultistateConstraintKalman2007},
KSWF is equipped with support for RS effect, compatibility for stereo camera setups,
full self-calibration, keyframe-based feature matching, and keyframe-based state management.

Lastly, in simulation and real world tests with KSWF, we validate that 
the camera and IMU calibration parameters are observable and 
self-calibration improves motion estimation under general motion,
and that the drift of existing monocular structureless filters in degenerate motion can be mitigated by the keyframe-based scheme.

The next section reviews recent developments on VIO from the aspects of observability analysis and 
real-time motion estimation with degenerate motion and distinguishes our study from related ones.
Section \ref{sec:method} describes the proposed KSWF including
measurement models, state vector, keyframe-based frontend and backend.
Its observability analysis is presented in Section \ref{sec:observability}.
Simulation study on sensor parameter observability is reported in Section \ref{sec:simulation}.
Section \ref{sec:real-world-tests} evaluates motion estimation and self-calibration on public benchmarks.
Section \ref{sec:conclusion} summarizes the work and points out future directions.

\section{Related Work}
\label{sec:related}
Regarding observability analysis that underlies on-the-fly sensor calibration and real-time VIO methods that deals with degenerate motion,
we briefly review investigations about camera-IMU-based sensor fusion.

\subsection{Observability Analysis and Self-Calibration}
As the camera-IMU systems are widely adopted, the necessity of calibration motivates the observability property analysis which has 
been explored from various aspects.
This analysis is usually done with the ORC of an observability matrix \cite{maesObservability2019} constructed from either Jacobians of the linearized system (\eg, \cite{ heschConsistencyAnalysisImprovement2014,yangDegenerateMotionAnalysis2019}) or 
gradients of Lie derivatives of the original nonlinear system (\eg, \cite{mirzaeiKalman2008,
kellyVisualinertialSensorFusion2011, heschCameraIMUbasedLocalizationObservability2014, jungObservability2020, martinelliObservability2020}).
Both types of ORC have revealed that the camera-IMU-based odometry in general has four unobservable
directions, \ie, the absolute translation and rotation about gravity \cite{heschConsistencyAnalysisImprovement2014, martinelliVisionIMUData2012}.
Free of approximation, the latter ORC approach lends itself to automated symbolic deduction, and thus, is capable of dealing with more complex systems.

Much effort has been devoted to the observability of sensor parameters which is the theoretical foundation for self-calibration.
In \cite{mirzaeiKalman2008}, the camera extrinsic parameters are shown to be observable when a calibration target is available.
With natural features, the extrinsic parameters prove to be observable when at least two axes of the accelerometer 
and two axes of the gyroscope are excited (nonzero) \cite{kellyVisualinertialSensorFusion2011}.
Yang \etal \cite{yangDegenerateMotionAnalysis2019} showed that the spatiotemporal parameters are observable with general motion by analyzing the linearized camera-IMU-based odometry system.
They further presented that the IMU intrinsic parameters are observable when all axes of the IMU are excited \cite{yangOnline2020}.
The same conclusion was drawn in \cite{jungObservability2020} for a stereo camera-IMU system by using Lie derivatives.
The camera intrinsic parameters (excluding lens distortion) are shown to be observable in \cite{tsaoObservabilityAnalysisPerformance2019} under general motion.
Though studies have shown that online calibration of the camera intrinsic, extrinsic, and IMU intrinsic parameters in combination 
is possible and beneficial to motion estimation \cite{liHighfidelity2014,huaiCollaborativeSLAMCrowdsourced2017},
a formal analysis of whether these parameters are observable is unavailable.

For the camera time offset to the IMU, it is asserted in \cite{yangDegenerateMotionAnalysis2019} that it is 
only unobservable when the IMU experiences constant angular rate and linear acceleration.
This assertion can also be interpreted from deductions in \cite{liOnlineTemporalCalibration2014}. 
To our knowledge, the observability of camera time offset in a VIO system has only been analyzed with
a linearized system in \cite{yangDegenerateMotionAnalysis2019} where the motion state variables are at varying times that depend on the time offset.
Despite being conceptually simpler, the approach based on Lie derivatives has not been used to analyze whether time offset and readout time of a RS camera in the VIO system is observable (see Appendix~\ref{app:time-offset} and \ref{app:readout-time}).

Regarding the above gaps, we conduct observability analysis of the VIO system with full self-calibration by using the Lie derivatives,
revealing the observables in monocular camera-IMU and stereo camera-IMU setups.

\subsection{Motion Estimation in Degenerate Motion}

Two types of geometrical approaches for VIO are prevalent, those based on optimization and filters.
As some measurements arrive, an optimization-based approach conducts repeated linearizations and updates to refine the motion and structure variables.
Example optimization-based VIO systems are OKVIS \cite{leuteneggerKeyframebased2015},
VINS-Mono \cite{qinVINSMono2018}, and \cite{usenkoVisualinertialMappingNonlinear2020}.
By contrast, a filter linearizes these measurements and updates the state variables and their covariance,
thereby saving much computation while supplying the covariance at no additional cost. 
Thus, filters are especially suitable for low-cost systems \cite{flintVisualbased2016}.

The early VIO systems are often formulated as filters, \eg \cite{strelowMotionEstimationImage2004,vethFusionImagingInertial2006}.
In the evolution of filters for VIO, the MSCKF proposed in \cite{mourikisMultistateConstraintKalman2007} had been a milestone, 
enabling real-time motion estimation with a monocular camera-IMU setup.
It had some bearing on many subsequent filtering approaches, \eg, 
\cite{jonesVisualinertialNavigationMapping2011, weissRealtimeOnboardVisualinertial2012, bloeschIteratedExtendedKalman2017}.
Among them, these methods \cite{liHighprecisionConsistentEKFbased2013,heschCameraIMUbasedLocalizationObservability2014,
wuSquareRootInverse2015,sunRobustStereoVisual2018, genevaOpenVINSResearchPlatform2019} 
can be viewed as derivatives of MSCKF, enhancing it to be consistent and to deal with multiple cameras.

However, structureless filters that update motion estimates with every arrived image, \eg, variants of MSCKF,
are prone to drift during hovering or stationary periods.
This issue was mitigated by switching to a last-in-first-out state management strategy in \cite{guoEfficientVisualinertialNavigation2014}.
Later, ZUPT was adopted in \cite{genevaOpenVINSResearchPlatform2019}.
These approaches require heuristics for motion detection which often are not portable. For instance, ZUPT may not work well with hovering.

To bridle drift of a structureless filter in degenerate motion, 
we leverage the concept of keyframes in both the feature association frontend and the state estimation backend.

Though keyframes have been widely adopted in optimization-based visual SLAM methods \cite{younesKeyframebasedMonocularSLAM2017}, \eg, \cite{leuteneggerKeyframebased2015,qinVINSMono2018},
where the keyframe schemes help reduce computation in optimizing local maps,
very few filters incorporate the keyframe concept perhaps because of efficiency in themselves.
The VIO filters in \cite{zhuEventbasedVisualInertial2017,sunRobustStereoVisual2018} used keyframe-based state management in the backend similar to VINS-Mono \cite{qinVINSMono2018}.
In \cite{flintVisualbased2016}, a preprocessing module selects a subset of frames as keyframes 
and poses of only keyframes are estimated in the backend inverse SWF. However, we empirically observed that only using keyframes made the estimated trajectory wiggle.
To deal with stationary periods, the EKF proposed in \cite{abeywardenaFastOnboardModelaided2016} keeps a cloned state variable 
for the latest keyframe in order to leverage epipolar constraints due to points observed in that keyframe and the current frame.
Overall, our proposed KSWF shares features with
\cite{liHighfidelity2014,leuteneggerKeyframebased2015,qinVINSMono2018,genevaOpenVINSResearchPlatform2019} 
to which the differences are summarized in Table~\ref{tab:related-work-comparison}.
In view of these studies, we believe that the proposed KSWF is the first keyframe-based filter that supports full self-calibration.
\begin{table}[!htb]
\caption{Differences and similarities of our work to related studies. RS: Rolling shutter. Prop.: Propagation.}
\label{tab:related-work-comparison}
\begin{tabular}{lccccc}
		\toprule
		Methods      & Ours    & 
		\begin{tabular}[c]{@{}l@{}}OKVIS\\\cite{leuteneggerKeyframebased2015}\end{tabular} & 
		\begin{tabular}[c]{@{}l@{}}Li \etal\\\cite{liHighfidelity2014}\end{tabular}
		& \begin{tabular}[c]{@{}l@{}}VINS-\\Mono\cite{qinVINSMono2018}\end{tabular}
	    & \begin{tabular}[c]{@{}l@{}}Open-\\VINS\cite{genevaOpenVINSResearchPlatform2019}\end{tabular} \\ \midrule
		\begin{tabular}[c]{@{}l@{}}Supported\\ setup\end{tabular}            & \begin{tabular}[c]{@{}l@{}}mono\\ stereo\end{tabular}          & \begin{tabular}[c]{@{}l@{}}mono \\ stereo\end{tabular}       & mono & mono                                                      & \begin{tabular}[c]{@{}l@{}}$\geq1$\\ cameras\end{tabular}         \\ \midrule
		\begin{tabular}[c]{@{}l@{}}Keyframe-\\ based\\ frontend\end{tabular} & $\checkmark$ & $\checkmark$ & $\times$ & $\times$& $\times$ \\ \midrule
		\begin{tabular}[c]{@{}l@{}}Keyframe-\\ based\\ backend\end{tabular}  & $\checkmark$ & $\checkmark$ & $\times$ & $\checkmark$ & $\times$ \\ \midrule
		\begin{tabular}[c]{@{}l@{}}Standstill\\ in mono\\ setup\end{tabular} & Work   & Work & Fail & Work & Fail \\ \midrule
		\begin{tabular}[c]{@{}l@{}}Calibrated\\ params\end{tabular}          & \begin{tabular}[c]{@{}l@{}}camera-\\ IMU\\ params\end{tabular} & \begin{tabular}[c]{@{}l@{}}camera \\ extrinsics\end{tabular} & \begin{tabular}[c]{@{}l@{}}camera-\\ IMU\\ params\end{tabular} & \begin{tabular}[c]{@{}l@{}}camera\\ extrinsics, \\ delay\end{tabular} & \begin{tabular}[c]{@{}l@{}}camera \\ extrinsics,\\ intrinsics\end{tabular}  \\ \midrule
		\begin{tabular}[c]{@{}l@{}}Deal with \\RS effect\end{tabular}         & \begin{tabular}[c]{@{}l@{}}IMU\\ prop.\end{tabular}            & N/A  & \begin{tabular}[c]{@{}l@{}}IMU\\ prop.\end{tabular}            & \begin{tabular}[c]{@{}l@{}}optic flow\\ correction\end{tabular} & \begin{tabular}[c]{@{}l@{}}interpol.\\ by poses\\ in state\end{tabular} \\ \bottomrule
	\end{tabular}
\end{table}

\section{Keyframe-based Sliding Window Filter with Self-Calibration}
\label{sec:method}
From two aspects, self-calibration and resilience to degeneration motion, this work studies the VIO problem, which is to track the pose (position and orientation)
and optionally calibrate the sensor parameters of a platform consisting of an IMU and 
at least one camera using inertial readings and image observations of
opportunistic landmarks with unknown positions.
Considering both aspects, we propose the KSWF with self-calibration.
It has two keyframe-based components: the feature association frontend which extracts feature matches from images, and the estimation backend which fuses feature matches
and inertial measurements to estimate motion and optional parameters.
Next, we present notation conventions, IMU and camera measurement models, state variables, the frontend, and the backend filter.

\subsection{Notation and Coordinate Frames}
\label{subsec:notation}
The Euclidean transformation from reference frame $\{B\}$ to $\{A\}$ is represented by a transformation matrix $\mbf{T}_{AB}$ which consists of a $3\times 3$ rotation matrix $\mbf{R}_{AB}$
and a 3D translation vector $\mbf{p}_{AB}^A$ which is coordinates of $\{B\}$'s origin in $\{A\}$, \ie,
$\mbf{T}_{AB} = \left[\begin{matrix}
\mbf{R}_{AB} & \mbf{p}_{AB}^A \\
\mbf{0}\tran & 1
\end{matrix}\right]$,
For brevity, we will write $\mbf{p}_{AB}^A$ as $\mbf{p}_{AB}$ or $\mbf{p}^A$ when its meaning is clear.

This work uses several right-handed coordinate frames, camera frames $\{C_k\}$, $k =0, 1$,
body frame $\{B\}$, and world frame $\{W\}$.
The right-down-forward frame $\{C_k\}$ is affixed to the lens optical center of a camera $k$.
Without loss of generality, we assign index 0 to the left camera of a stereo rig and refer to it as the main camera.
$\{B\}$ is the frame affixed to the platform which estimated poses by a VIO method refer to.
It is defined by using the accelerometer $x$- and $y$- input axes. 
Therefore, the gyroscope input frame $\{\omega\}$ can have a relative rotation to $\{B\}$.
Moreover, a quasi-inertial world frame $\{W\}$ is 
instantiated at the start with accelerometer measurements 
such that its $z$-axis is along negative gravity.

\subsection{IMU Measurements}
To consider systematic and random errors 
in the three-axis gyroscope and accelerometer, 
we use an IMU model close to \cite{rehderExtendingKalibrCalibrating2016, schubertTUMVIBenchmark2018} which 
accounts for random biases and noises, and systematic errors (also known as IMU intrinsic parameters) including scale factor errors,
misalignment, relative orientations, and $g$-sensitivity.

The accelerometer measures the specific force applied to the IMU in $\{B\}$, $\mbf{a}_s^B$. 
The measurement $\mbf{a}_m$ is affected by
systematic errors encoded in a lower triangular matrix $\mbf{M}_a$, 
accelerometer bias $\mbf{b}_a$ and noise process $\mbs{\nu}_a$,
\begin{equation}
	\label{eq:accel_model}
	\begin{split}
		\mbf{a}_s^B &= \mbf{R}_{WB}\tran \cdot (\dot{\mbf{v}}_{WB} + \begin{bmatrix}0 & 0 & g\end{bmatrix}\tran) \\
		\mbf{a}_s^B  &= \mbf M_a (\mbf{a}_m - \mbf{b}_a - \mbs{\nu}_a) \\
		\dot{\mbf{b}}_a &= \mbs{\nu}_{ba},
	\end{split}
\end{equation}
where $\mbf v_{WB}$ is platform velocity in $\{W\}$, 
$g$ is gravity magnitude, and
$\mbf{b}_a$ is assumed to be driven by Gaussian white noise process 
$\mbs{\nu}_{ba}$.
$\mbf{M}_a$ represents 3-DOF (degree-of-freedom) scale factor errors, 3-DOF misalignment of the accelerometer.

The gyroscope measures angular rate of the platform in $\{B\}$, $\mbs{\omega}_{WB}^B$.
The measurement $\mbs{\omega}_m$ is affected by systematic errors encoded in a 
$3\times 3$ matrix $\mbf{M}_g$, $g$-sensitivity effect encoded in a $3\times 3$ matrix $\mbf{T}_s$,
gyroscope bias $\mbf{b}_g$, and noise process $\mbs{\nu}_g$,
\begin{equation}
	\label{eq:gyro_model}
	\begin{split}
		\mbs{\omega}_{WB}^B &= \mbf{M}_g [\mbs{\omega}_m - \mbf{b}_g - \mbs{\nu}_g - 
		\mbf{T}_s (\mbf{a}_m - \mbf{b}_a - \mbs{\nu}_a)] \\
		\dot{\mbf{b}}_g &= \mbs{\nu}_{bg}.
	\end{split}
\end{equation}
where $\mbf{b}_g$ is assumed to be driven by Gaussian white noise process $\mbs{\nu}_{bg}$.
$\mbf{M}_g$ represents 3-DOF scale factor errors, 3-DOF misalignment,
and 3-DOF orientation between gyroscope input axes and $\{B\}$.
Here, $\mbf M_a$ and $\mbf M_g$ are defined as such for generality,
but they may be simplified to be diagonal or lower triangular given extra knowledge of the IMU.

The power spectral densities of $\mbs{\nu}_{a}$, $\mbs{\nu}_{g}$, 
$\mbs{\nu}_{ba}$, and $\mbs{\nu}_{bg}$, 
are usually assumed to be $\sigma^2_a\mbf{I}_3$, $\sigma^2_g \mbf{I}_3$, $\sigma^2_{ba}\mbf{I}_3$, and $\sigma^2_{bg}\mbf{I}_3$, respectively,
but they may have different diagonal values, and nonzero off-diagonal entries, \eg, for a 2-DOF gyro.

To predict pose and velocity and their covariance, propagation with IMU data by the trapezoidal rule is carried out as described in 
\cite[Appendix A]{huaiCollaborativeSLAMCrowdsourced2017}, \cite{jekeliInertialNavigationSystems2001}.
We also extend the propagation to support integration backward in time so as to
predict camera poses for features in images captured by a RS camera.
The propagation step uses the gravity norm provided by the user \eqref{eq:accel_model}.

\subsection{Camera Measurements}
The proposed method supports image streams from one camera or two coarsely synchronized cameras.
In the stereo setup, two frames of the two cameras are grouped into a frame bundle when they have close timestamps per the camera clock.
In a frame bundle of index $j$, we denote the original time of an image $k$, $k = 0, 1$, stamped by the camera clock, by $t_{j}^{C_k}$.
Camera $k$'s clock has an offset $t_d^k$ relative to the IMU clock.
Then, the timestamp by the IMU clock for the central row of image $k$, $t_j^k$, is given by 
\begin{equation}
t_j^k = t_{j}^{C_k} + t_{d}^k.
\end{equation}

In the filter, state variables will be stamped by the estimated IMU time for the main camera's images. 
Suppose that the latest time offset estimate of the main camera as frame bundle $j$ arrives is $t_{d,j}^0$,
then the time for state variables associated with bundle $j$ is set to a constant value $t_j$ given by 
\begin{equation}
	\label{eq:state_epoch}
	t_j = t_{j}^{C_0} + t_{d,j}^0.
\end{equation}

With mid-exposure time of an image, we can compute the observation time per the IMU clock of any row in an image.
Denote the frame readout time for camera $k$ by $t_{r}^k$.
Suppose that a landmark $i$ has an observation at row $v$ in image $k$,
then its timestamp $t_{i, j}^k$ is computed by
\begin{align}
	\label{eq:feature_time}
	t_{i, j}^k = t_{j}^{k} + (\frac{v}{h} - \frac{1}{2}) t_{r}^k
\end{align}
where $h$ is the height of image $k$.
Using the central row of the RS image as the reference epoch instead of its first row
reduces the average prediction range and hence computation when predicting camera poses with IMU data for landmark observations.

We use the conventional reprojection model for landmark observations in images.
Consider a landmark $\mbf{L}_i$ observed in image $k$, $k=0, 1$, of frame bundle $j$ at $\mbf{z}_{i,j}^k = [u, v]\tran$, 
and denote the reprojection function of camera $k$ by $\mbf h^k(\cdot)$ which factors in all camera parameters, $\mbf{x}_C^k$, including
the camera extrinsic and intrinsic parameters, camera time offset $t_d^k$, 
and frame readout time $t_r^k$.

The reprojection model varies slightly depending on how the landmark is parameterized.
If the landmark is expressed in $\{W\}$, 
$\mbf{p}_i^W = [x_i, y_i, z_i, 1]\tran$, 
the model is given by
\begin{equation}
\label{eq:reprojection}
\mbf{z}_{i, j}^k = \mbf{h}^k(
\mbf{T}_{BC_k}^{-1} \mbf{T}_{WB(t_{i,j}^k)}^{-1} 
\mbf{p}_i^W) + \mbf{w}_c
\end{equation}
where $B(t_{i,j}^k)$ denotes the body frame at feature time $t_{i,j}^k$, and we assume that an image observation is affected by
Gaussian noise $\mbf{w}_c \sim N(\mbf{0}, \sigma_c^2\mbf{I}_2)$.
This reprojection model is used for observability analysis in Section~\ref{sec:observability} for simplicity.

If landmark $\mbf{L}_i$ is anchored in image $b$, $b \in \{0, 1\}$, of frame bundle $a$, and represented with its inverse depth,
the reprojection model will involve the body frame at bundle $a$'s assigned epoch $t_a$ \eqref{eq:state_epoch}, $\{B(t_a)\}$.
Now $\mbf{L}_i$ is expressed in the camera frame at $t_a$ of image $b$, $\{C_b(t_a)\}$, \ie,
$\mbf{p}_i^{C_b(t_a)} = \frac{1}{\rho_i}[\alpha_i, \beta_i, 1, \rho_i]\tran$, with inverse depth $\rho_i$ and observation direction $[\alpha_i, \beta_i, 1]$,
the observation model is given by
\begin{equation}
\label{eq:reprojection_anchor}
\mbf{z}_{i, j}^k
= \mbf{h}^k(
\mbf{T}_{BC_k}^{-1} \mbf{T}_{WB(t_{i,j}^k)}^{-1} 
\mbf{T}_{WB(t_a)} \mbf{T}_{BC_{b}} \cdot
\mbf{p}_{i}^{C_b(t_a)}) + \mbf{w}_c.
\end{equation}
This model is used in our experiments because anchored landmark representation does not cause inconsistency \cite{huaiConsistent2021}.

For the camera observation, the estimator will need measurement Jacobians of $\mbf{z}_{i, j}^k$.
Among them, one key component is Jacobians of the pose at feature time $t_{i,j}^k$, $\mbf{T}_{WB(t_{i,j}^k)}$,
relative to the pose and velocity at state time $t_j$, $\mbf{T}_{WB(t_j)}$ and $\mbf{v}_{WB(t_j)}$, respectively.
Both are computed along with IMU propagation from $t_j$ to $t_{i,j}^k$.
Jacobians relative to time parameters, \eg, $t_d^k$, $k=0, 1$,
are computed with the propagated velocity $\mbf{v}_{WB}(t_{i,j}^k)$ and fitted angular rate
$\mbs{\omega}_{WB}^B(t_{i,j}^k)$.
Jacobians relative to IMU parameters are ignored as in
\cite{liVisionaidedInertialNavigation2014} because their entries are usually small (with first order terms up to $O(t_r)$).

Our estimation framework supports multiple camera models. 
When the camera intrinsic parameters are to be calibrated, we recommend using the pinhole radial tangential 8-parameter model \cite{bradskiLearningOpenCVComputer2008} or the Kannala Brandt 6-parameter model \cite{kannalaGeneric2006}
as the observability analysis in Section~\ref{sec:observability} shows that their parameters are observable.
Without loss of generality, we assume the pinhole radial tangential 8-parameter model
\eqref{eq:camera_params} is used in the following unless stated otherwise.

\subsection{State Variables in the Filter}
\label{subsec:state-and-error-state}
The state vector of the KSWF at time $t$, $\mbf{x}(t)$, consists of current navigation state variable $\mbs{\pi}(t)$,
IMU parameters $\mbf{x}_S$, parameters of $N$ cameras $\mbf{x}_C$,
and a sliding window of past navigation state variables $\mbf{x}_W$
at known epochs $\{t_j\}$
where $j$ enumerates $N_{kf}$ keyframe bundles and
$N_{tf}$ recent temporal frame bundles, and landmarks $\mbf x_L$.
That is,
\begin{align}
\label{eq:state_vector}
\mbf{x}(t) &= \{\mbs{\pi}(t), \mbf{x}_{S}, \mbf{x}_C,
\mbf{x}_W, \mbf x_L\}\\
\label{eq:nav_states}
\mbs{\pi}(t) &= \{\mbf{p}_{WB}(t), \mbf{R}_{WB}(t), \mbf{v}_{WB}(t)\} \\
\label{eq:IMU-parameters}
\mbf x_S &= \left\{\mbf b_g(t) \enskip \mbf b_a(t)
\enskip \vec{\mbf{M}}_g \enskip \vec{\mbf{M}}_s 
\enskip \vec{\mbf{M}}_a \right\} \\
\label{eq:ncamera_params}
\mbf{x}_C &= \{\mbf{x}_{C}^k | k = 0, 1\} \\
\label{eq:slidingwindow_states}
\mbf{x}_W &= \{\mbs{\pi}(t_j) | j = 0, 1, \cdots N_{kf} + N_{tf} -1 \} \\
\mbf x_L &= \{\mbf {L}_i | i = 0, 1, \cdots \}
\end{align}
where $\vec{(\cdot)}$ concatenates rows of a matrix into a vector.

\begin{table}[!htb]
	\ra{1.3}
	\caption{State variables estimated by KSWF for a stereo camera-IMU setup.
		The pinhole radial tangential model is assumed.}
	\label{tab:state-variables}
	\begin{tabular}{@{}lll@{}}
		\toprule
		State variables & Time-varying                                      & Time-invariant      \\ \midrule
		Motion          & $\mbf{T}_{WB(t)}$, $\mbf{v}_{WB(t)}$ & \begin{tabular}[c]{@{}l@{}}$\{\mbf{T}_{WB(t_j)}, \mbf{v}_{WB(t_j)}\}$\\ $j=0, 1, \cdots, N_{kf} + N_{tf} - 1$\end{tabular} \\
		IMU             & $\mbf{b}_g,  \mbf{b}_a$                     & $\mbf{M}_a, \mbf{M}_g,\mbf{M}_s$                                   \\ \midrule
		Camera          & \multicolumn{2}{l}{Time-invariant}                                      \\ \midrule
		Extrinsics      & \multicolumn{2}{l}{$\mbf{T}_{BC_0}$, $\mbf{T}_{BC_1}$}          \\
		Intrinsics      & \multicolumn{2}{l}{$\{f_x^k, f_y^k, c_x^k, c_y^k, k_1^k, k_2^k, p_1^k, p_2^k\}$, $k = 0, 1$}                                      \\
		Time offset     & \multicolumn{2}{l}{$t_r^k$, $k = 0$, 1}                                   \\
		Readout time    & \multicolumn{2}{l}{$t_d^k$, $k = 0$, 1}                                   \\ \bottomrule
	\end{tabular}
\end{table}

For a monocular camera-IMU setup, the camera parameters suitable for calibration are
\begin{equation}
	\label{eq:camera_params}
	\mbf{x}_C = \{\mbf{T}_{BC_0}, f_x^0, f_y^0, c_x^0, c_y^0, k_1^0, k_2^0, p_1^0, p_2^0, t_d^0, t_r^0\}
\end{equation}
Parameters that can be estimated by the filter for a stereo camera-IMU setup are listed in Table~\ref{tab:state-variables}.

Each navigation state in the sliding window $\mbs{\pi}(t_j)$ associated with frame bundle $j$ is at a known epoch $t_j$ \eqref{eq:state_epoch}.
Tying navigation state variables to known epochs is conceptually simpler than tying them to true epochs which are uncertain as in 
\cite{genevaOpenVINSResearchPlatform2019, liVisionaidedInertialNavigation2014}.
The past navigation variables also include velocity so that 
the RS camera pose for a landmark observation can be
propagated from the relevant $\mbs{\pi}(t_j)$ with IMU measurements.

Camera parameters $\mbf{x}_C$ and IMU intrinsic parameters are assumed to be unknown constants as in \cite{liHighfidelity2014, genevaOpenVINSResearchPlatform2019},
for they change little in the timespan of the sliding window.
However, if a sensor parameter is in fact not constant, 
noise can be added to account for its drift over time as done for IMU biases.

The error $\delta\mbf{x}$ for a state variable $\mbf{x}$ in a vector space is defined to be
$\mbf{x} = \hat{\mbf{x}} + \delta\mbf{x}$,
where $\hat{\mbf{x}}$ is its estimate.
For the rotation matrices on a Lie group, the error state is defined by the Lie algebra.
For example, $\mbf{R}_{WB}$'s error state $\delta \mbs{\theta}_{WB}$ is defined by
$\mbf{R}_{WB} = \exp(\delta \mbs{\theta}_{WB}^\times) \hat{\mbf{R}}_{WB}
 \approx (\mbf{I} + \delta \mbs{\theta}_{WB}^\times) \hat{\mbf{R}}_{WB}$,
where $(\cdot)^\times$ obtains the skew-symmetric matrix of a 3D vector.

\subsection{Feature Association Frontend}
\label{subsec:feature-association}
\begin{figure}[!htbp]
	\centering
	\includegraphics[width=0.75\columnwidth]{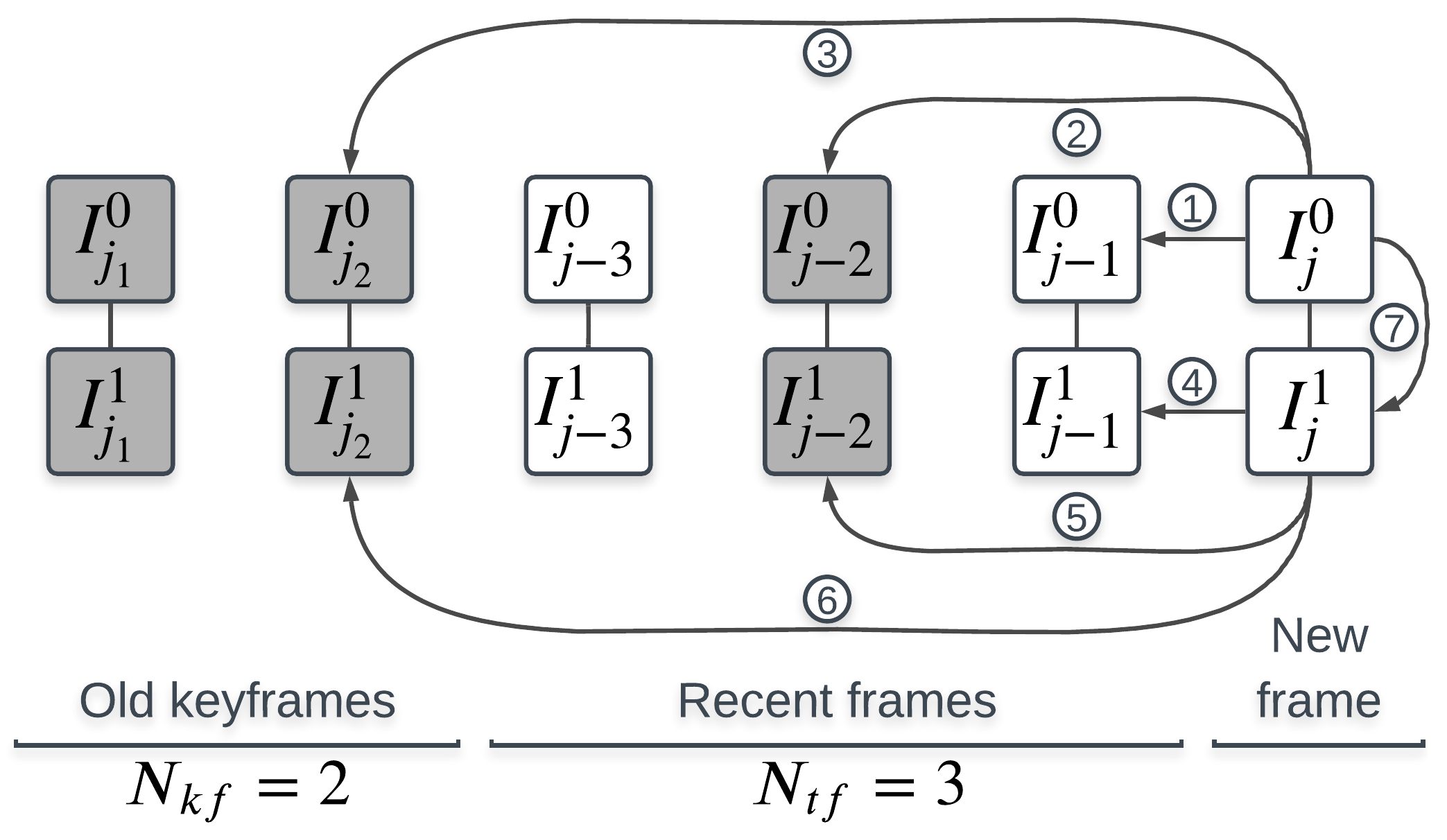}
	\caption{The keyframe-based feature matching consists of matching the current
		frame bundle to the last frame bundle, \numcircledmod{1} and \numcircledmod{4}, matching the
		current frame bundle to two previous keyframe bundles, \numcircledmod{2}, \numcircledmod{3},
		\numcircledmod{5}, and \numcircledmod{6}, and matching between images of the
		current frame bundle, \numcircledmod{7}. Showing keyframe bundles as shaded frame bundles, this
		diagram assumes a platform with stereo cameras.
		For a single camera setup, only \numcircledmod{1}-\numcircledmod{3} are needed.
	}
	\label{fig:feature-matching}
\end{figure}

The frontend extracts image features by using descriptors,
and matches descriptors from the current frame bundle to the last bundle as well as several previous keyframe bundles, as illustrated in 
Fig.~\ref{fig:feature-matching}.
Our frontend matches the current frame bundle to at most two previous keyframe bundles because matching to even earlier keyframes results in so few matches that their validity cannot be checked by 5-point RANSAC \cite{nisterEfficientSolutionFivePoint2003}.
We choose the BRISK \cite{leuteneggerBRISKBinaryRobust2011} to detect and describe keypoints for its sub-pixel keypoint location (w.r.t. ORB \cite{rubleeORB2011}), efficiency, and ability to handle large disparity (w.r.t. KLT \cite{bradskiLearningOpenCVComputer2008}) which is required in matching to a keyframe. 

Unlike OKVIS \cite{leuteneggerKeyframebased2015} and the ORB-SLAM trilogy \cite{camposORBSLAM32020} where the frontend uses 3D landmarks owned by the backend estimator for 3D-2D matching, our frontend only performs 2D-2D matching of keypoints between frames as in VINS-Mono \cite{qinVINSMono2018} and OpenVINS \cite{genevaOpenVINSResearchPlatform2019}, and thus is generally decoupled from the backend. The insight is that the coupling greatly increases implementation and maintenance complexity while bringing about marginal accuracy improvement.
Another difference to these frameworks is that our frontend regularly updates the camera parameters for 5-point RANSAC \cite{nisterEfficientSolutionFivePoint2003} by using estimates sent from the backend estimator.

The frontend distinguishes ordinary frame bundles and keyframe bundles 
by using the view overlap criteria in \cite{leuteneggerKeyframebased2015}.
For every image $k$, $k=0, 1$, in the current frame bundle, among its features $\mathcal F_k$, we find those associated with feature tracks in existing keyframe bundles, $\mathcal M_k$.
Then, we compute the overlap $o_k$ between
the convex hull of $\mathcal M_k$ and that of $\mathcal F_k$, and the ratio $r_k$ between 
$\vert\mathcal M_k\vert$ and the number of features in image $k$, $\vert\mathcal F_k\vert$.
A keyframe bundle is selected if the maximum of area overlaps $\{o_k| k \in [0, 1]\}$ is
less than a threshold $T_o$ (typically 60\%) or 
if the maximum of match ratios $\{r_k| k \in [0, 1]\}$ is less than $T_r$ (typically 20\%).

\subsection{Filter Workflow}
\label{subsec:filter}
\begin{figure}[!t]
	\centering
	\includegraphics[width=\columnwidth]{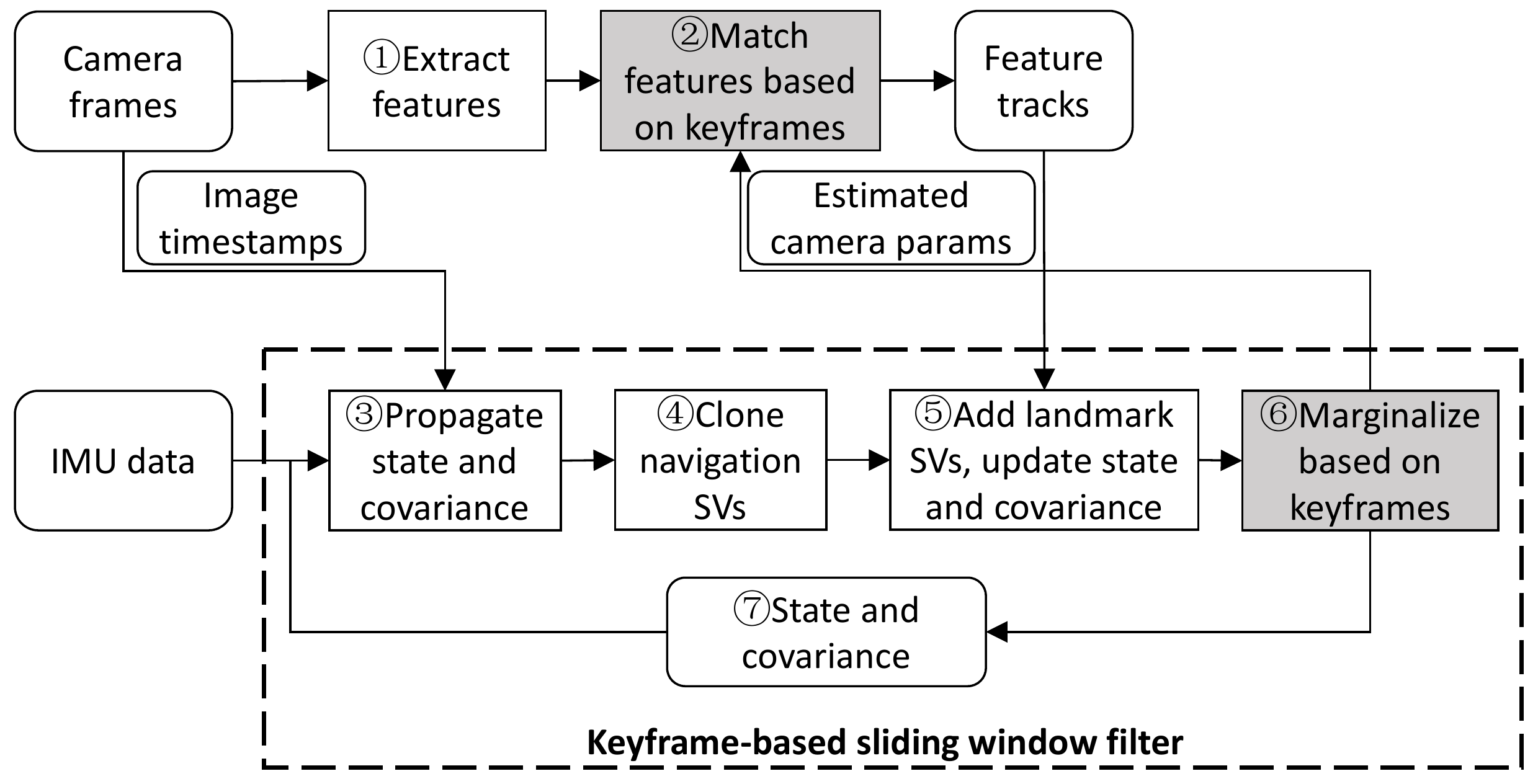}
	\caption{Flowchart of the keyframe-based sliding window filter. SV: state variable.
		\numcircledmod{1}-\numcircledmod{2} are discussed in \ref{subsec:feature-association}.
		\numcircledmod{3}-\numcircledmod{6} are presented in \ref{subsec:filter}.
		\numcircledmod{7} is described in \ref{subsec:state-and-error-state}.}
	\label{fig:flowchart}
\end{figure}

The backend of KSWF estimates state variables tied to frame and keyframe bundles using feature observations and 
marginalizes redundant state variables (Fig.~\ref{fig:flowchart}).
The filter manages motion-related state variables depending on whether
they are associated with keyframe bundles.
The keyframe-based state management scheme is facilitated by the selection of keyframe bundles in the frontend.

The proposed filter is outlined in Algorithm~\ref{algo:KSWF}.
At the start, the VIO system is initialized to zero position and zero velocity.
The orientation is set to meet the requirement that $z$-axis of $\{W\}$ is along negative
gravity by using a few accelerometer data.
Since the filter enforces that requirement, the initial coarse orientation will be refined subsequently.
The biases are set by averaging the IMU data if the optic flow in the first three frames is small.
Otherwise, they are set to zero.
Remaining IMU and camera parameters are initialized to values from datasheets or experience.
The standard deviations of state variables
are initialized to sensible values (see Table~\ref{tab:init_value_std} 
for simulation).
The simplistic initialization works well even when the system starts in motion (see Fig.~\ref{fig:stationary-mh}).
Nevertheless, the initialization can be refined by a visual-inertial BA for the first few keyframes which deals with inaccurate calibration.

\begin{figure}[!htb]
\centering
\includegraphics[width=0.95\columnwidth]{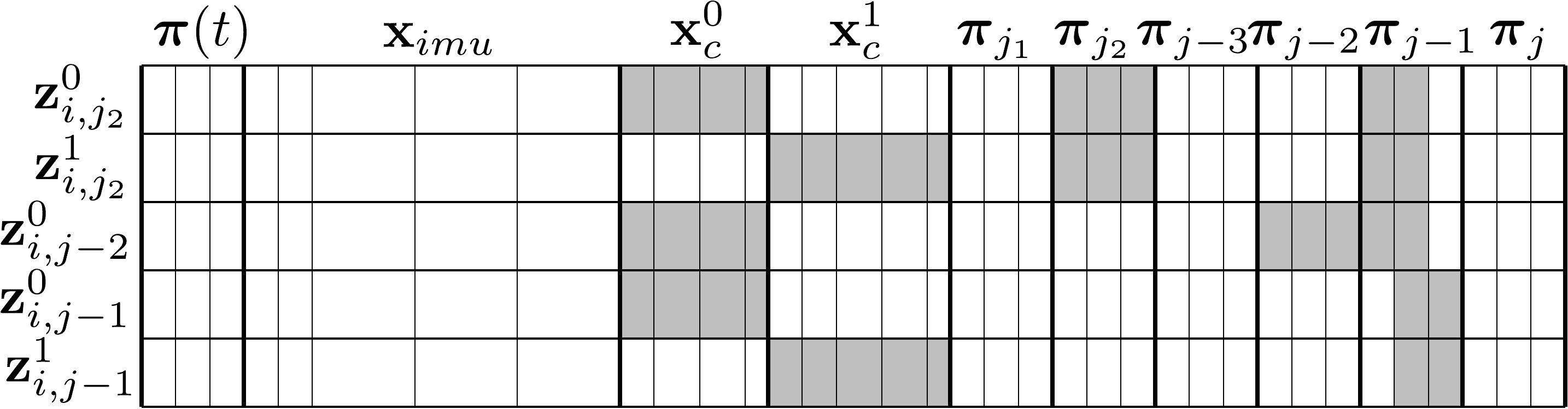}
\caption{Schematic drawing of observation Jacobians of a landmark $i$ to be marginalized, where
nonzero entries are shaded.
For clarity, the landmark state variables are not shown.
In accordance with Fig.~\ref{fig:feature-matching}, as frame bundle $j$ arrives, landmark $i$ is assumed to
complete its feature track at frame bundle $j-1$ which will serve as its anchor
\eqref{eq:reprojection_anchor}. After the navigation state variable for frame bundle
$j$ has been cloned as $\mbs{\pi}_j:=\mbs{\pi}(t_j)$, the computed
Jacobians for landmark $i$ are stacked as shown and ready for canceling out the
Jacobian for $i$'s parameters (Section \ref{subsec:filter}).}
\label{fig:jacobians}
\end{figure}

As a frame bundle $j$ arrives, the navigation state $\mbs{\pi}(t)$ and 
the covariance matrix are propagated with IMU data to the state epoch $t_j$
(Box~\numcircledmod{3} of Fig.~\ref{fig:flowchart}).
Then, $\mbs{\pi}(t)$ is cloned and appended to the state vector and the
covariance matrix is also expanded for the cloned $\mbs{\pi}(t_j)$
(Line~\ref{KSWF:line:propagate}, Box~\numcircledmod{4} of Fig.~\ref{fig:flowchart}).
As $\mbs{\pi}(t_j)$ is at a known epoch $t_j$,
unlike \cite{genevaOpenVINSResearchPlatform2019} or \cite[(30)]{liVisionaidedInertialNavigation2014}, we
need not to account for uncertainty in its time when augmenting
covariance for $\mbs{\pi}(t_j)$. Instead, we need to compute the measurement Jacobian
relative to camera time offsets in the update step (Line~\ref{KSWF:line:disappear-end}).

In parallel, feature descriptors are extracted from images of bundle $j$ and
matched to those of the last bundle $j-1$ and several recent keyframe bundles
(\eg, bundle $j-2$ and $j_2$ in Fig.~\ref{fig:feature-matching}).
After that, three types of feature tracks will be used for update (Box~\numcircledmod{5} of Fig.~\ref{fig:flowchart}): tracks that disappear in bundle $j$ (Lines~\ref{KSWF:line:disappear}-\ref{KSWF:line:disappear-end}),
observations of landmarks in the state vector (Lines~\ref{KSWF:line:instate}-\ref{KSWF:line:instate-end}), 
and tracks that can triangulate a new landmark to the state (Lines~\ref{KSWF:line:newlandmark}-\ref{KSWF:line:newlandmark-end}).
They differ in how to prepare for the update. For disappeared tracks, 
Jacobians relative to their landmark parameters need to be canceled out by matrix nullspace projection \cite{mourikisMultistateConstraintKalman2007} (Line~\ref{KSWF:line:canceljac}).
For the third type, triangulated new landmarks will be augmented to the state and covariance
(Line~\ref{KSWF:line:addnewlandmark}).
For all three observation types, the update is carried out identical to the classic EKF \cite{jekeliInertialNavigationSystems2001,huaiCollaborativeSLAMCrowdsourced2017}.
Note that it is possible to use an iterated update scheme for disappeared features.
However, empirically, despite additional computation, 
its improvement on accuracy across datasets was marginal \cite{bloeschIteratedExtendedKalman2017}.
Moreover, we do not rescale the variables despite their vastly different orders of magnitude, because the filter uses error state variables which are consistently small in value and standard deviation.

For a landmark $i$ associated with a disappeared feature track, stacked Jacobians of its observations relative to the
state vector are visualized in Fig.~\ref{fig:jacobians}.
For a landmark of large depth, the observation Jacobian relative to the landmark's parameters (\ie, the landmark Jacobian) can be rank-deficient,
and this fact should be considered in canceling out the landmark Jacobian and in the following Mahalanobis gating test.

To bound computation, redundant frame bundles are selected and marginalized from
the filter (Lines~\ref{KSWF:line:detect-dud-frames}-\ref{KSWF:line:remove-dud-variables}, Box~\numcircledmod{6} of
Fig.~\ref{fig:flowchart}), once the number of navigation
state variables in the sliding window exceeds $N_{kf} + N_{tf}$.
In a marginalization operation, at least $N_r$ redundant bundles 
($N_r$ is 3 for a monocular setup, 2 for a stereo setup) are
chosen (Line~\ref{KSWF:line:detect-dud-frames}) since two reprojection
measurements for an unknown landmark is
uninformative \cite{sunRobustStereoVisual2018}. To meet this requirement, redundant bundles are chosen
first among the recent non-keyframe bundles while excluding the most recent $N_{tf}$ bundles
and secondly among the oldest keyframe bundles. For the case of
Fig.~\ref{fig:feature-matching}, $N_{kf}=2$ and $N_{tf}=3$, the
redundant bundles are keyframe bundle $j_1$ and frame bundle $j-3$.

With these redundant bundles, we update the filter with observations of landmarks,
each observed more than twice in these bundles
(Line~\ref{KSWF:line:update-with-dud-frames}).
For such a landmark, if it can be triangulated with its entire observation history, its
observations in the redundant bundles are used for EKF update. 
Other observations in the redundant bundles are discarded.
After the update, state variables and
entries in the covariance matrix for these redundant bundles are removed
(Line~\ref{KSWF:line:remove-dud-variables}).

\begin{algorithm}
\caption{KSWF workflow.}
	\label{algo:KSWF}
	\begin{algorithmic}[1]
		\State Initialize state and covariance
		\label{KSWF:line:initialize}
		\While{a frame bundle $j$ arrives}
		\State Propagate state variable $\mbs{\pi}(t)$ to $t_j$ with IMU data
		\label{KSWF:line:propagate}
		\State Augment covariance for the cloned state variable $\mbs{\pi}(t_j)$
		\State Extract descriptors and match to the last frame bundle and previous keyframe bundles
		\label{KSWF:line:descriptor}
		\For{disappeared feature tracks}\label{KSWF:line:disappear}
		\State Triangulate the point landmark
		\label{KSWF:line:triangulate}
		\State Compute residuals and Jacobians of reprojection errors
		\State Cancel out Jacobians for the landmark
		\label{KSWF:line:canceljac}
		\State Remove outliers with Mahalanobis test
		\label{KSWF:line:mahal}
		\EndFor
		\State Update state and covariance
		\label{KSWF:line:disappear-end}
		
		\For{feature tracks of landmarks in state}\label{KSWF:line:instate}
		\State Compute residuals and Jacobians of reprojection errors
		\State Remove outliers with Mahalanobis test
		\EndFor
		\State Update state and covariance\label{KSWF:line:instate-end}
		
		\For{feature tracks of well-observed landmarks but not in state}
		\label{KSWF:line:newlandmark}
		\State Triangulate the point landmark
		\State Compute residuals and Jacobians of reprojection errors
		\State Remove outliers with Mahalanobis test
		\EndFor
		\State Augment state and covariance for these landmarks\label{KSWF:line:addnewlandmark}
		\State Update state and covariance
		\label{KSWF:line:newlandmark-end}
		
		\If{Redundant frame bundles are detected}
		\label{KSWF:line:detect-dud-frames}
		\State Update with features in these frame bundles \Comment{As with the disappeared features}
		\label{KSWF:line:update-with-dud-frames}
		\State Remove associated state variables and covariance entries
		\label{KSWF:line:remove-dud-variables}
		\EndIf
		\EndWhile
	\end{algorithmic}
\end{algorithm}

To ensure filter consistency, for IMU propagation and camera measurements, Jacobians are evaluated
at propagated values of position and velocity (\ie, first estimates
\cite{liHighprecisionConsistentEKFbased2013}),
and at the latest estimates of other variables (\eg, IMU biases, landmarks expressed by anchored inverse depth), since they
do not affect the unobservable directions \cite{huaiConsistent2021}.

\section{Observability Analysis}
\label{sec:observability}
By using the ORC of Lie derivatives, this section shows that the VIO system with self-calibration is weakly observable
under general motion while excluding the four immanent unobservable dimensions.
We first review the basics of observability analysis with Lie derivatives. 
Then, this method is applied to the monocular global shutter (GS) camera-IMU-based odometry with self-calibration while ignoring time variables.
Discussions on extensions are also given.
Lastly, we consider observability of time offset and RS readout time.

\subsection{Observability Analysis Fundamentals}
According to \cite{martinelliObservability2020}, a state at time $t_0$, $\mbf x(t_0)$, is weakly observable if there is a neighborhood in which all its neighbors can be
distinguished from itself by the knowledge of outputs and general (unconstrained) inputs in a time interval $\mathcal{I} = [t_0, t_0 + T]$.
For a noise-free system that is affine in the control inputs $\mbf u_i$, $i=1, 2, \cdots, p$,
\begin{equation}
\label{eq:affine-input-sys}
\dot{\mbf{x}} = \mbf{f}_0(\mbf x) + \sum_{i = 1}^{p}\mbf f_i(\mbf x) \mbf u_i, \quad 
\mbf{y} = \mbf{h(x)}
\end{equation}
the sufficient condition for weak observability of its state $\mbf x$ is that the observability matrix built from the outputs $\mbf h(\mbf x)$ has full rank
\cite{hermannNonlinearControllabilityObservability1977, martinelliObservability2020}.
The observability matrix $\mathcal{O}$ consists of gradients of Lie derivatives of
$\mbf h(\mbf x)$ along vector fields $\mbf f_i$ of the control inputs $\mbf u_i$.
For a vector output function $\mbf h$, its first order Lie derivative along the vector field $\mbf f_i = [\mbf f_{i1}\enskip \mbf f_{i2} \enskip \cdots \enskip \mbf f_{is}]$ of $s$ columns is defined by
\begin{equation}
	\mathcal{L}^1_{\mbf f_i} \mbf{h} = 
	\begin{bmatrix}
		\nabla \mbf h \cdot \mbf f_{i1} \\
		\nabla \mbf h \cdot \mbf f_{i2} \\
		\vdots \\
		\nabla \mbf h \cdot \mbf f_{is}
	\end{bmatrix}
\end{equation}
The zeroth order Lie derivative is defined to be
\begin{equation}
	\mathcal{L}^0 \mbf h = \mbf h.
\end{equation}
Higher order Lie derivatives can be computed recursively by 
\begin{equation}
\mathcal{L}^2_{\mbf f_i, \mbf f_j} \mbf{h} = \mathcal{L}^1_{\mbf f_j} \mathcal{L}^1_{\mbf f_i} \mbf{h}.
\end{equation}
With these Lie derivatives of the observation function, the entire observability matrix $\mathcal{O}$ is given by
\begin{equation}
\mathcal{O} = \begin{bmatrix}
\nabla \mathcal{L}^0 \mbf h \\
\nabla \mathcal{L}^1_{\mbf f_i} \mbf h \\
\nabla \mathcal{L}^2_{\mbf f_i, \mbf f_j} \mbf h \\
\vdots
\end{bmatrix},
\end{equation}
where $i, j = [0, 1, \cdots, p]$.
$\mathcal{O}$ can be viewed as a codistribution spanned by row vectors (also known as covectors).
That is, $\mathcal{O}$ can be written as 
\begin{equation}
\mathcal{O} = \mathrm{span}\{\nabla \mathcal{L}^0 \mbf h, \nabla \mathcal{L}^1_{\mbf f_i} \mbf h, \nabla \mathcal{L}^2_{\mbf f_i, \mbf f_j} \mbf h, \cdots\}
\end{equation}

The observability property of the affine-input system can be revealed by incrementally building up the codistribution and checking its dimension.
Denote the codistribution with gradients of Lie derivatives up to order $k$ by 
$\mathcal{O}_k$, \eg, $\mathcal{O}_0 = \mathrm{span}\{ \nabla \mbf h \}$ and 
$\mathcal{O}_1 = \mathrm{span}\{\nabla \mbf h,
\nabla \mathcal{L}^1_{\mbf f_0} \mbf h, \nabla \mathcal{L}^1_{\mbf f_1} \mbf h, \cdots, \nabla \mathcal{L}^1_{\mbf f_p} \mbf h\}$.
As deduced in \cite[Algorithm 4.2]{martinelliObservability2020}, if $\mathrm{rank}(\mathcal{O}_{k-1}) = \mathrm{rank}(\mathcal{O}_k)$, the incremental procedure completes.
By then, if $\mathrm{rank}(\mathcal{O}_{k}) = \dim(\mbf x)$, the system is weakly observable.

To simplify observability analysis, it is beneficial to proactively identify 
observable dimensions of the state $\mbf x$ in building up the codistribution.
Considering the codistribution at step $k$, $\mathcal{O}_k$, this can be done by using the below properties.
By \cite[Corollary 2.1]{maesObservability2019} (see a sketch of proof in Appendix~\ref{app:proof}), a component of $\mbf x$, $\mbf x_i$, is weakly observable 
if removing the column corresponding to $\mbf x_i$ from $\mathcal{O}_k$ reduces its rank by 1.
Alternatively, according to \cite[Theorem 4.8]{martinelliObservability2020}, $\mbf x_i$ is weakly observable 
if the unit covector $\nabla \mbf x_i$ belongs to the codistribution $\mathcal{O}_k$, \ie,
adding covector $\nabla \mbf x_i$ to $\mathcal{O}_k$ does not increase its rank.

\subsection{Weak Observability without Time Parameters}
This section investigates whether state variables of the monocular GS camera-IMU-based odometry with self-calibration are weakly observable
using observations from an unknown point landmark, except the well-known four unobservable directions.
The below analysis easily extends to multiple point landmarks without changing the conclusions.

We first build up the differential system with variables equivalent to those in \eqref{eq:state_vector}.
First, by ignoring noises, the calibrated IMU measurements are computed by
\begin{equation}
\begin{split}
\mbf{a}_t &= \mbf{M}_a (\mbf{a}_m - \mbf{b}_a), \\
\mbs{\omega}_t &= \mbf{M}_g (\mbs{\omega}_m - \mbf{b}_g) -
\mbf{M}_g \mbf{T}_s (\mbf{a}_m - \mbf{b}_a).
\end{split}
\end{equation}
Next, the unknown landmark $\mbf L_i$ is represented by inverse depth $\rho$ and direction $\mbs{\gamma}$ in the camera frame $\{C\}$,
\begin{equation}
\mbs{\gamma} = [\gamma_1, \gamma_2]\tran 
= [p^C_{i,x} / p^C_{i,z}, p^C_{i,y} / p^C_{i,z}]\tran, \enskip
\rho =  1 / \mbf p^C_{i,z},
\end{equation}
where $[p^C_{i,x}\enskip p^C_{i,y} \enskip p^C_{i,z}]$ are time-varying coordinates of the landmark in $\{C\}$.
Moreover, we use the gravity vector $\mbf g^B$ expressed in $\{B\}$ to encode gravity magnitude, and roll and pitch of the camera-IMU system.
Overall, the system state for VIO is simplified to 
\begin{equation}
\mbf{x} = \{\mbs{\gamma}, \rho, \mbf{v}^{B}, \mbf{g}^B,
\mbf{b}_g, \mbf{b}_a, \vec{\mbf{M}}_g,
\vec{\mbf{T}}_s, \vec{\mbf{M}}_a, \mbf{x}_{C}\}
\end{equation}
where $\mbf{x}_C = \{\mbf{p}_{CB}, \mbf{q}_{CB}, f_x, f_y, c_x, c_y, k_1, k_2, p_1, p_2\}$,
excluding temporal parameters $t_d^0$ and $t_r^0$ and dropping the camera index superscript for brevity.
As we express $\mbf{q}_{CB} = [q_w, q_x, q_y, q_z]\tran$, the state dimension is $\mathrm{dim}(\mbf x) = 54$.
Comparing to \eqref{eq:state_vector}, we see that the alternative system state covers all variables in the original monocular VIO system, 
while conveniently excluding $t_d^0$, $t_r^0$, and the unobservable $\mbf p_{WB}$ and yaw.

Assuming that biases are constant in a short timespan,
with basic derivations, the system model is found to be
\begin{equation}
\label{eq:system-model}
\begin{split}
\begin{bmatrix}
	\dot{\mbs{\gamma}} \\ \dot{\rho} \\ \dot{\mbf{v}}^{B} \\ \dot{\mbf{g}}^{B}
\end{bmatrix} &= 
\begin{bmatrix} 
\begin{array}{c}
	C_{\mbs{\gamma}\rho}[
	(\mbf{R}_{CB} \mbs{\omega}_t) \times (\rho \mbf{p}_{CB} - \bar{\mbs{\gamma}}) -\\
	\mbf{R}_{CB} \rho \mbf{v}^B]
\end{array} \\
	\mbf{v}^B \times \mbs{\omega}_t + \mbf{a}_t + \mbf{g}^B \\
	\mbf{g}^B \times \mbs{\omega}_t
\end{bmatrix} \\
&= \mbf f_0 + \mbf{f}_1 \mbf a_m + \mbf f_2 \mbs{\omega}_m \\
\dot{ \mbf{b}}_g &= \dot{\mbf{b}}_a = \mbf{0}_{3\times 1}, \quad \dot{\mbf q}_{CB} = \mbf 0_{4\times 1} \\
\dot{\mbf{M}}_g &= \dot{\mbf{M}}_s = \dot{\mbf{M}}_a = \mbf{0}_{3\times 3}, \quad
\dot{\mbf{x}}_{C} = \mbf{0}_{15\times 1}
\end{split}
\end{equation}
where the shorthand notations are 
\begin{equation}
	\bar{\mbs{\gamma}} = [\mbs{\gamma}\tran \quad 1]\tran, \quad
	C_{\mbs{\gamma} \rho} = \begin{bmatrix}
		\mbf{I}_2 & -\mbs{\gamma} \\ 0_{1\times2} & -\rho
	\end{bmatrix}.
\end{equation}
To save space, the coefficient vector fields $\mbf f_i$, $i=0,1,2$ are not expanded.

There are three types of observations to the system, point observation $\mbf h_1$,
gravity magnitude $h_2$, and the unit quaternion constraint $h_3$,
\begin{align}
\label{eq:oa-observations}
\mbf{h}_1 &= \left[\begin{array}{c}
	f_x [\gamma_1 (1 + k_1 r^2 + k_2 r^4) + 2 p_1 \gamma_1 \gamma_2 + \\ p_2 (r^2 + 2 \gamma_1^2)] + c_x\\
	f_y [\gamma_2 (1 + k_1 r^2 + k_2 r^4) + p_1 (r^2 + 2\gamma_2^2) + \\ 2 p_2 \gamma_1 \gamma_2] + c_y
\end{array}\right] \\
\label{eq:gravity-observation}
h_2 &= (\mbf{g}^B)\tran \mbf{g}^B \\
\label{eq:quaternion-constraint}
h_3 &= \mbf{q}_{CB}\tran \mbf{q}_{CB}	
\end{align}
where $r^2 = \mbs{\gamma}\tran \mbs{\gamma}$.

We analyze the system of \eqref{eq:system-model}, \eqref{eq:oa-observations}- \eqref{eq:quaternion-constraint} by incrementally constructing codistributions.
With first order Lie derivatives, we find $\mathrm{rank}(\mathcal{O}_1) = 18$.
With second order Lie derivatives, $\mathrm{rank}(\mathcal{O}_2) = 47$.
By using the earlier deduction technique, we identify that eight parameters of the pinhole radial tangential model,
are weakly observable.
The same procedure shows that the FOV 5-parameter model \cite{devernayStraightLinesHave2001}, the extended unified camera 6-parameter model \cite{khomutenkoEnhancedUnifiedCamera2016},
and the Kannala Brandt 6-parameter model \cite{kannalaGeneric2006} are weakly observable with the second order codistribution $\mathcal{O}_2$.

As a result, we can replace $\mbf h_1$ with
$\mbf{h}_1' = \mbs{\gamma}$ and remove camera intrinsic parameters from $\mbf{x}$,  giving a shorter state vector $\mbf x'$, simplifying the arduous analysis.
For the new system, the third order codistribution has full rank,
$\mathrm{rank}(\mathcal{O}_3) = 46 = \dim(\mbf x')$, implying that 
the system is weakly observable.
One set of covectors spanning $\mathcal{O}_3$ is given by

\begin{equation}
\begin{split}
\mathcal{O}_3 &= \mathrm{span}\{
\nabla \mathcal{L}^0_{}\mbf h_1,
\nabla \mathcal{L}^0_{} h_2,
\nabla \mathcal{L}^0_{} h_3,
\nabla \mathcal{L}^1_{\mbf f_0}\mbf h_1,
\nabla \mathcal{L}^1_{\mbf f_1}\mbf h_1, \\ &\quad
\nabla \mathcal{L}^1_{\mbf f_2}\mbf h_1,
\nabla \mathcal{L}^2_{\mbf f_0,\mbf f_0}\mbf h_1,
\nabla \mathcal{L}^2_{\mbf f_1,\mbf f_0}\mbf h_1,
\nabla \mathcal{L}^2_{\mbf f_2,\mbf f_0} h_{11}, \\ &\quad
\nabla \mathcal{L}^2_{\mbf f_0,\mbf f_{11}}\mbf h_1,
\nabla \mathcal{L}^2_{\mbf f_{11},\mbf f_{11}} h_{11},
\nabla \mathcal{L}^2_{\mbf f_0,\mbf f_{12}}\mbf h_1, \\ &\quad
\nabla \mathcal{L}^2_{\mbf f_0,\mbf f_{13}}\mbf h_1,
\nabla \mathcal{L}^2_{\mbf f_0,\mbf f_{21}}\mbf h_1,
\nabla \mathcal{L}^2_{\mbf f_0,\mbf f_{22}} h_{11}, \\ &\quad
\nabla \mathcal{L}^3_{\mbf f_0,\mbf f_0,\mbf f_0}\mbf h_1, 
\nabla \mathcal{L}^3_{\mbf f_{11},\mbf f_0,\mbf f_0} h_{11},
\nabla \mathcal{L}^3_{\mbf f_0,\mbf f_1,\mbf f_0} h_{11}, \\ &\quad
\nabla \mathcal{L}^3_{\mbf f_0,\mbf f_0,\mbf f_{11}} h_{11}
\}
\end{split}
\end{equation}
where $\mbf h_1 \triangleq [h_{11}, h_{12}]\tran$.
The 46-row spanning set involves Lie derivatives computed along vector fields of all control inputs, $\mbf a_m$ and $\mbs \omega_m$,
implying that weak observability of the system requires that all six axes of the IMU are unconstrained in the interval $\mathcal{I}$.

\subsection{Discussions on Extensions}
It is worth noting that the known gravity magnitude is needed for calibrating scale factors of the accelerometer.
If $h_2$ is unavailable, the above system of \eqref{eq:system-model} and \eqref{eq:oa-observations} is identifiable up to a metric scale with the unobservable direction
\begin{equation}
	\label{eq:unobservable-dim}
	\mbf N = \begin{bmatrix}
		\mbf{0}_{2}\tran & -\rho & {\mbf{v}^B}\tran & {\mbf{g}^B}\tran & \mbf{0}_{24}\tran & \vec{\mbf{M}}_a\tran & \mbf{p}_{CB}\tran & \mbf{0}_{12}\tran
	\end{bmatrix}\tran
\end{equation}
which satisfies $\mathcal{O}_k\mathbf{N} = \mathbf{0}$.

For the case of stereo cameras, if the IMU is unavailable,
it is easy to show that the intrinsic and extrinsic parameters of the stereo camera rig (of 22 DOFs) 
are observable given 4D observations of 21 landmarks and the baseline length.
Otherwise, for the stereo camera-IMU system, following the above analysis based on Lie derivatives,
it can be shown that the intrinsic and extrinsic parameters of both cameras and the IMU intrinsic parameters 
as listed in Table~\ref{tab:state-variables} are observable given 4D observations of an unknown landmark and the gravity magnitude or the baseline length.
Intuitively, one of the two latter constraints resolves the scale ambiguity.

When the system is standstill, whether at the beginning or middle of estimation, 
all state variables turn into unknown constants.
None of them can be resolved given the set of observation equations unless some parameters are assumed well estimated.
For instance, the IMU biases can be estimated if $\mathbf{M}_g$, $\mathbf{T}_s$, 
$\mathbf{M}_a$, and $\mbf g^W$ are roughly known.
Moreover, the system pose can be obtained given sufficient observations of landmarks of 
approximately known positions along with the camera intrinsic parameters.

\subsection{Observability of Time Parameters}
The camera time offset of the VIO system can be estimated unless special conditions arise.
These conditions are summarized by the below theorem proved in Appendix~\ref{app:time-offset}.
\begin{theorem}
\label{theorem}
For a nonlinear system affine in inputs, constant control inputs or observations make time offset weakly unobservable.
\end{theorem}
Looking from a distance, this statement agrees well with intuition.
Moreover, when we transform the control inputs (by using variables in the state $\mbf x$)
into a new set of control inputs and rewrite the original system with the new inputs,
the time offset is unobservable if the new inputs are constant.

Thus, for VIO, the conditions to make $t_d$ weakly unobservable is
$\mbf a_m = \mathrm{const}$ and 
$\mbs{\omega}_m = \mathrm{const}$, or $\mbf h = \mathrm{const}$.
Since the VIO system can be written in terms of control inputs transformed with the state variables, for instance, $\mbf a_t, \mbs \omega_t$,
the time offset is also unobservable when the alternative control inputs are constant.
In \cite{liOnlineTemporalCalibration2014}, the time offset is identified to be unobservable under constant local angular rate.
Later, the time offset is verified to be unobservable in \cite{yangDegenerateMotionAnalysis2019} with constant local angular rate and constant local velocity or global acceleration.
Our condition encompasses these above conditions, and is stronger and more generic.

As for the RS effect, we have the following corollary proved similarly to Theorem~\ref{theorem} in Appendix~\ref{app:readout-time}.
\begin{corollary}
\label{corollary}
The frame readout time is generally observable except when the observations are trivially constant.
\end{corollary}

In summary, the state in the VIO system \eqref{eq:system-model}, camera time offset, and readout time can be estimated under general motion,
meaning that KSWF with self-calibration is feasible in theory.

\section{Simulation Study}
\label{sec:simulation}
By simulation in two scenarios, this section shows that the presented sensor parameters can be estimated with observations of opportunistic point landmarks, 
and that the proposed KSWF can accurately estimate platform motion with inaccurately calibrated sensors.

\begin{figure}[!htb]
	\centering
	\includegraphics[width=0.8\columnwidth]{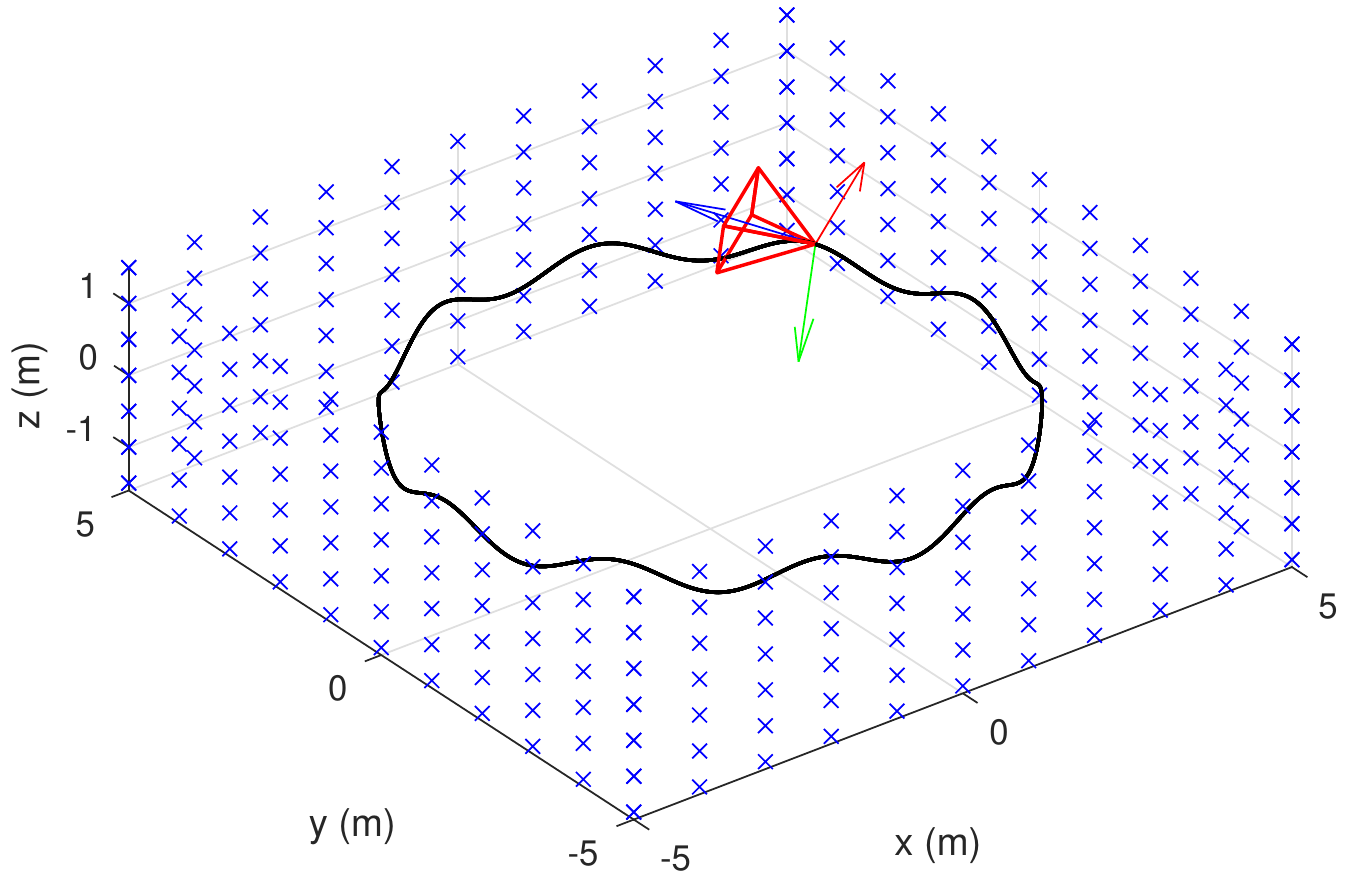}
	\includegraphics[width=0.95\columnwidth]{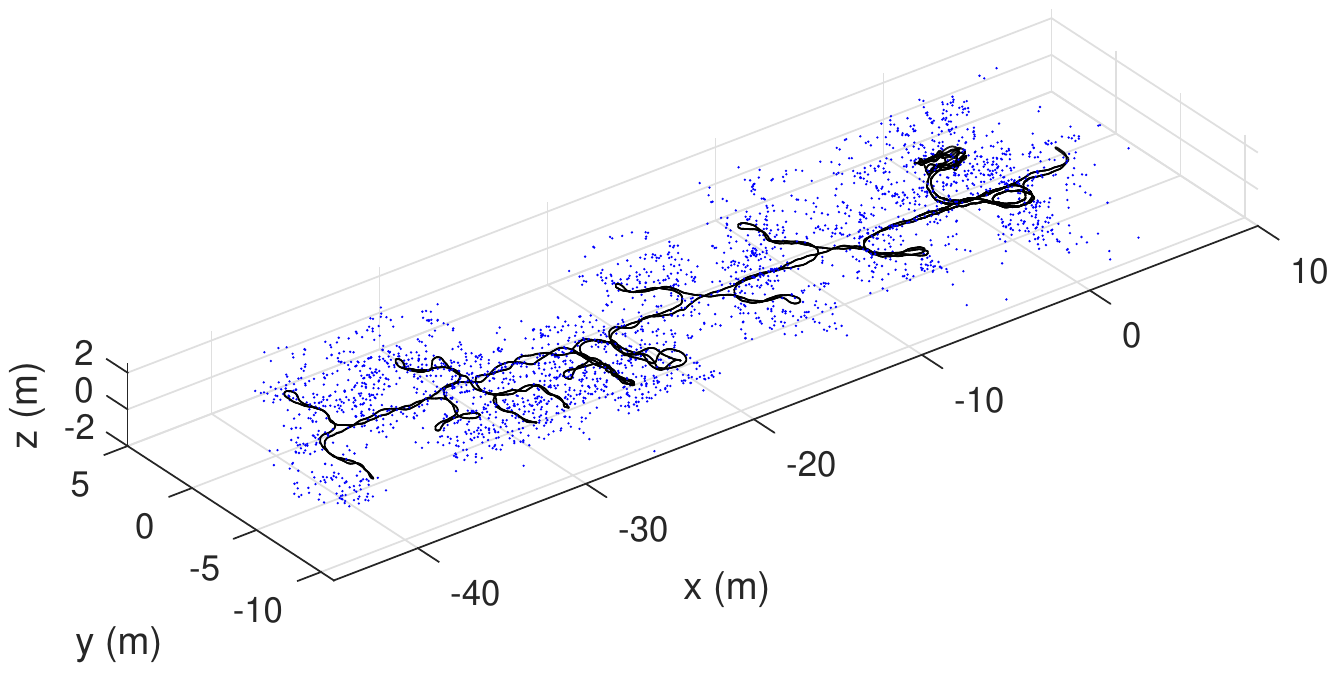}
	\caption{Simulated setups for the monocular RS camera-IMU system. 
		Top: Landmarks on walls and a general wavy circle trajectory. 
		Bottom: Scene and trajectory generated from the TUM VI corridor3 sequence \cite{schubertTUMVIBenchmark2018}.}
	\label{fig:scenarios}
\end{figure}

The first scenario was setup with a monocular RS camera-IMU system and an artificial scene of points distributed on four walls.
The camera-IMU platform traversed the scene for five minutes with a wavy circle motion (Fig.~\ref{fig:scenarios} top) at average velocity 1.26 $m/s$.
For the second scenario, both the trajectory of the monocular RS camera-IMU system and the landmarks 
were generated from the TUM VI corridor3 sequence \cite{schubertTUMVIBenchmark2018}. Specifically, the sequence was first processed by the KSWF to obtain a trajectory estimated at 200 Hz, and then by maplab \cite{schneiderMaplab2018} with loop closure to obtain a trajectory of keyframes at about 10 Hz and 2700 landmarks. Finally, the discrete reference trajectory was obtained by optimizing the KSWF trajectory with these keyframe poses as constraints. This procedure is needed because the TUM VI ground truth only covers the start and the end of the sequence.
In simulation, the continuous-time reference trajectory was represented by a $R^3$ spline function and a cumulative $SO(3)$ spline function composed of fifth-order B-splines \cite{sommerContinuoustime2016}, fitted to the discrete pose samples, as shown in Fig.~\ref{fig:scenarios} bottom.
Both setups used the same camera of the diagonal FOV 103$^\circ$ and a resolution $752 \times 480$.
For the wavy circle, the camera operated at 10 Hz and the IMU at 100 Hz.
For the corridor3, the camera operated at 20 Hz and the IMU at 200 Hz.
Image observations were obtained with the pinhole radial tangential model and affected by
white Gaussian noise with standard deviation of 1 pixel in each dimension.
The RS effect was simulated by solving for the feature observation time \eqref{eq:feature_time} as 
the only root of the projection function~\eqref{eq:reprojection}.
IMU data were affected by biases, bias random walk, and additive white noise.
IMU noise parameters were chosen to be realistic for a consumer-grade IMU and are given in Table~\ref{tab:imu_noise}.
Sensor parameters used to simulate images and inertial data were set to
mean values in Table~\ref{tab:init_value_std}.

A simulation frontend was created to associate landmark observations between consecutive
frames and between a frame and its previous two keyframes.
To handle cluttered observations in an image, a grid of cell size 32$\times$24 is used to keep only the keypoint for the closest landmark observed in each cell.
The average feature track length is 19.9 for the wavy circle scenario and 8.7 for the corridor3 scenario, and the average number of observed landmarks in an image is 63 for the former and 142 for the latter.

\input{tables/compared-methods.tex}

To study the effect of self-calibration,
four settings of KSWF with varying degrees of self-calibration were chosen (see Table~\ref{tab:features-odometry-methods}):
only calibrating IMU biases and camera extrinsic parameters (\ie, KSWF min cal.), additionally calibrating IMU intrinsic parameters (\ie, KSWF cal. IMU),
additionally calibrating camera intrinsic parameters and $t_d$ and $t_r$ (\ie, KSWF cal. camera), and the default full calibration (\ie, KSWF).
In all four settings, both IMU biases and camera extrinsic parameters were calibrated on the fly, 
as is attainable with most VIO methods, \eg, OKVIS\cite{leuteneggerKeyframebased2015}, ROVIO\cite{bloeschIteratedExtendedKalman2017}, and VINS-Mono \cite{qinVINSMono2018}.
To evaluate motion estimation, comparison was made to OKVIS \cite{leuteneggerKeyframebased2015} and a structureless KSWF (SL-KSWF) which uses observations for
update as a feature track disappears and does not add landmarks to the state vector.
Note that for OKVIS, RS effect and camera time offset were not added in simulating data.

\begin{table}[!htb]
\centering
\caption{IMU noise parameters.}
\begin{tabular}{llll}
	\toprule
	$\sigma_{bg}$ & $\sigma_{ba}$ & $\sigma_g$ & $\sigma_a$ \\ \midrule
	$2\cdot10^{-5}$ & $5.5\cdot10^{-5}$ &	$1.2\cdot10^{-3}$ & $8\cdot10^{-3}$ \\ \midrule
	 $rad/s^2/\sqrt{Hz}$ & $m/s^3/\sqrt{Hz}$ & $rad/s/\sqrt{Hz}$ & $m/s^2/\sqrt{Hz}$ \\
	 \bottomrule
\end{tabular}
\label{tab:imu_noise}
\end{table}

\input{tables/init-value-std.tex}

The compared estimators were initialized with true pose and velocity
affected by Gaussian noise 
$N(\mbf{0}, 0.05^2 \mbf{I}_3 \hspace{0.2em} m^2/s^4)$.
For KSWF and its variants, the initial values for sensor parameters were drawn
from Gaussian distributions given in Table~\ref{tab:init_value_std}.
Except SL-KSWF, a landmark was added to the state vector if it had been observed six times.
OKVIS was initialized with either sensor models used in simulation or inaccurate sensor models.
For the former (\ie, OKVIS calibrated), sensor parameters were initialized with
mean values in Table~\ref{tab:init_value_std} except time offset
and readout time which were set to zero.
For the latter (\ie, OKVIS uncalibrated), sensor parameters were corrupted with noises in Table~\ref{tab:init_value_std}.

\input{tables/sim-rmse.tex}

The simulation and estimation were repeated 100 times for an estimator (see also the supplementary video).
A run is considered successful if the final position error is less than 100 m.
The pose RMSE values at the end were computed over successful runs and are reported in Table~\ref{tab:sim-rmse}
for the wavy circle and corridor3.
With inaccurate sensor parameters, both KSWF and SL-KSWF with full self-calibration 
achieved better or comparable accuracy to OKVIS with good calibration, largely outperforming OKVIS starting from inaccurate parameters.
One point worth mentioning is that though SL-KSWF showed good accuracy, it was fairly likely to diverge starting from poor calibration.
Referring to Table~\ref{tab:sim-rmse}, additionally calibrating mere IMU or camera parameters, the accuracy suffered much compared to full self-calibration.
As fewer parameters were estimated on-the-fly, the KSWF diverged more often given inaccurate parameters.
These results validate that full self-calibration benefits motion estimation when the prior calibration is inaccurate.

\begin{figure*}[!htb]
\centering
\includegraphics[width=0.32\linewidth]{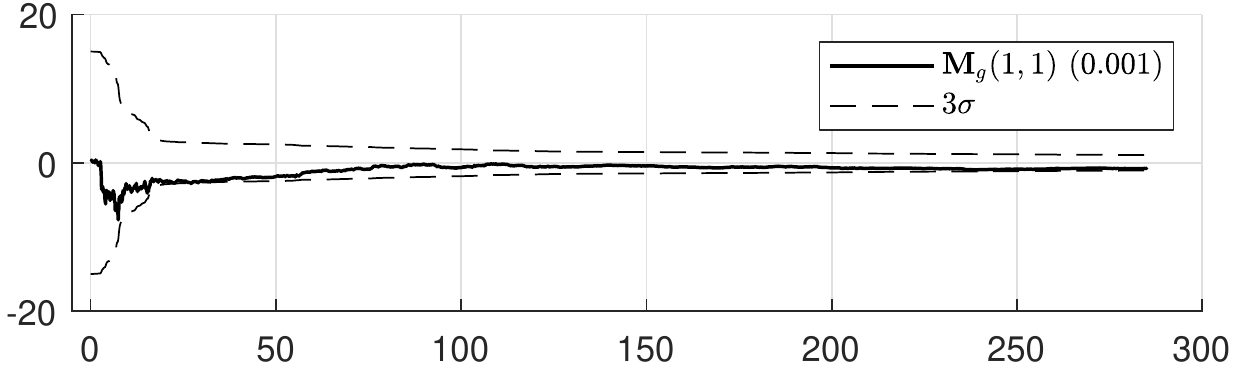}
\includegraphics[width=0.32\linewidth]{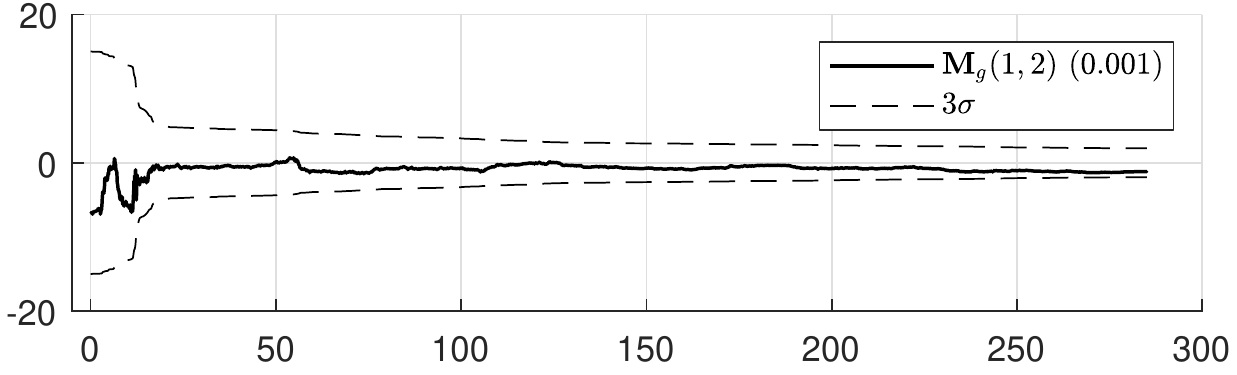}
\includegraphics[width=0.32\linewidth]{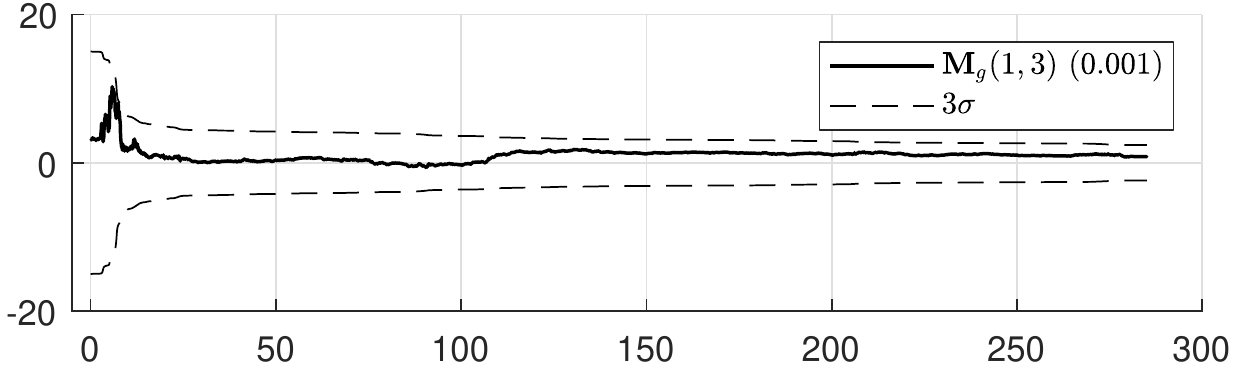}
\includegraphics[width=0.32\linewidth]{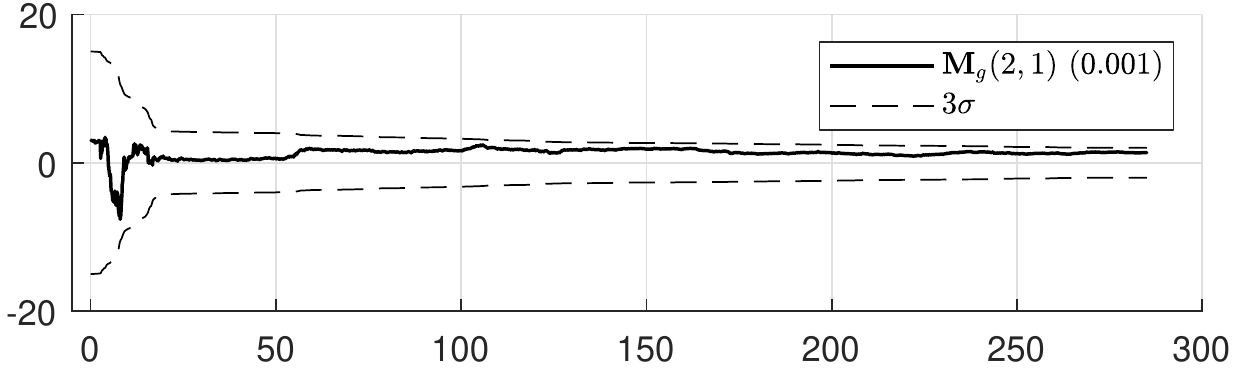}
\includegraphics[width=0.32\linewidth]{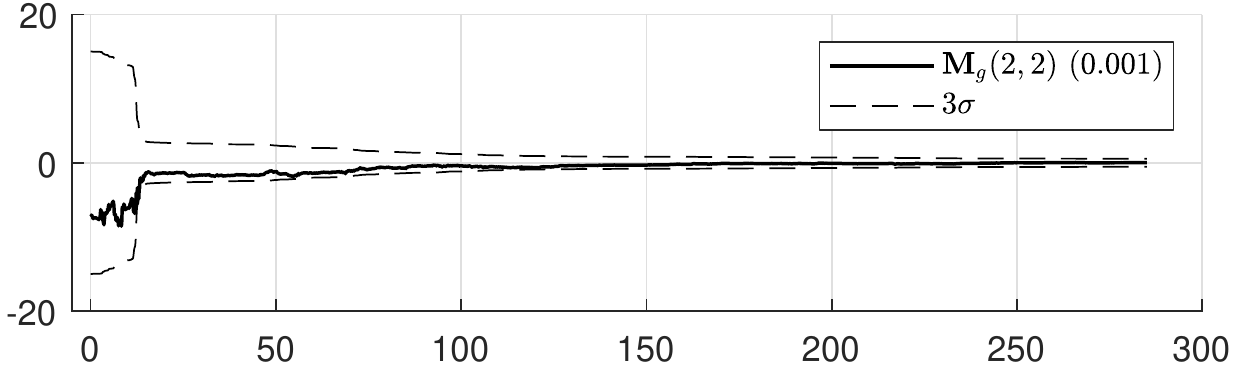}
\includegraphics[width=0.32\linewidth]{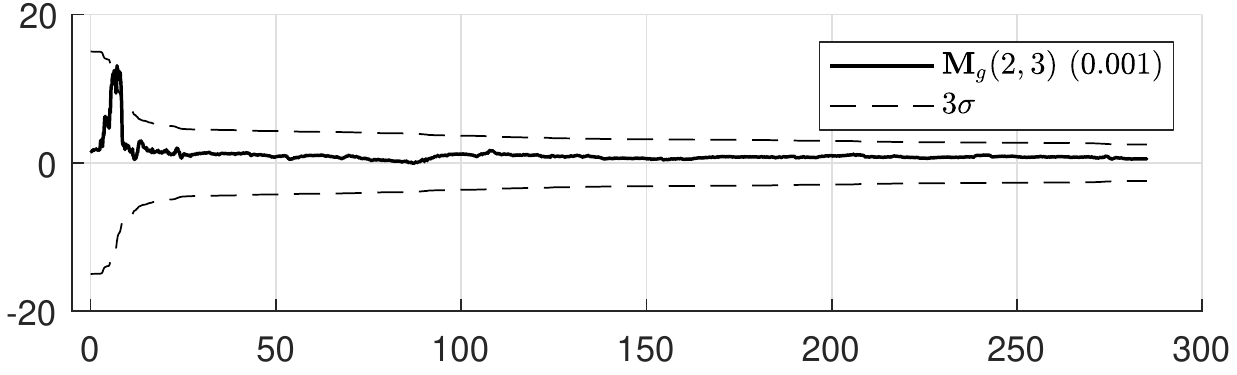}
\includegraphics[width=0.32\linewidth]{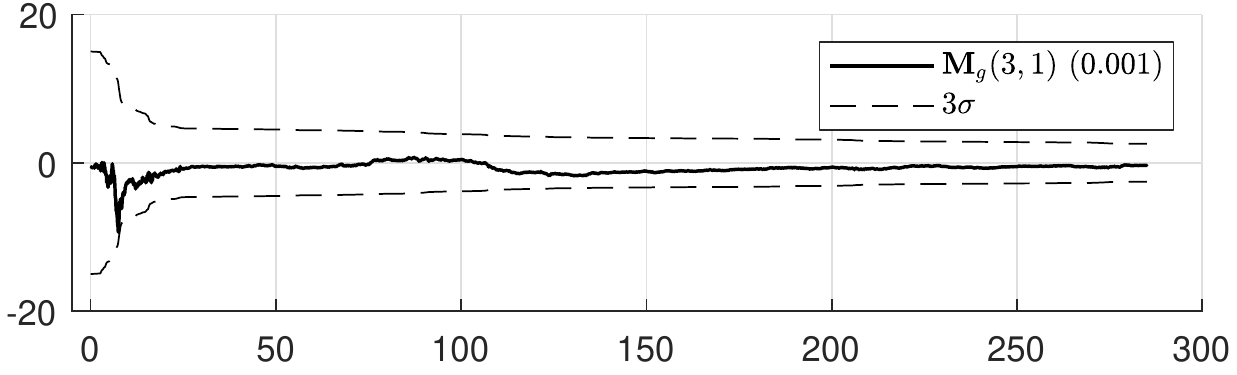}
\includegraphics[width=0.32\linewidth]{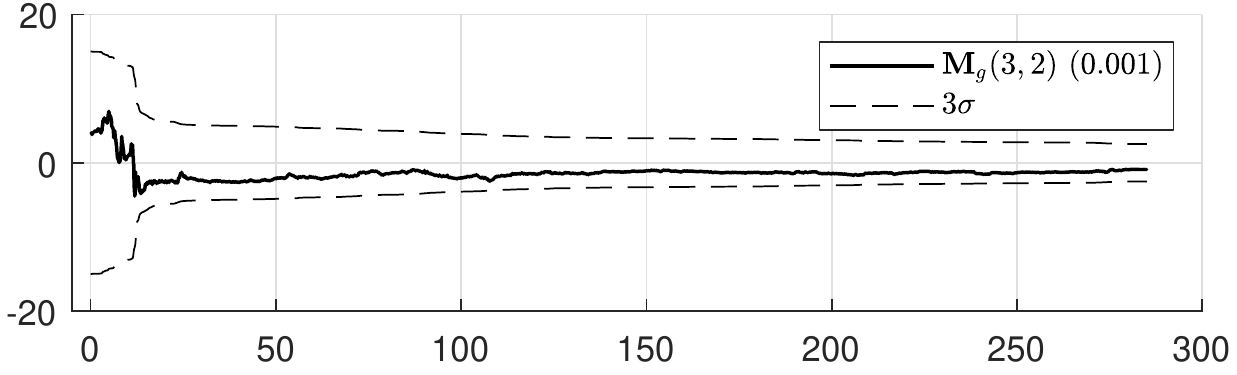}
\includegraphics[width=0.32\linewidth]{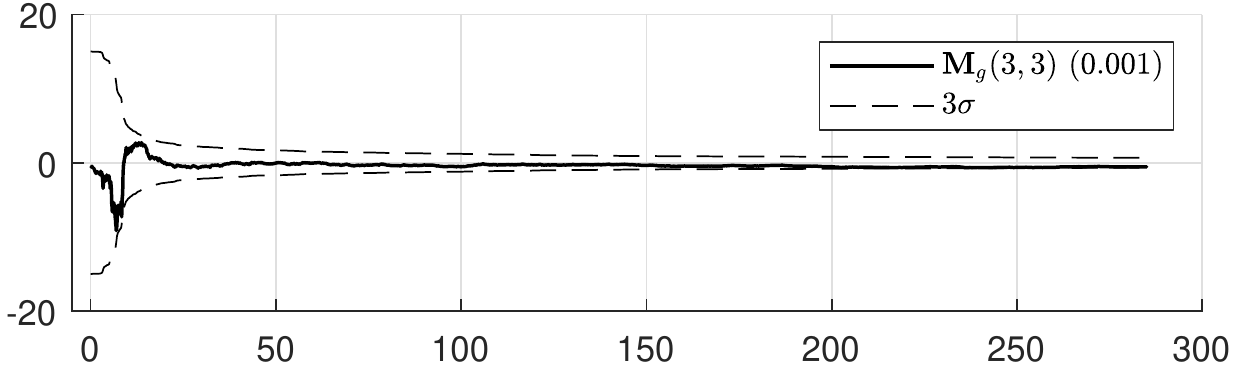}
\includegraphics[width=0.32\linewidth]{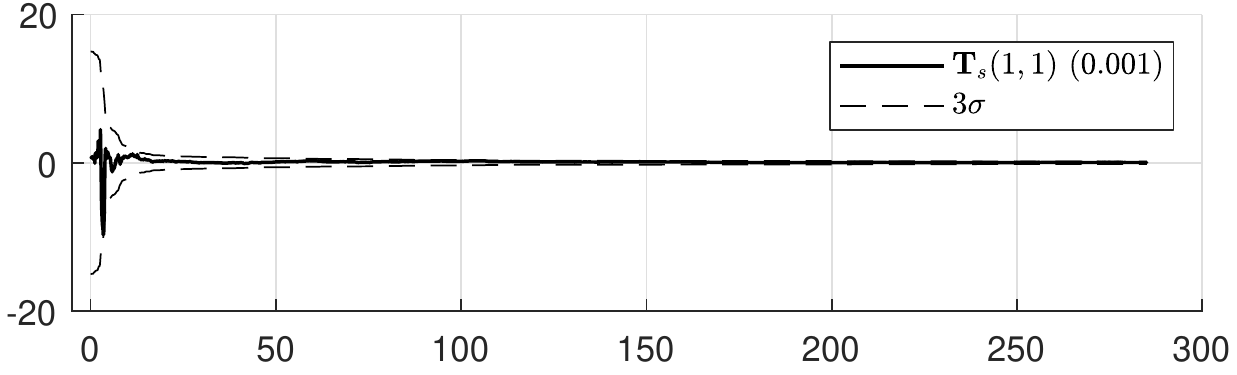}
\includegraphics[width=0.32\linewidth]{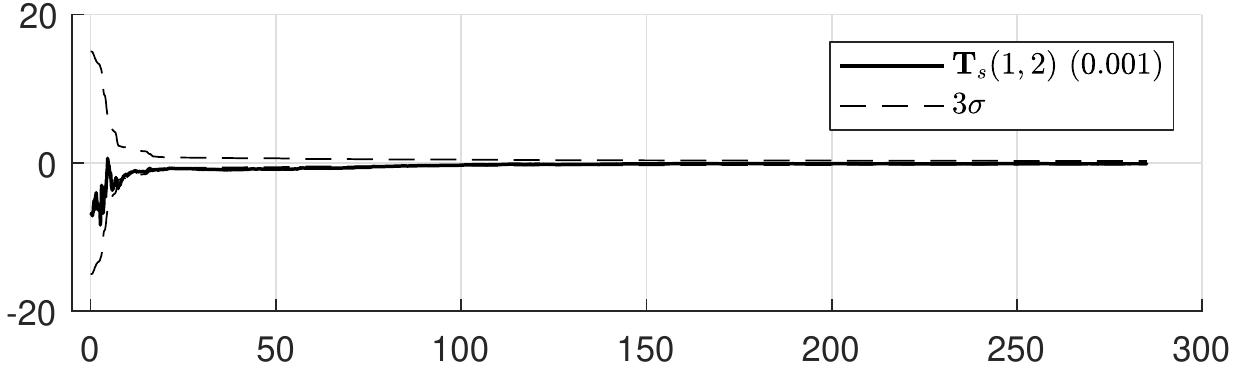}
\includegraphics[width=0.32\linewidth]{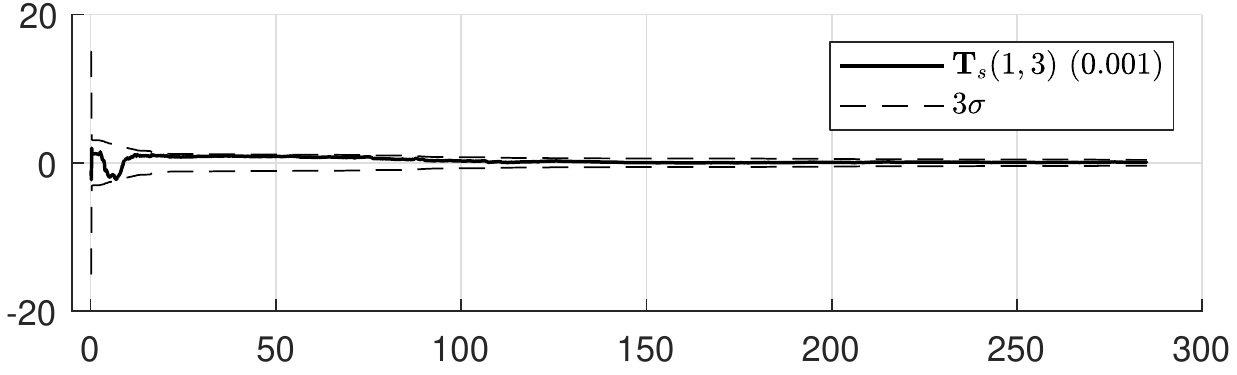}
\includegraphics[width=0.32\linewidth]{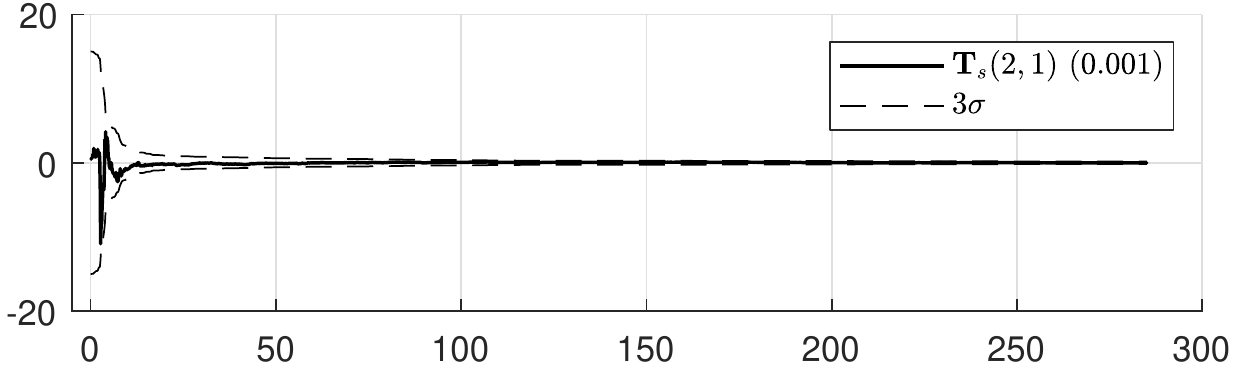}
\includegraphics[width=0.32\linewidth]{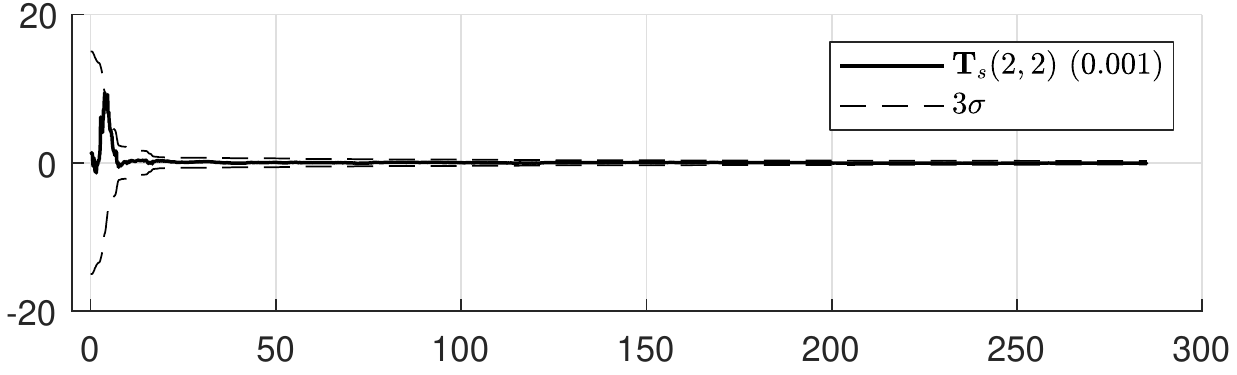}
\includegraphics[width=0.32\linewidth]{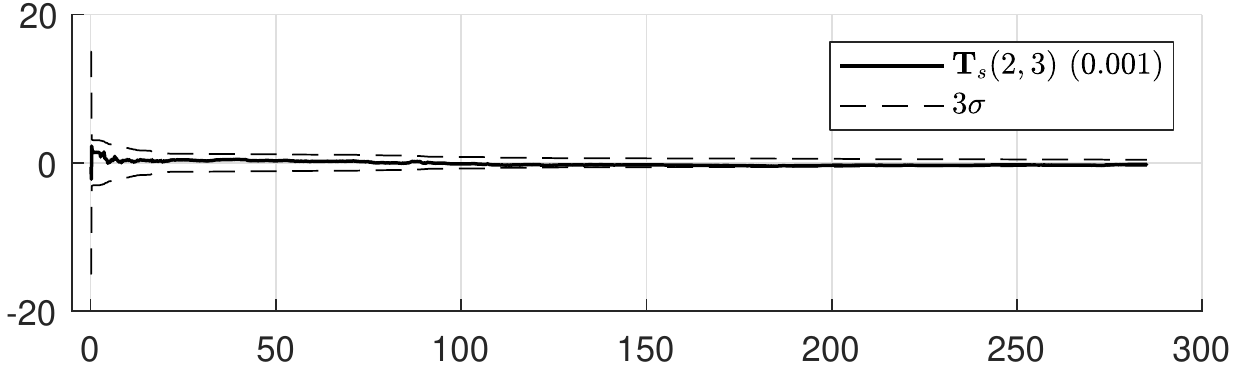}
\includegraphics[width=0.32\linewidth]{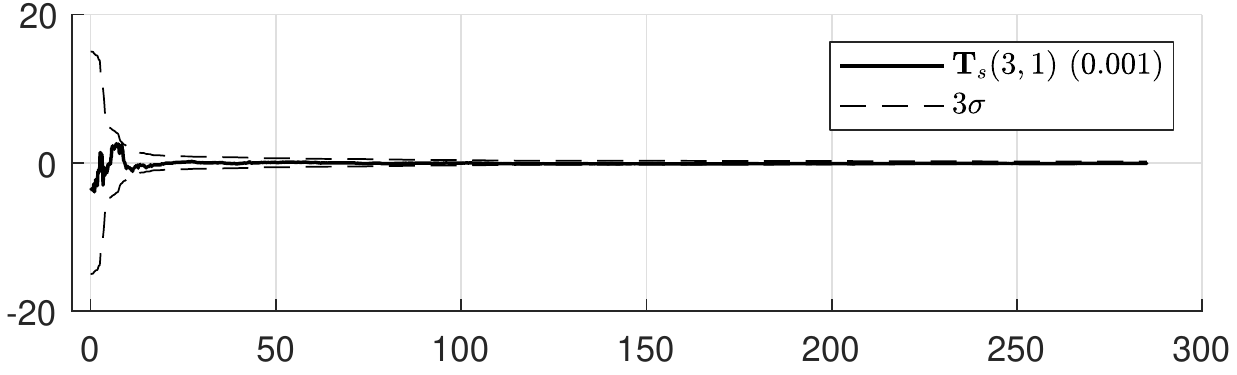}
\includegraphics[width=0.32\linewidth]{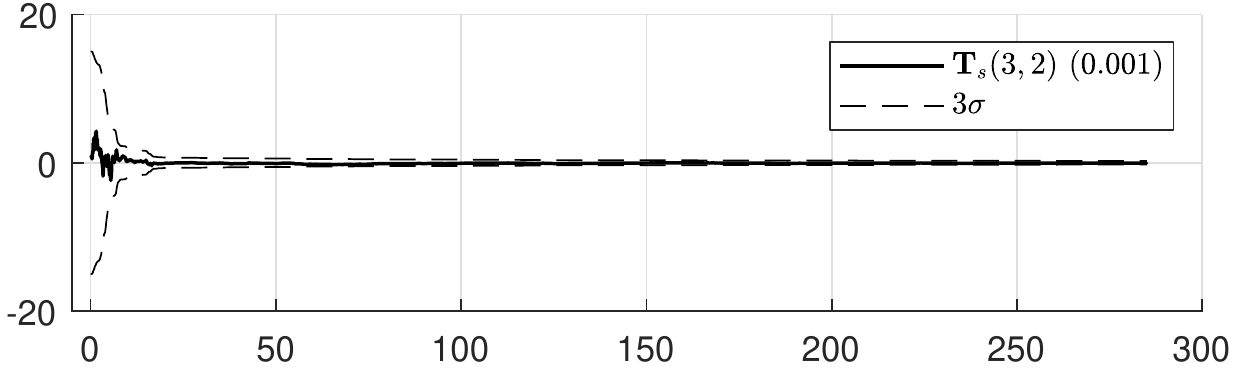}
\includegraphics[width=0.32\linewidth]{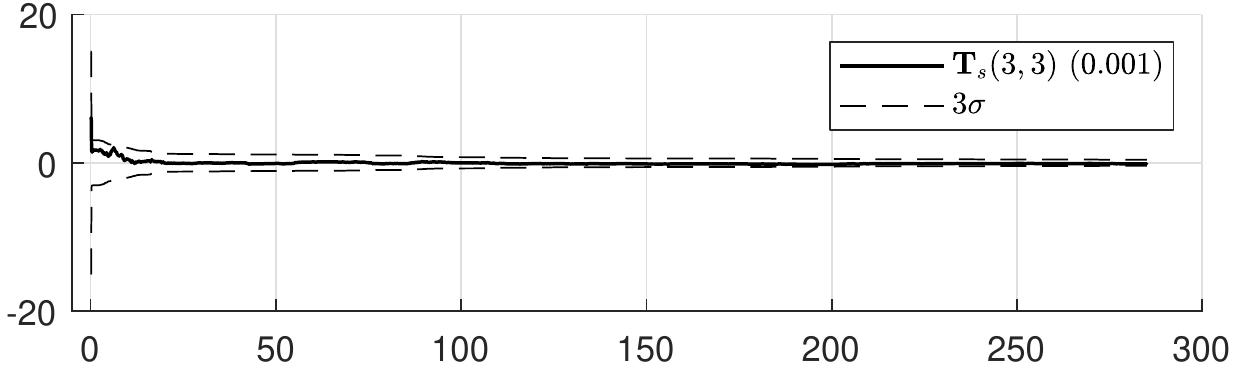}
\includegraphics[width=0.32\linewidth]{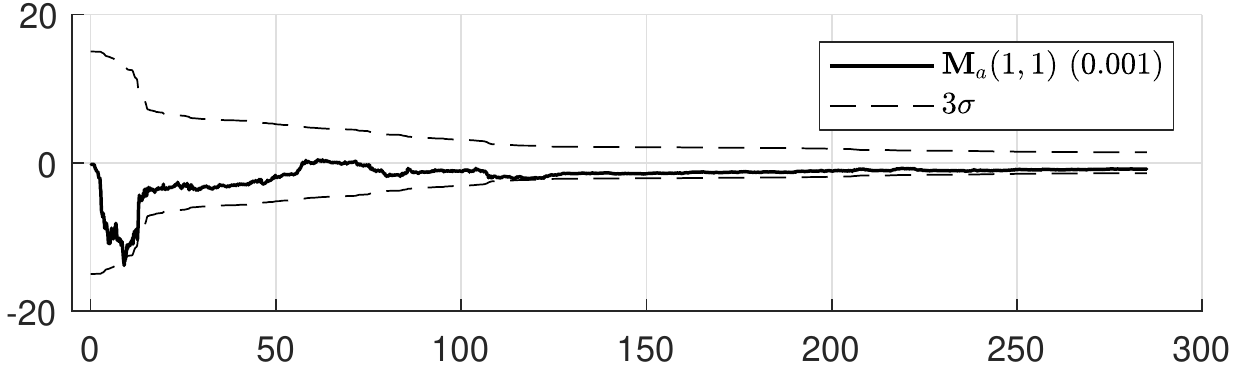}
\includegraphics[width=0.32\linewidth]{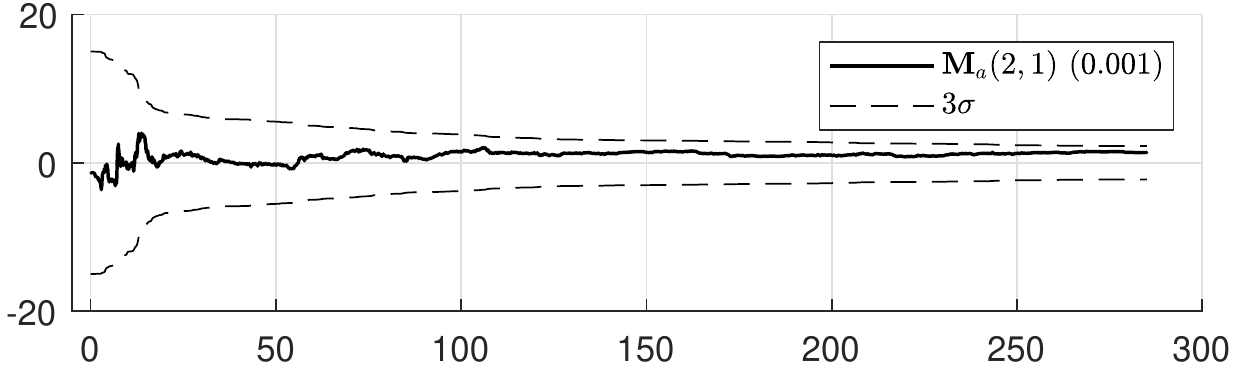}
\includegraphics[width=0.32\linewidth]{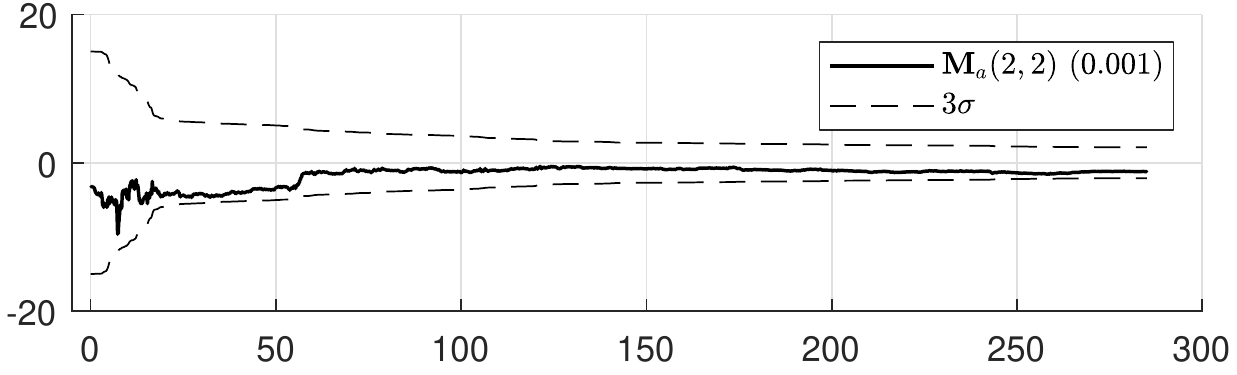}
\includegraphics[width=0.32\linewidth]{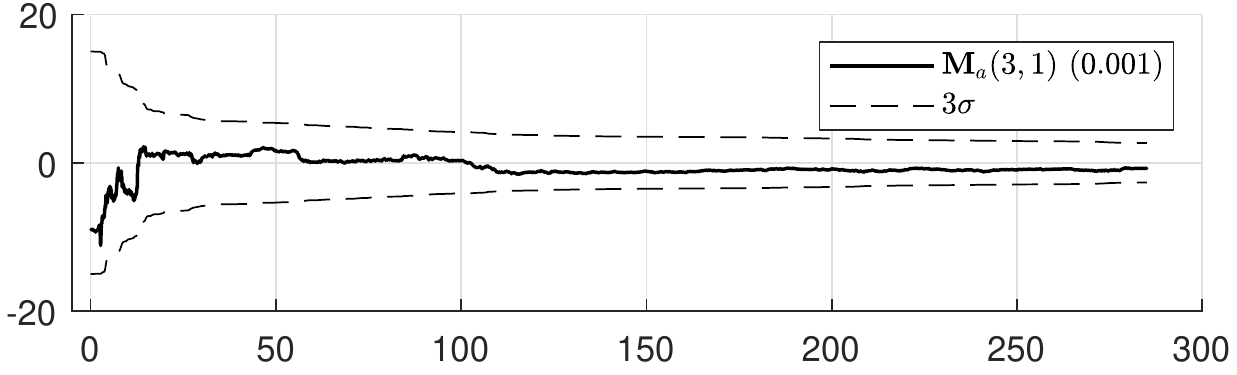}
\includegraphics[width=0.32\linewidth]{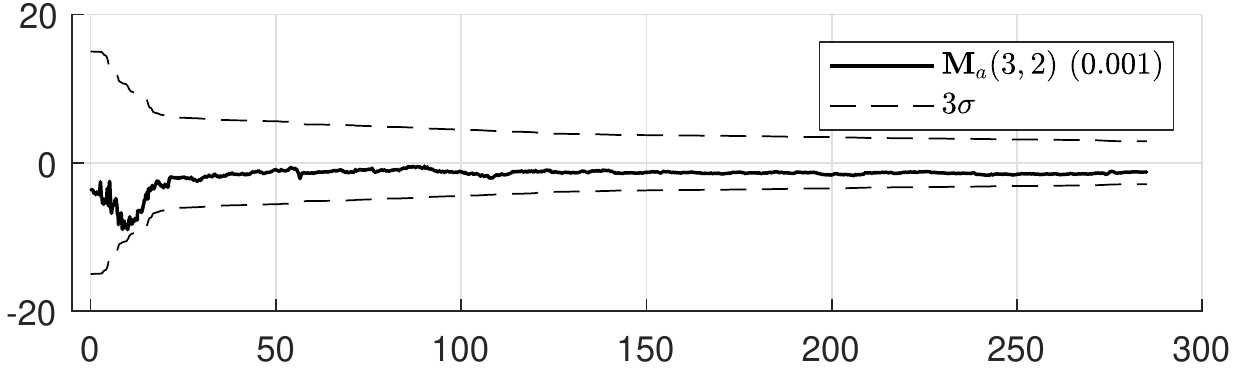}
\includegraphics[width=0.32\linewidth]{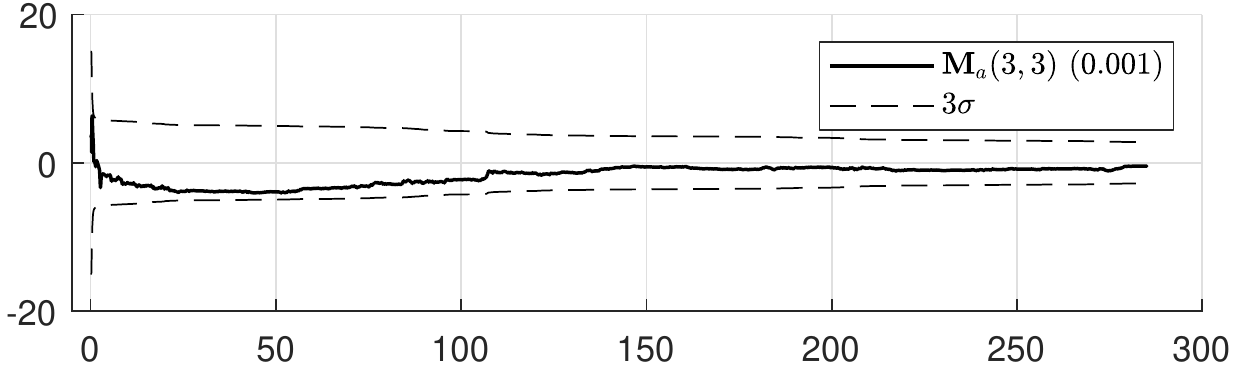}
\includegraphics[width=0.32\linewidth]{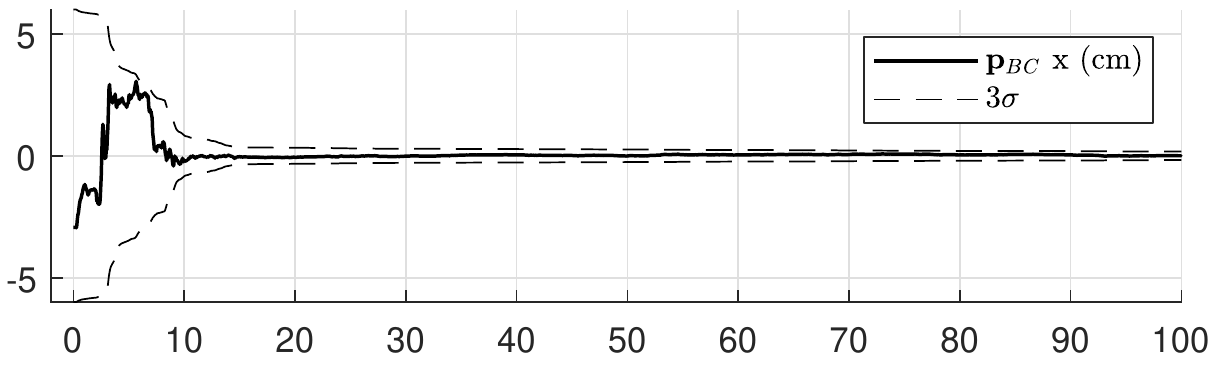}
\includegraphics[width=0.32\linewidth]{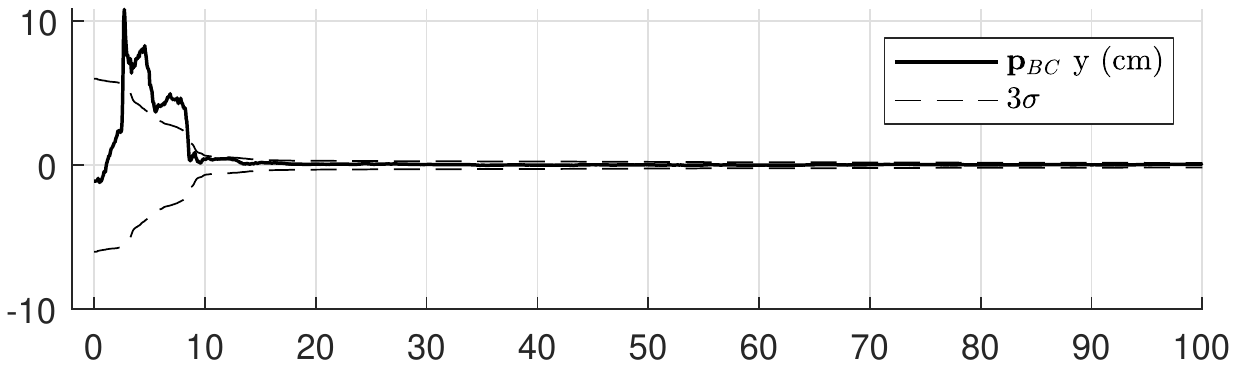}
\includegraphics[width=0.32\linewidth]{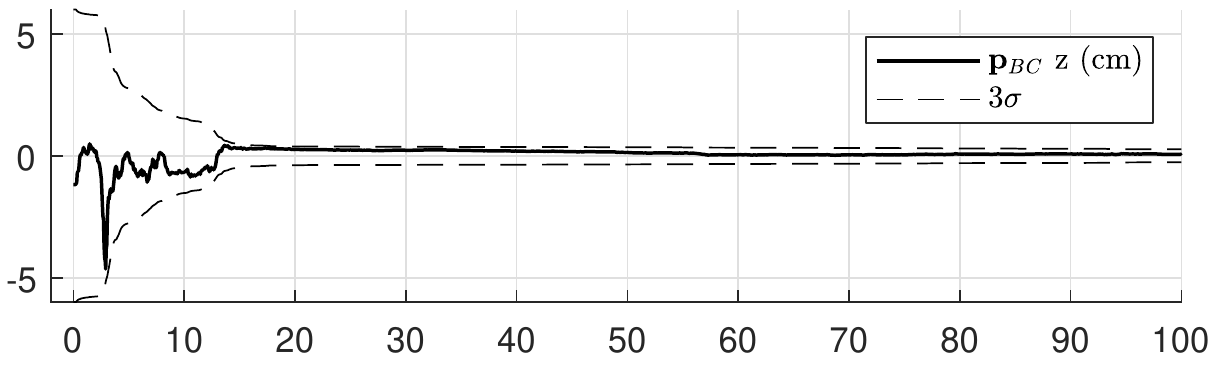}
\includegraphics[width=0.32\linewidth]{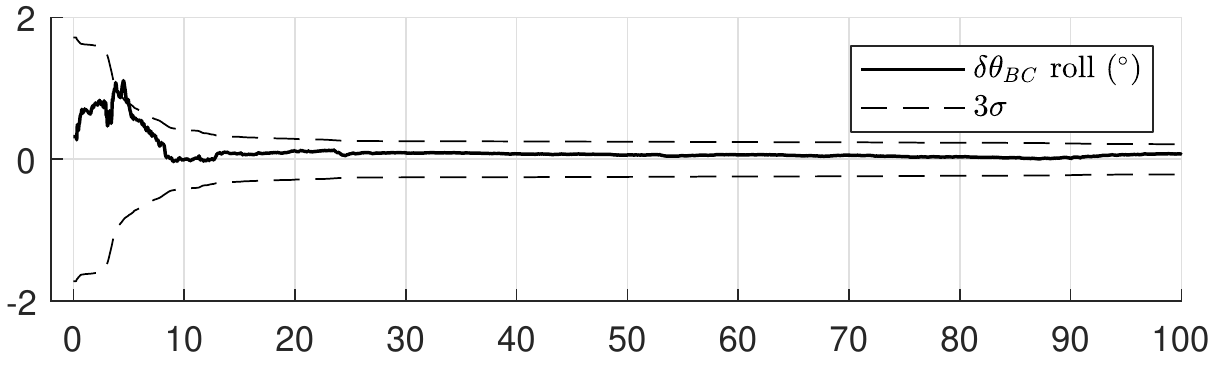}
\includegraphics[width=0.32\linewidth]{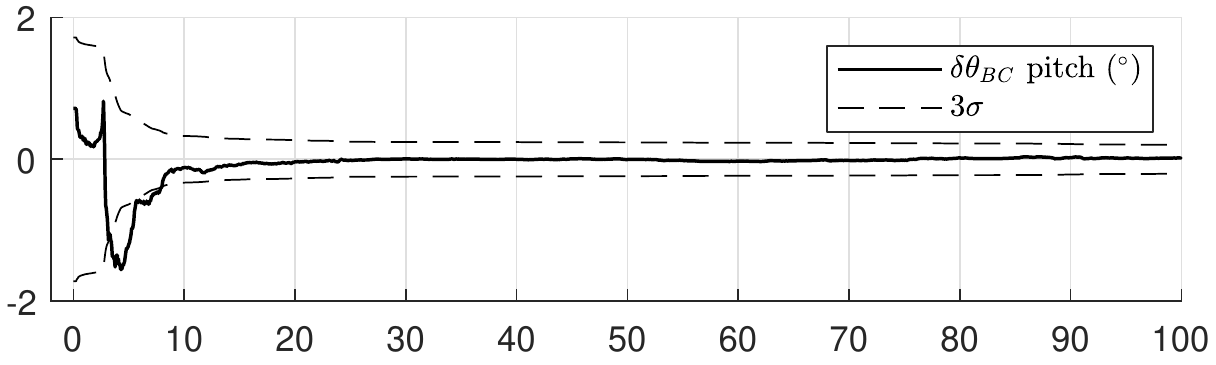}
\includegraphics[width=0.32\linewidth]{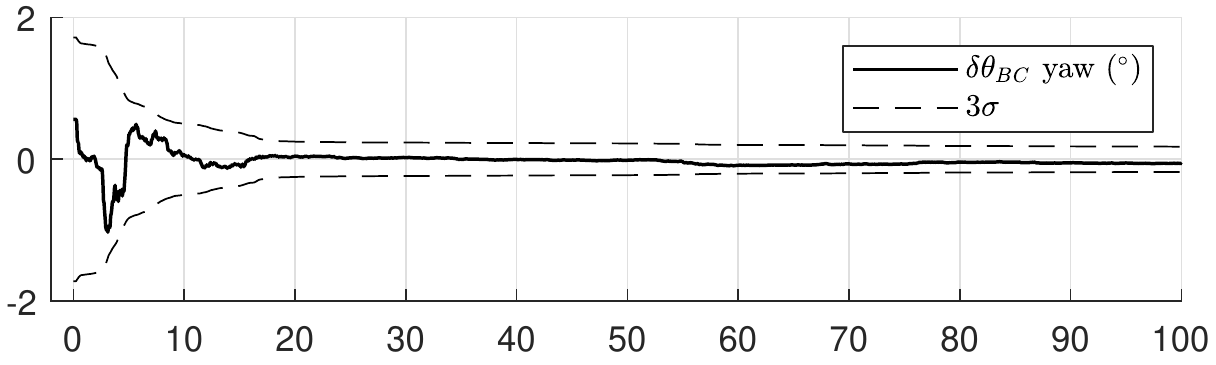}
\includegraphics[width=0.32\linewidth]{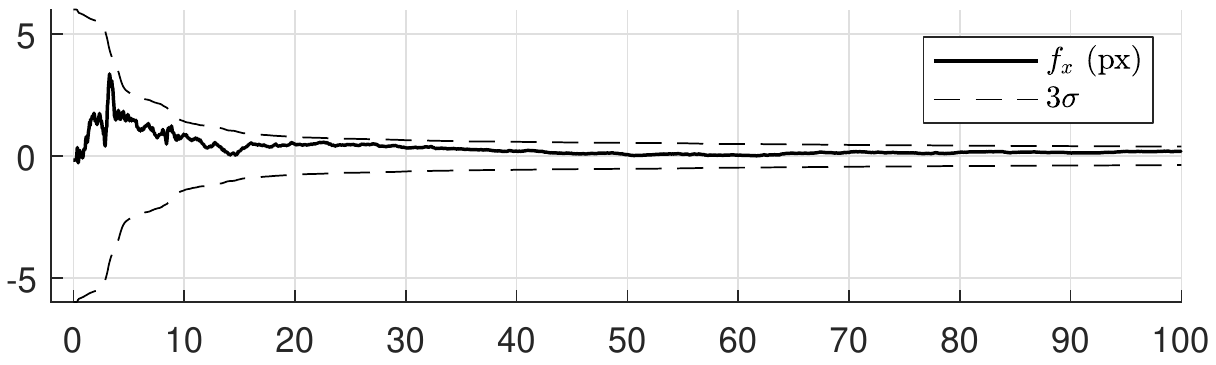}
\includegraphics[width=0.32\linewidth]{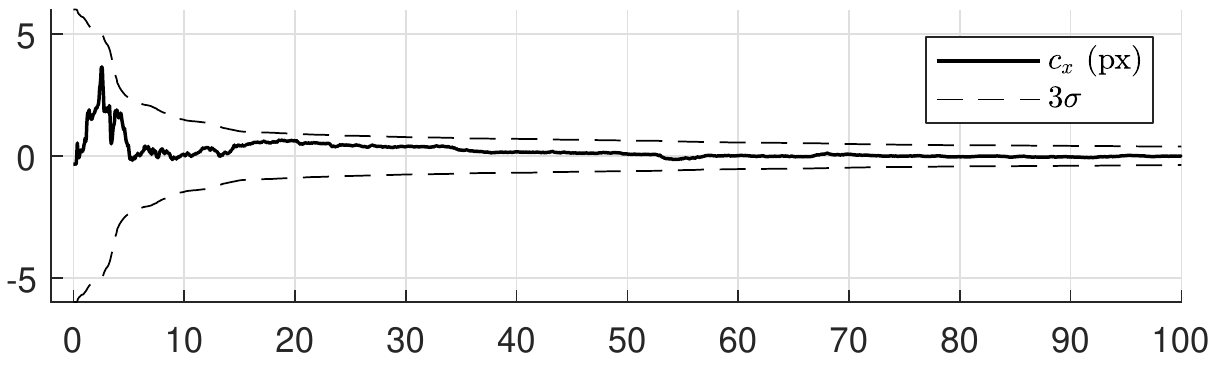}
\includegraphics[width=0.32\linewidth]{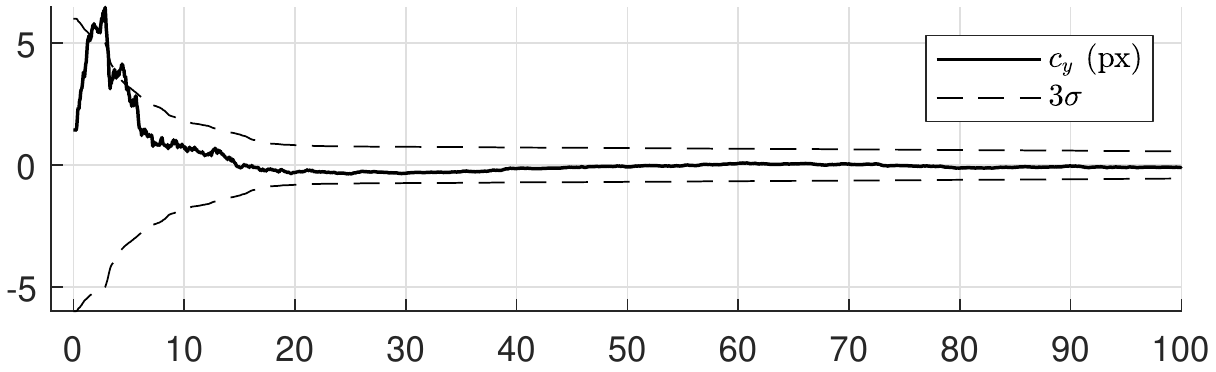}
\includegraphics[width=0.32\linewidth]{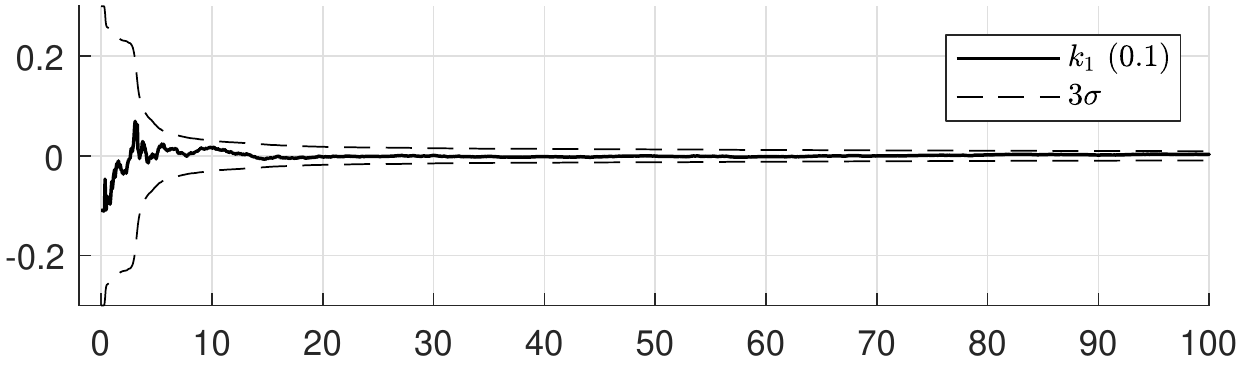}
\includegraphics[width=0.32\linewidth]{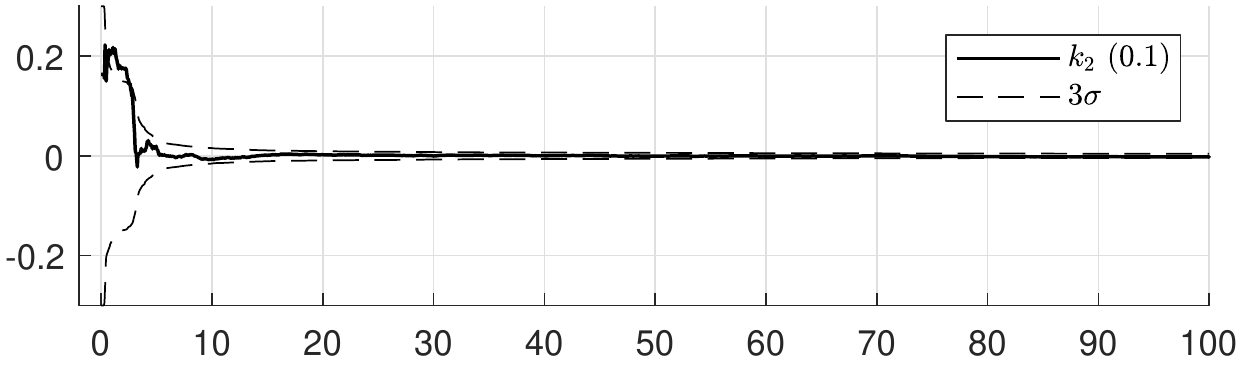}
\includegraphics[width=0.32\linewidth]{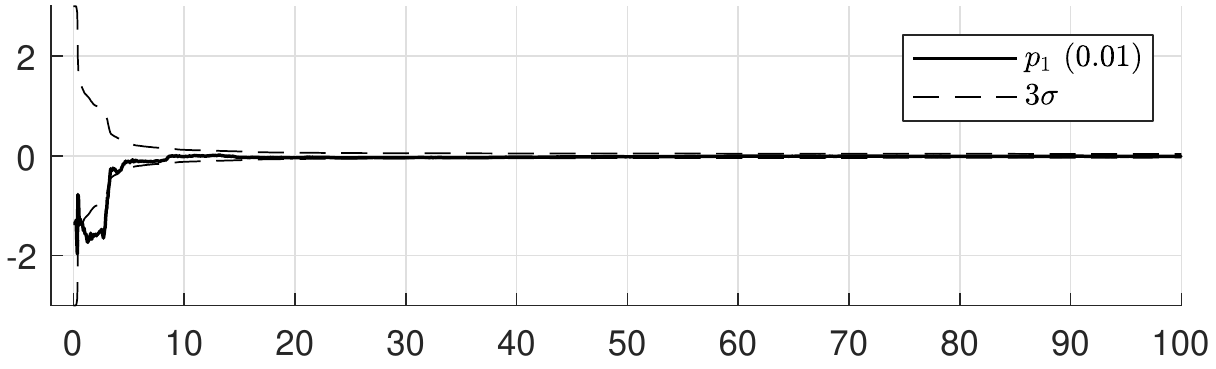}
\includegraphics[width=0.32\linewidth]{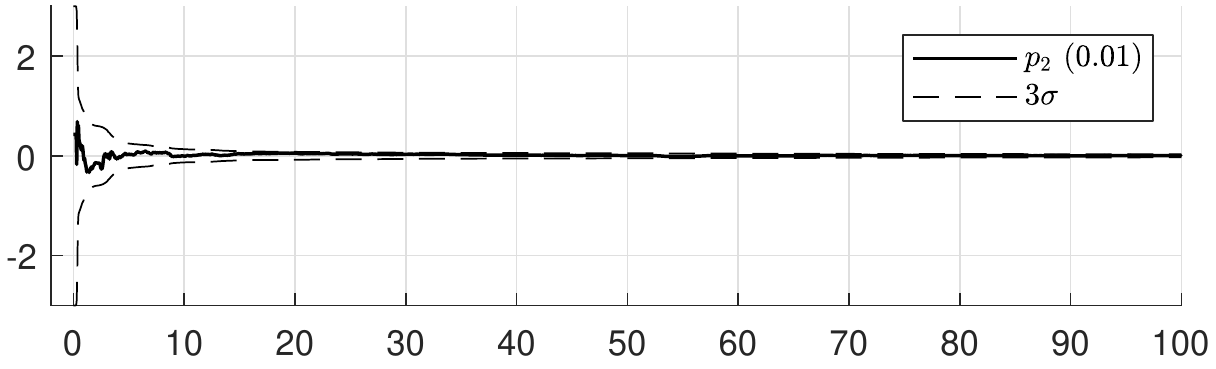}
\includegraphics[width=0.32\linewidth]{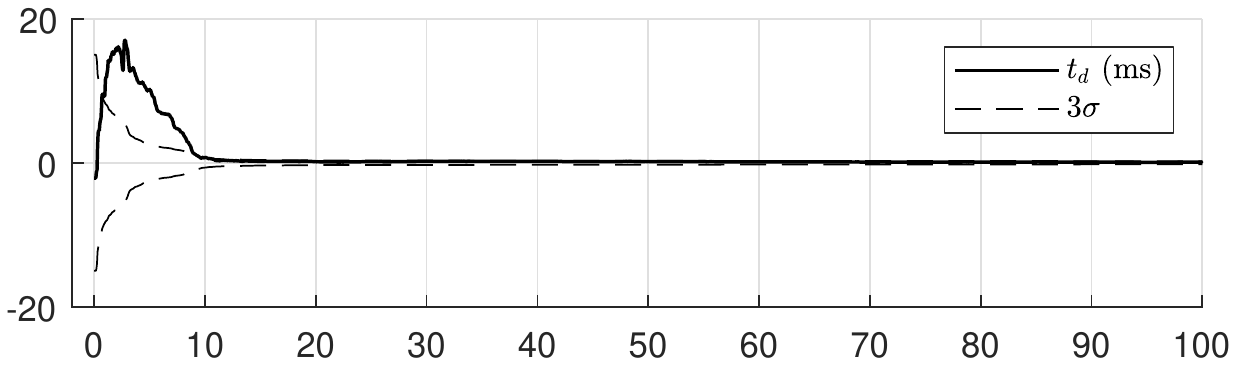}
\includegraphics[width=0.32\linewidth]{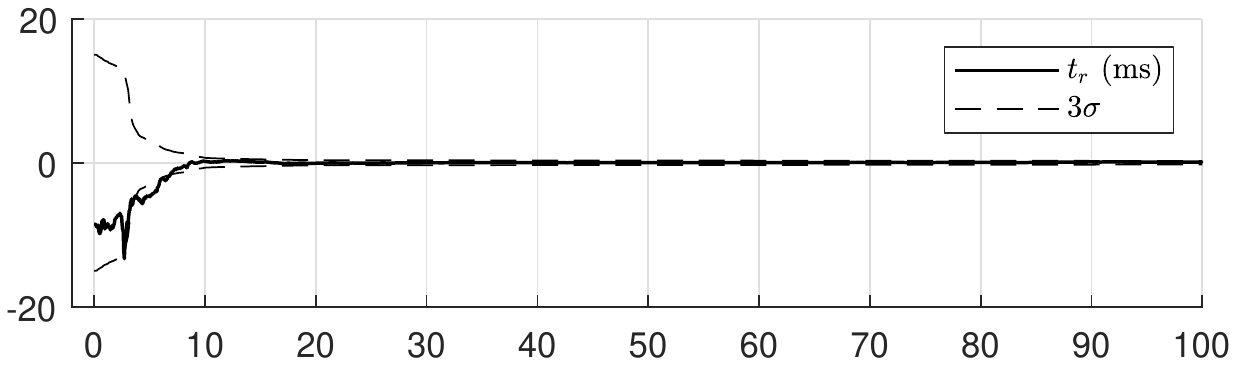}
\caption{Errors (solid lines) and $3\sigma$ bounds (dashed lines) of estimated camera-IMU system parameters in a sample run of KSWF on the data simulated from TUM VI corridor3 sequence.
The elements of $\mbf T_s$ are in metric units.
The $x$ axis represents time in seconds.
}
\label{fig:sim-calib-param-curves}
\end{figure*}

With the corridor3, we reveal evolution of the estimated sensor parameters.
For a sample run, errors in the calibrated parameters and corresponding $3\sigma$ bounds over time are drawn in Fig.~\ref{fig:sim-calib-param-curves}.
All components decreased in uncertainty over time though at varying rates.
The decreasing errors and tightening $3\sigma$ bounds validate that these parameters can be estimated reliably, corroborating analysis presented in Section~\ref{sec:observability}.

In summary, the simulation results show that KSWF can accurately estimate motion and fully calibrate the camera-IMU system with general motion.

\section{Real-World Experiments}
\label{sec:real-world-tests}
This section presents experiments on real data
to validate advantages of keyframe-based filtering and full self-calibration.
We report motion estimation with standstill periods first, and then  
motion estimation with full self-calibration (see also the supplementary video).
A previous evaluation of the SL-KSWF has been reported in \cite{huaiVersatile2020}.

To quantify estimation performance, pose accuracy was measured with Root Mean Square (RMS) of the Absolute Trajectory Error (ATE)
consisting of rotational RMSE, and translational RMSE \cite{zhangTutorial2018}.
To compute the two metrics, the trajectory estimated by a VIO method was aligned to the
ground truth with a transformation of a yaw-only rotation and a 3D translation.

In general, the compared methods were configured with as many default parameters as possible,
and they processed data synchronously without enforcing real time execution unless this was not possible. 
To reduce randomness, results from five runs were used to compute the error metrics.

Throughout tests for KSWF, we used $N_{kf} = 5$ and $N_{tf} = 5$. 
A landmark was added to the state vector if it had been observed six times, 
and would be marginalized if it eluded three consecutive frame bundles.
At maximum, 50 landmarks were kept in the state vector.

\subsection{Motion Estimation with Standstills}
\label{subsec:motion-stationary}
We examined the drift issue of a monocular structureless filter in standstill by comparing several sliding window filters.
The first four were KSWF, the non-keyframe version of KSWF (SWF) where every frame was taken as a keyframe and feature matching was only done between consecutive frames,
and their structureless counterparts, SL-KSWF and SL-SWF, respectively.
For the four filters, the self-calibration capability was turned off for clear comparison.
Additionally, we also compared with ROVIO \cite{bloeschIteratedExtendedKalman2017} which does not use a sliding window but keeps landmarks in the state vector and 
MSCKF-Mono \cite{zhuEventbasedVisualInertial2017} which uses frame-wise feature matching and keyframes in the backend filtering.
In essence, the SL-SWF is very close to MSCKF-Mono since both are structureless and use frame-wise feature matching.

\begin{figure}[!htb]
	\centering
	\includegraphics[width=0.8\columnwidth]{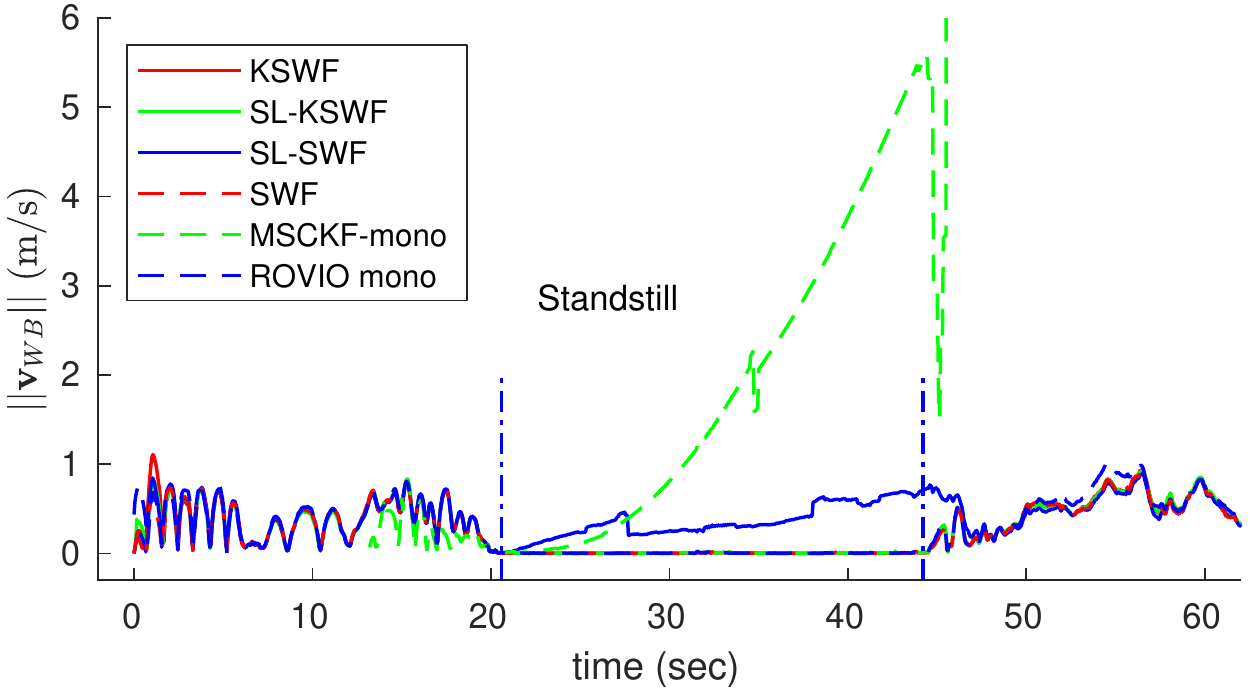}
	\includegraphics[width=0.8\columnwidth]{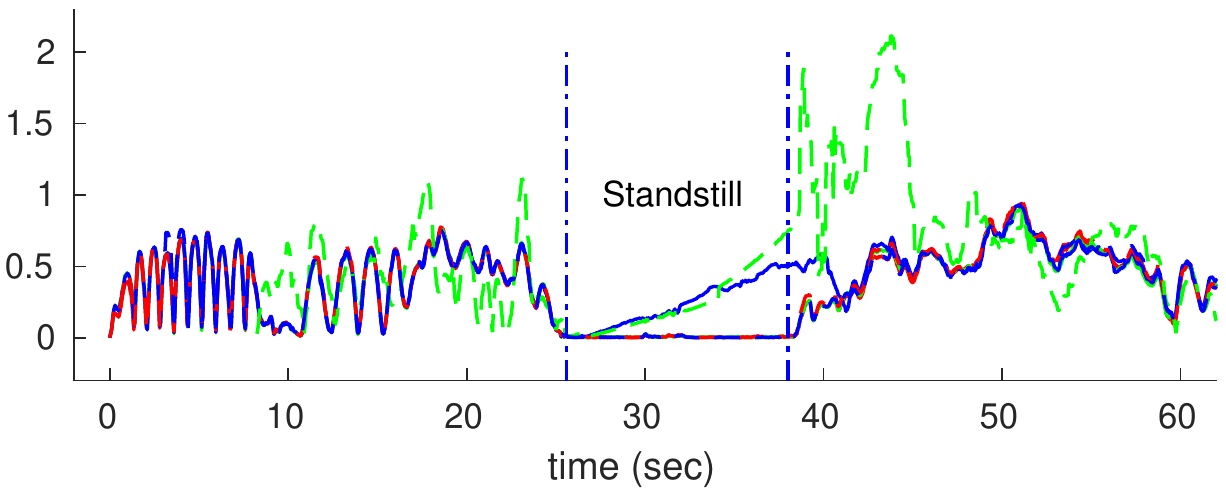}
	\includegraphics[width=0.8\columnwidth]{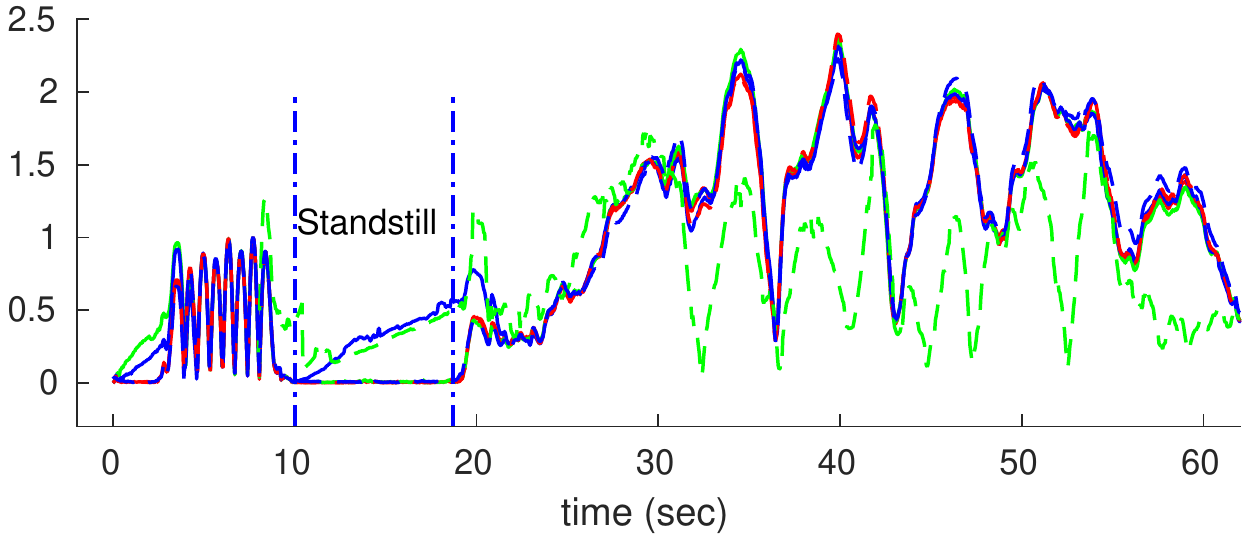}
	\includegraphics[width=0.8\columnwidth]{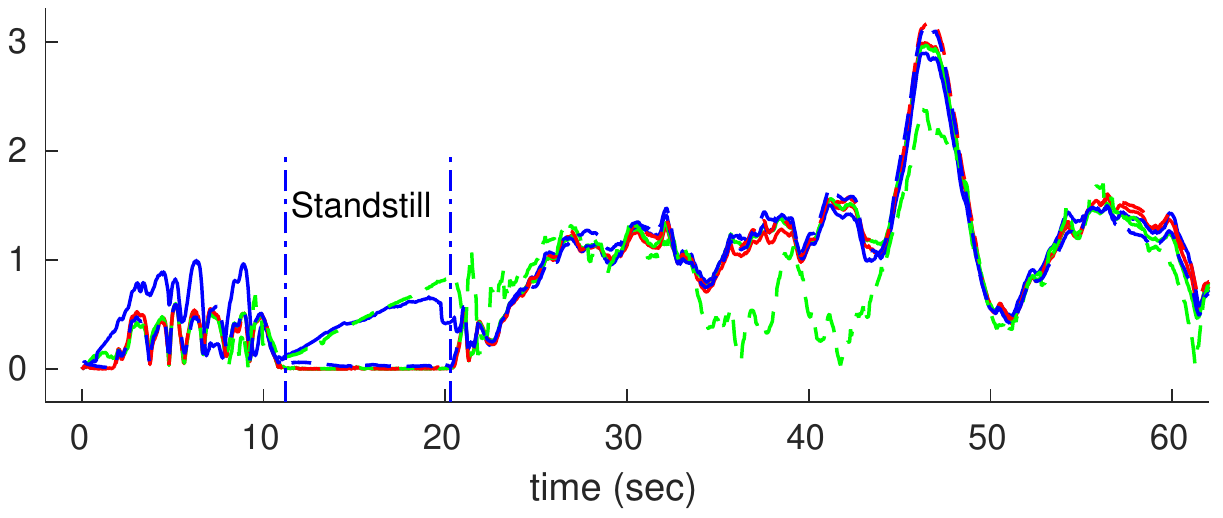}
	\includegraphics[width=0.8\columnwidth]{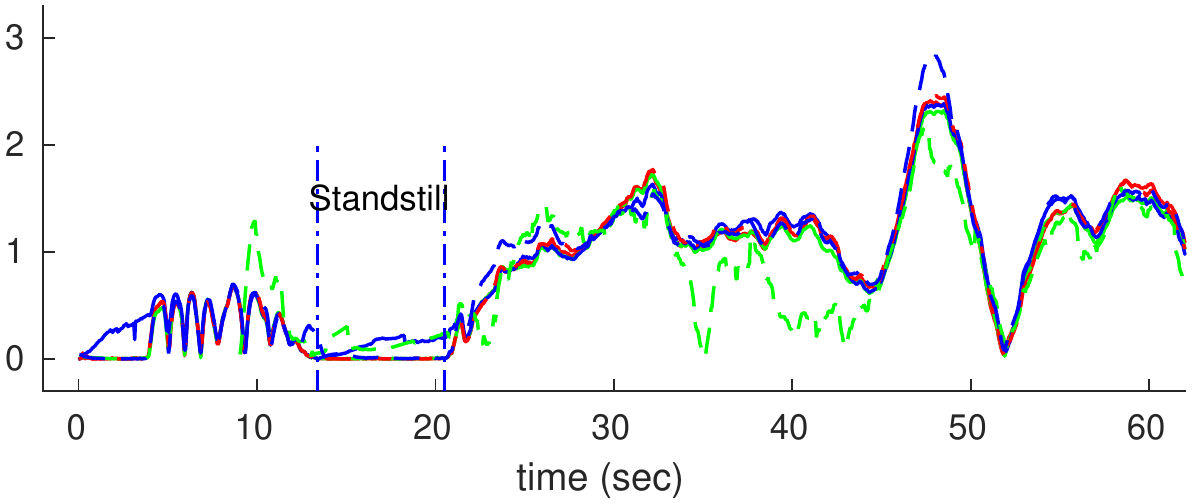}
	\caption{Velocity estimated by KSWF, SL-KSWF, SWF, SL-SWF, MSCKF-mono, and ROVIO for EuRoC MH 1-5 sequences.
		All methods ran in monocular mode using the left camera images. 
		SL-SWF and MSCKF-mono have nonzero velocity during standstills in contrast to other methods.}
	\label{fig:stationary-mh}
\end{figure}

This test was conducted on the five Machine Hall sessions of the EuRoC benchmark ~\cite{burriEuRoCMicroAerial2016} which have stationary periods.
The datasets were captured indoors by a micro aerial vehicle (MAV) outfitted with GS stereo cameras and an industrial-grade IMU.
This test used data captured by the IMU and the left camera for which the pinhole radial tangential model was used in the compared filters.

All filters initialized successfully and gave good motion estimates before a standstill.
Then we could clearly see the effect of standstill on motion estimation.
Fig.~\ref{fig:stationary-mh} visualizes the absolute velocity before and after standstills.
The area under the velocity curve during standstill roughly represents the position drift.
When the MAV landed, the frame-wise structureless filters, both MSCKF-Mono and SL-SWF, suffered position drifts,
although MSCKF-Mono organizes state variables in terms of keyframes in the backend.
Interestingly, both filters managed to estimate the velocity more or less accurately after the stationary periods.
We think that the drift of MSCKF-Mono and SL-SWF is caused by frame-wise feature matching which provides many low disparity matches during standstills.

By contrast, filters that keep landmarks in the state, ROVIO, KSWF, and SWF,
and filters that use keyframe-based matching, SL-KSWF and KSWF,
reported almost zero velocity in standstill.
On one hand, landmarks in the state vector serve as memory, constraining the filter during degenerate motion.
On the other hand, feature matching relative to previous keyframes which do not depend on time retrieves matches of large disparity even during stationary periods.

This comparative test illustrates that keyframe-based feature matching helps structureless filters deal with degenerate motion, \eg, standstill, 
and keeping landmarks in the state vector also resolves this issue.

\subsection{Motion Estimation with Self-Calibration}
\label{subsec:motion-inaccurate-calib}
To show the feasibility and benefits of self-calibration, tests were conducted on raw room sequences of the TUM VI benchmark \cite{schubertTUMVIBenchmark2018}.
These sequences were captured by a handheld device with stereo GS cameras with fisheye lenses of diagonal FOV 195$^\circ$ and a consumer-grade IMU.
Ground truth motion was recorded by a motion capture system, supporting the validation of VIO methods.
The authors provided raw sequences, calibration results for the sensor setup, and calibrated sequences where the IMU data were corrected with the estimated IMU intrinsic parameters.
For this test, we chose the six room sequences because they have diverse motion patterns.
The camera-IMU calibration results provided in the benchmark served as reference to validate the full self-calibration.

To prepare the raw sequences, images were down-scaled to 512 $\times$ 512 as in \cite{schubertTUMVIBenchmark2018},
and timestamps of camera and IMU messages were shifted by a constant to be aligned with the ground truth for the sake of evaluation.

Then, these sequences were processed by KSWF and several recent VIO methods listed in Table~\ref{tab:features-odometry-methods}.
Five KSWF variants were used including
KSWF (with full calibration by default), SL-KSWF, KSWF with minimal self-calibration, 
KSWF that calibrates IMU intrinsics, 
and KSWF that calibrates camera intrinsics, $t_d^i$, and $t_r^i$, $i=0,1$.
For comparison, OKVIS \cite{leuteneggerKeyframebased2015}, VINS-Mono \cite{qinVINSMono2018}, 
and OpenVINS \cite{genevaOpenVINSResearchPlatform2019} were used.
Note that the last variant of KSWF is similar to OpenVINS \cite{genevaOpenVINSResearchPlatform2019} in configuration.
VINS-Mono only processed the left camera images and IMU data.

These methods were mostly configured using parameters provided by the authors, \eg, OpenVINS and VINS-Mono.
For OKVIS, the configuration file provided by the TUM VI benchmark was used as a starting point.
Then, for all estimators, sensor parameters were initialized to nominal values.
If a parameter was not to be calibrated, it was locked to the initial value by zeroing its variance.
Initial values for IMU intrinsic parameters are $\mbf M_g = \mbf I$, $\mbf T_s = \mbf 0$, and $\mbf M_a = \mbf I$.
The camera extrinsic parameters were set zero in translation and a nominal rotation in orientation.
The Kannala Brandt 8-parameter model \cite{kannalaGeneric2006} were used for the stereo cameras with initial values $f_x^i = f_y^i = 190$ (px),
$c_x^i = 256$ (px), $c_y^i = 256$ (px), $k_1^i = 0$, $k_2^i = 0$, $k_3^i = 0$, and $k_4^i = 0$ for $i = 0, 1$.
The temporal parameters had initial values $t_d^i = 0$ s and $t_r^i = 0.02$ s for $i = 0, 1$.
Because the raw sequences have large IMU errors, for all estimators, the IMU noise parameters were inflated from values provided by authors with a simple grid search.

\begin{figure}[!htb]
	\centering
	\includegraphics[width=0.8\columnwidth]{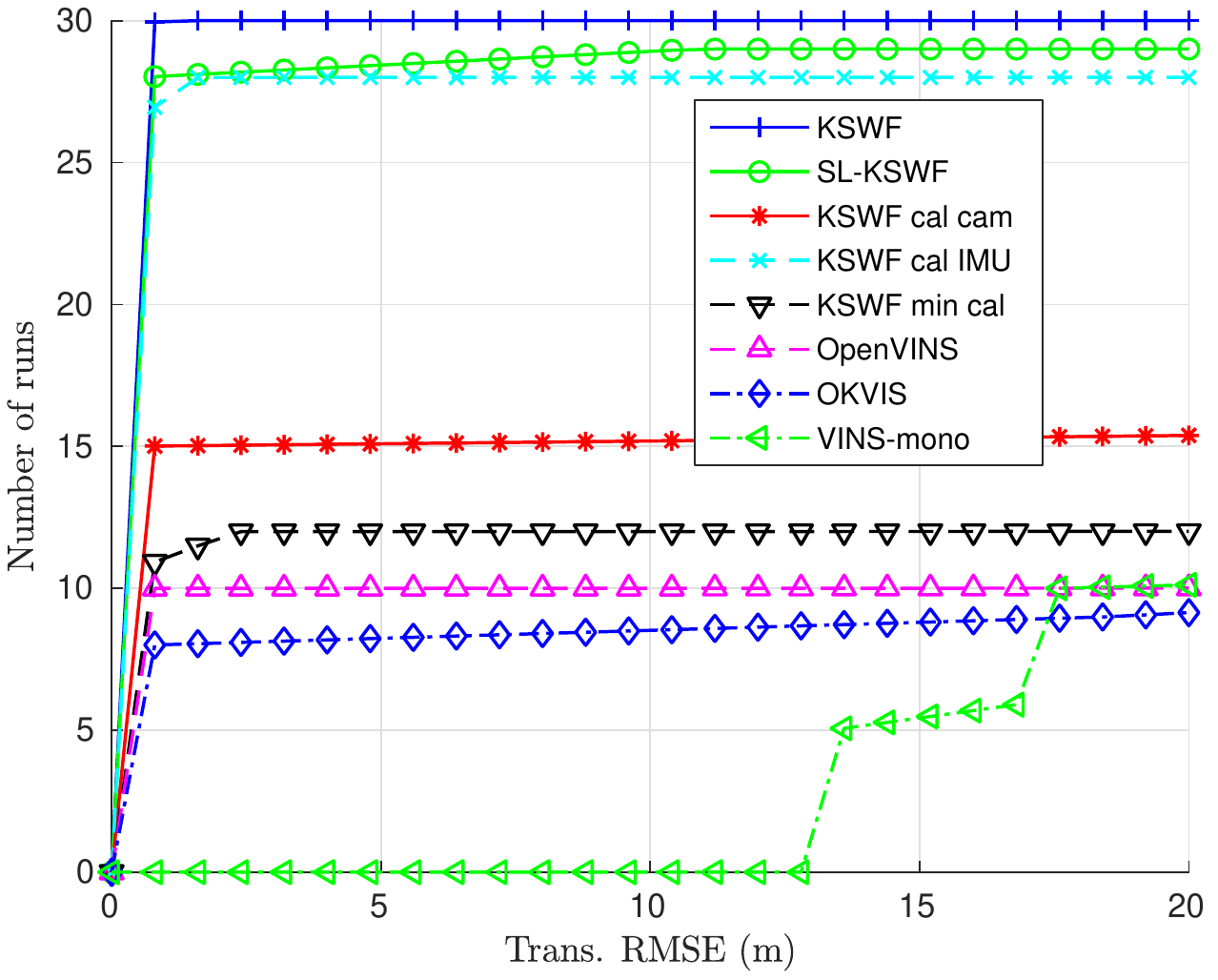}
	\includegraphics[width=0.8\columnwidth]{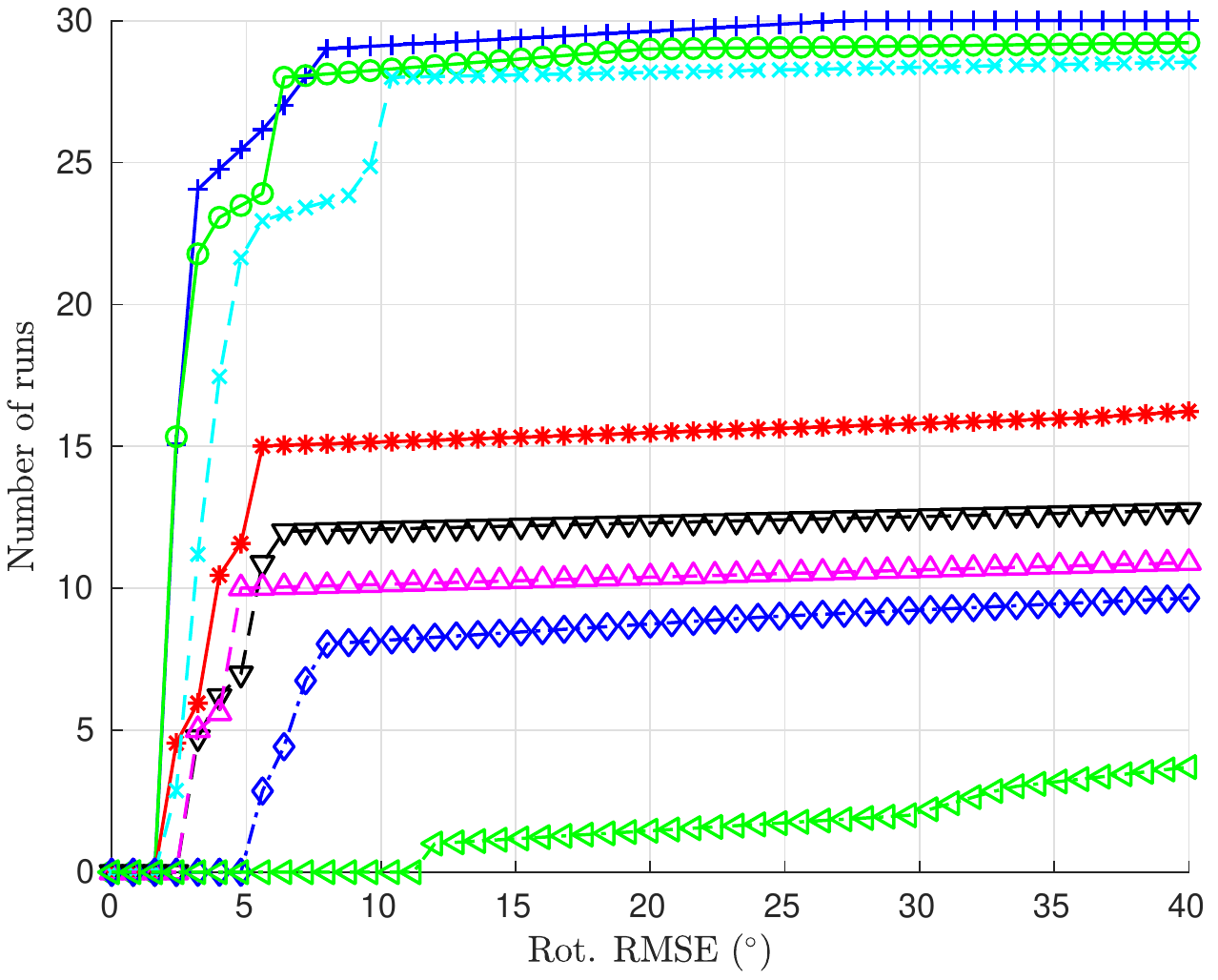}
	\caption{Translation and rotation RMSE of odometry methods, KSWF with default full calibration, SL-KSWF with default full calibration,
KSWF that calibrates camera intrinsic parameters, KSWF that calibrates IMU intrinsic parameters,
KSWF with minimal calibration, OKVIS, OpenVINS, VINS-mono, on the TUM VI six raw stereo room sequences. 
All methods calibrate IMU biases and camera extrinsic parameters on the fly. Each method ran five times on a sequence.}
	\label{fig:cumulative-rmse}
\end{figure}

For these estimators starting from inaccurate parameters, the error metrics of estimated motion versus the number of runs on the raw room sequences are drawn in Fig~\ref{fig:cumulative-rmse}.
The relative pose errors and number of failed runs with absolute translation error \textgreater 100 m or absolute rotation error \textgreater 15$^\circ$ are listed in Table~\ref{tab:tumvi-room-rpe}.
Methods with very limited self-calibration, \eg, OKVIS and VINS-Mono, suffered significantly in motion estimation accuracy and often diverged.
This contrasts with their performance on the calibrated room sequences with accurate sensor parameters, 
for which all compared estimators estimated motion accurately without obvious drift, 
achieving \textless 0.2\% relative translation error \cite{zhangTutorial2018}.
Comparing the variants of KSWF, as the level of self-calibration increased, the motion estimation became less brittle and more accurate.
This is particularly true when the IMU intrinsics were calibrated, likely because the consumer-grade IMU is notoriously affected by scale factor errors and misalignment.
Interestingly, OpenVINS and KSWF that calibrated camera parameters had similar performance as both calibrated almost the same set of parameters.

\begin{figure*}[!htb]
\centering
\includegraphics[width=0.9\columnwidth]{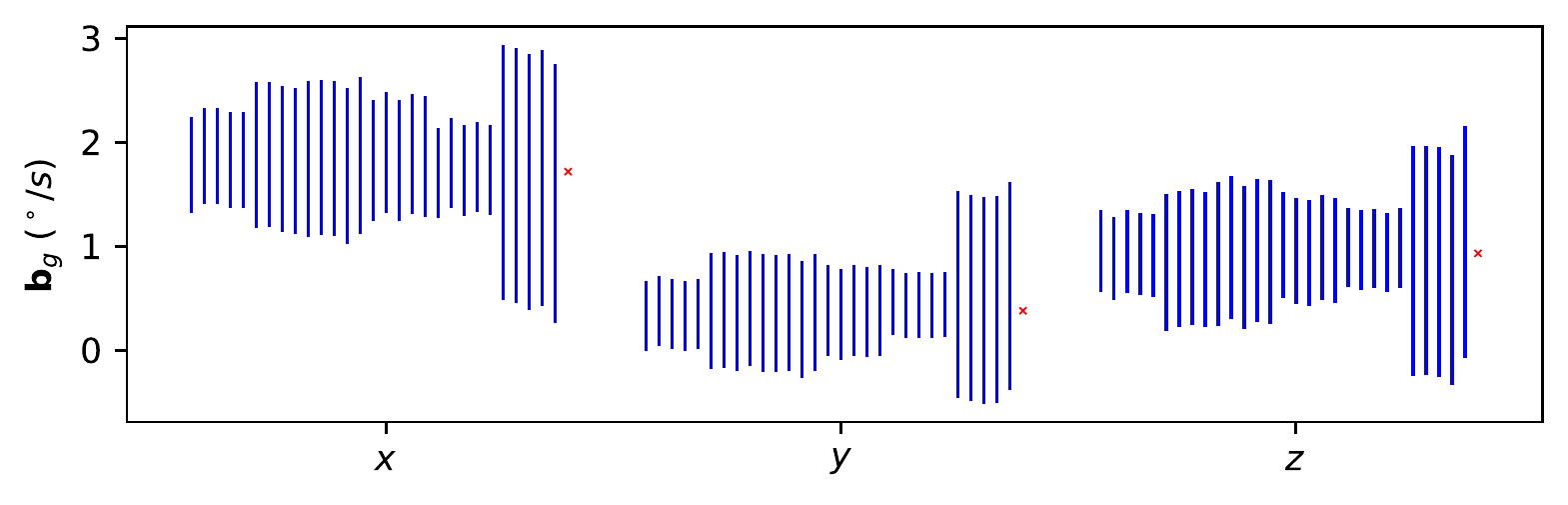}
\includegraphics[width=0.9\columnwidth]{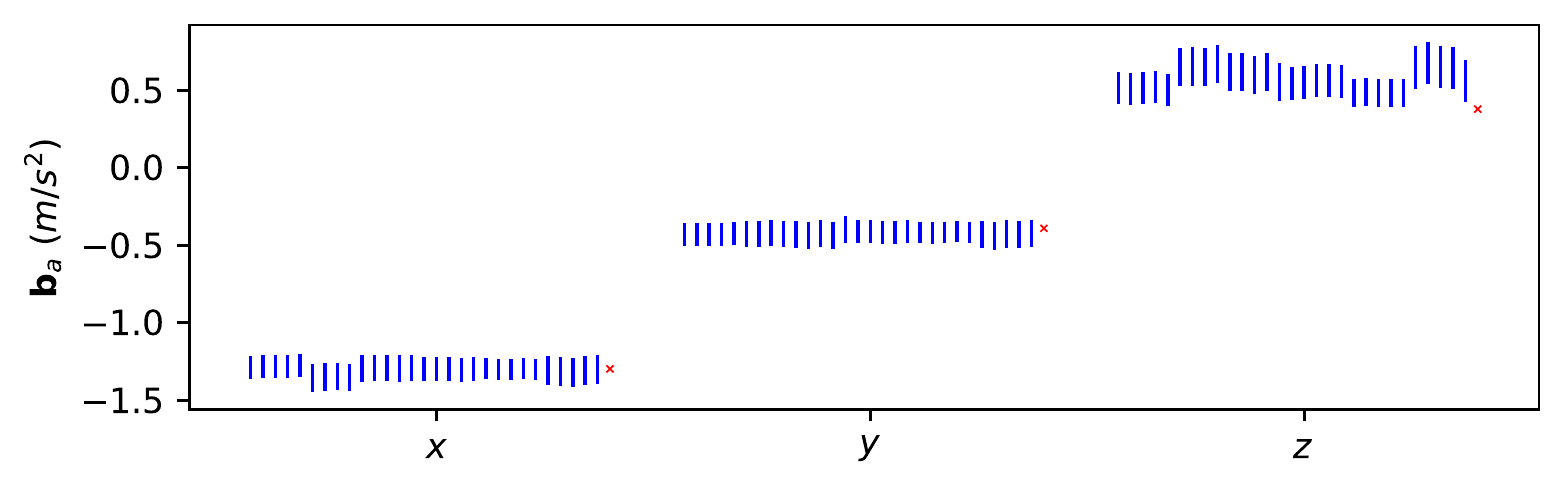}
\includegraphics[width=0.9\columnwidth]{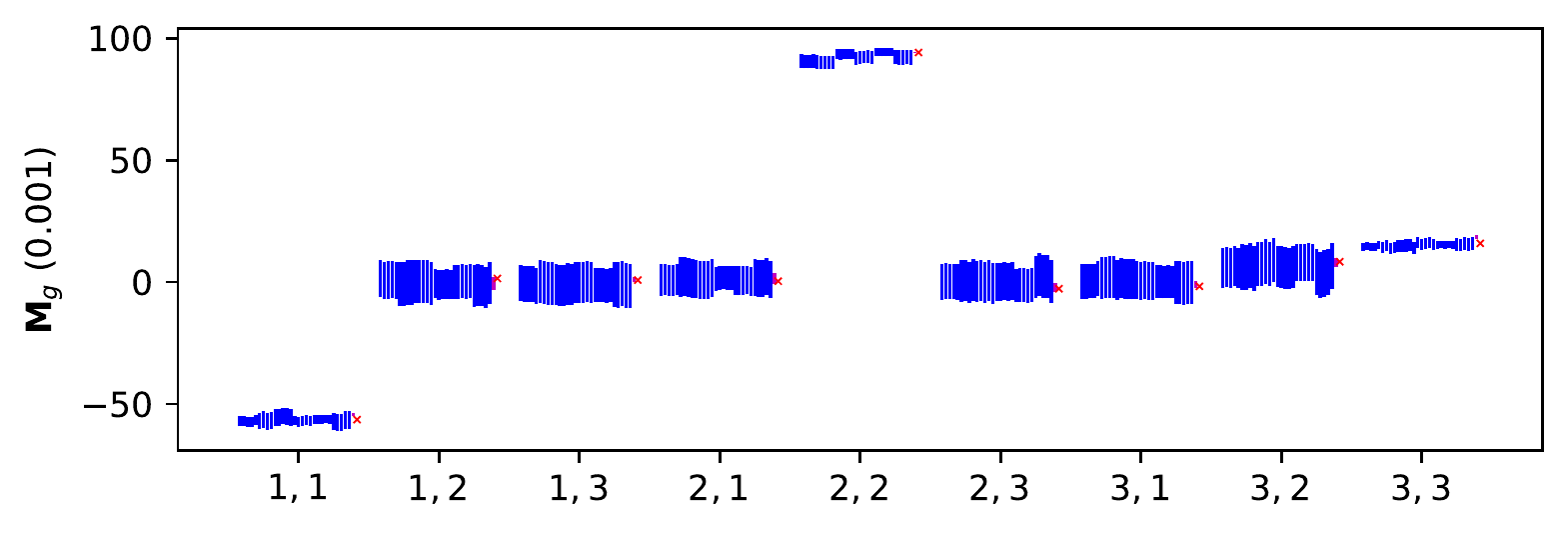}
\includegraphics[width=0.9\columnwidth]{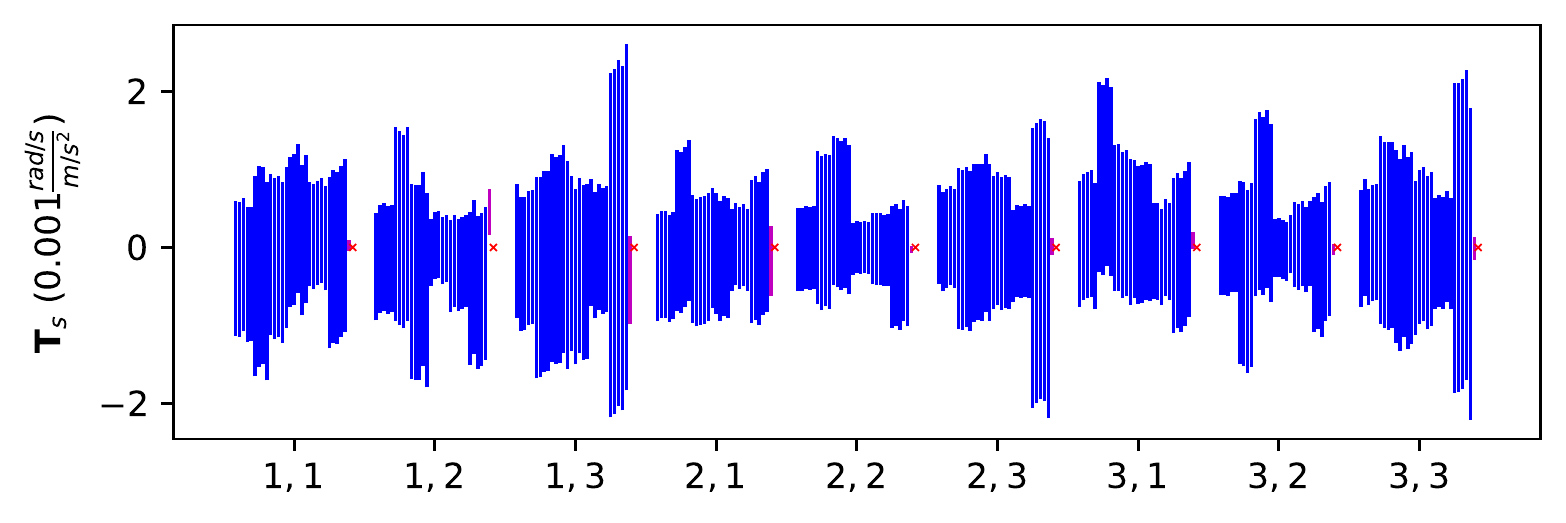}
\includegraphics[width=0.9\columnwidth]{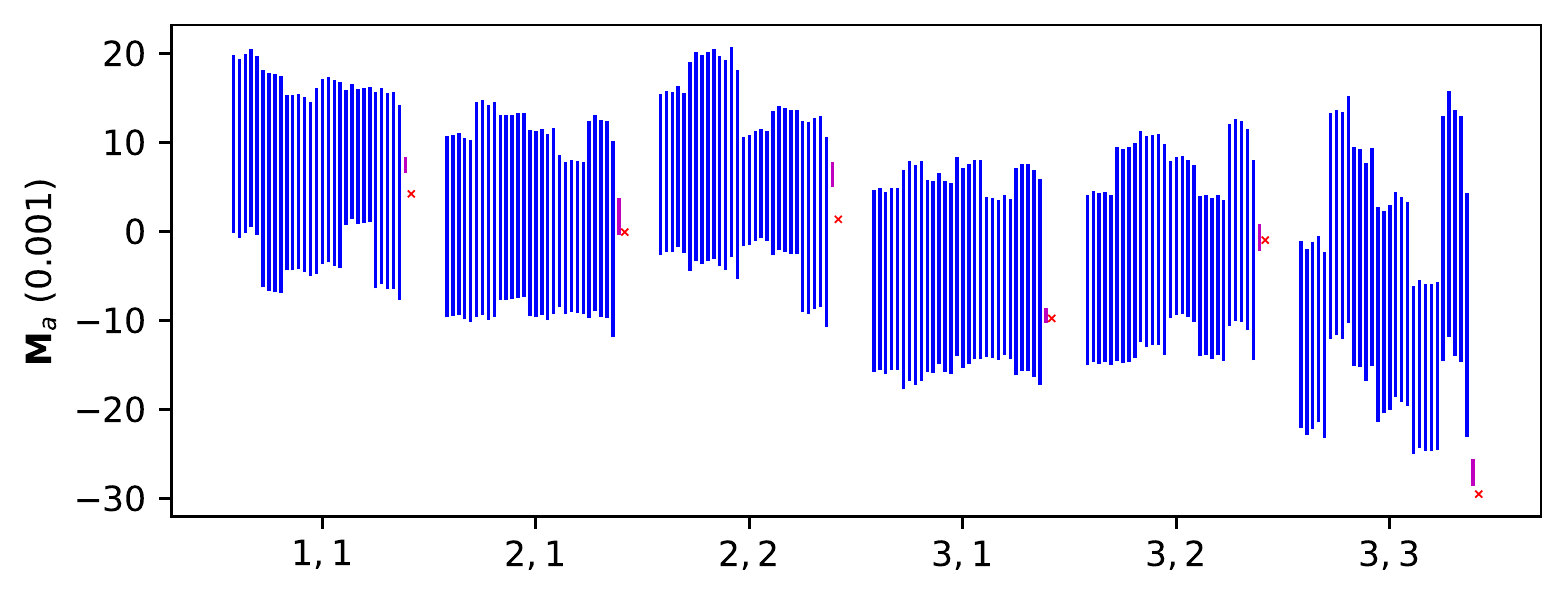}
\includegraphics[width=0.9\columnwidth]{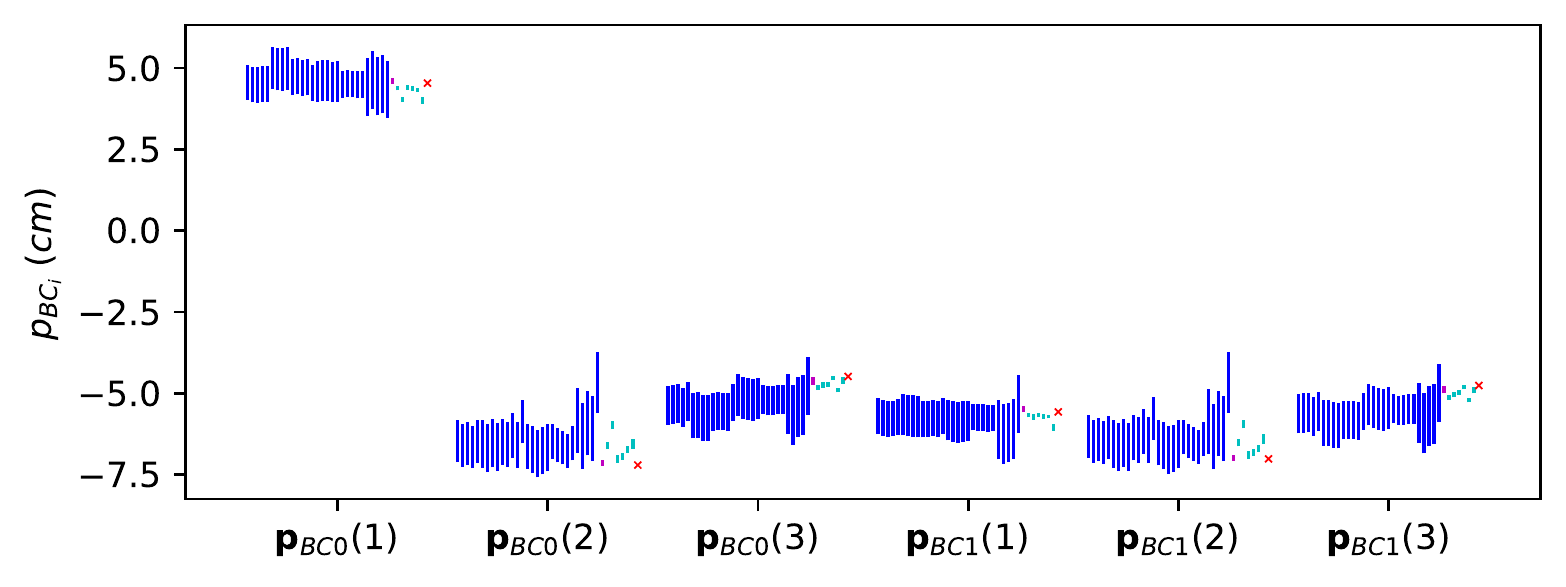}
\includegraphics[width=0.9\columnwidth]{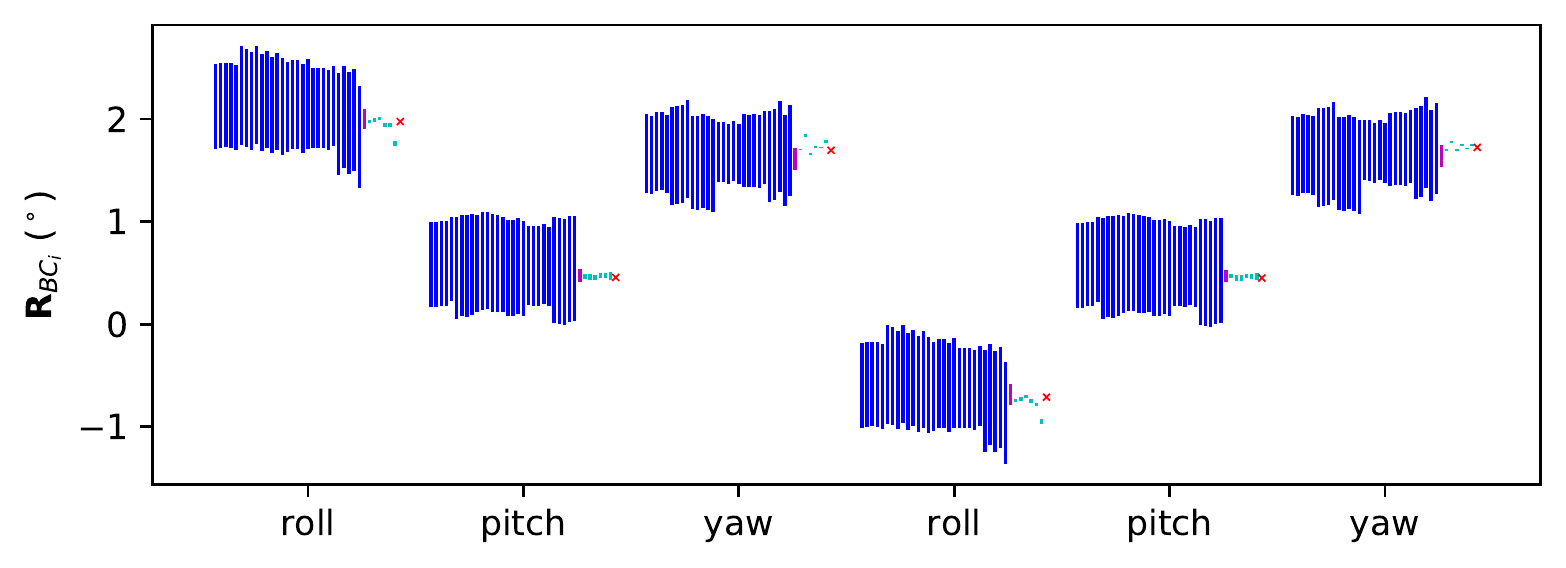}
\includegraphics[width=0.9\columnwidth]{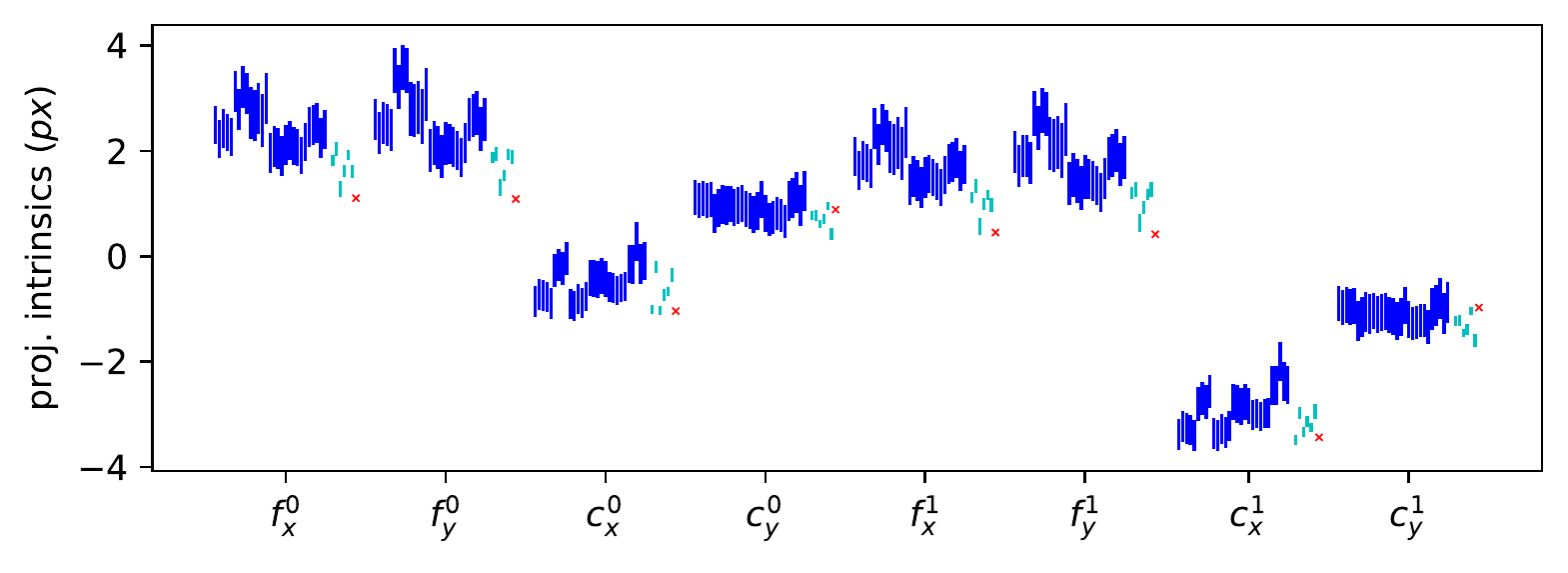}
\includegraphics[width=0.9\columnwidth]{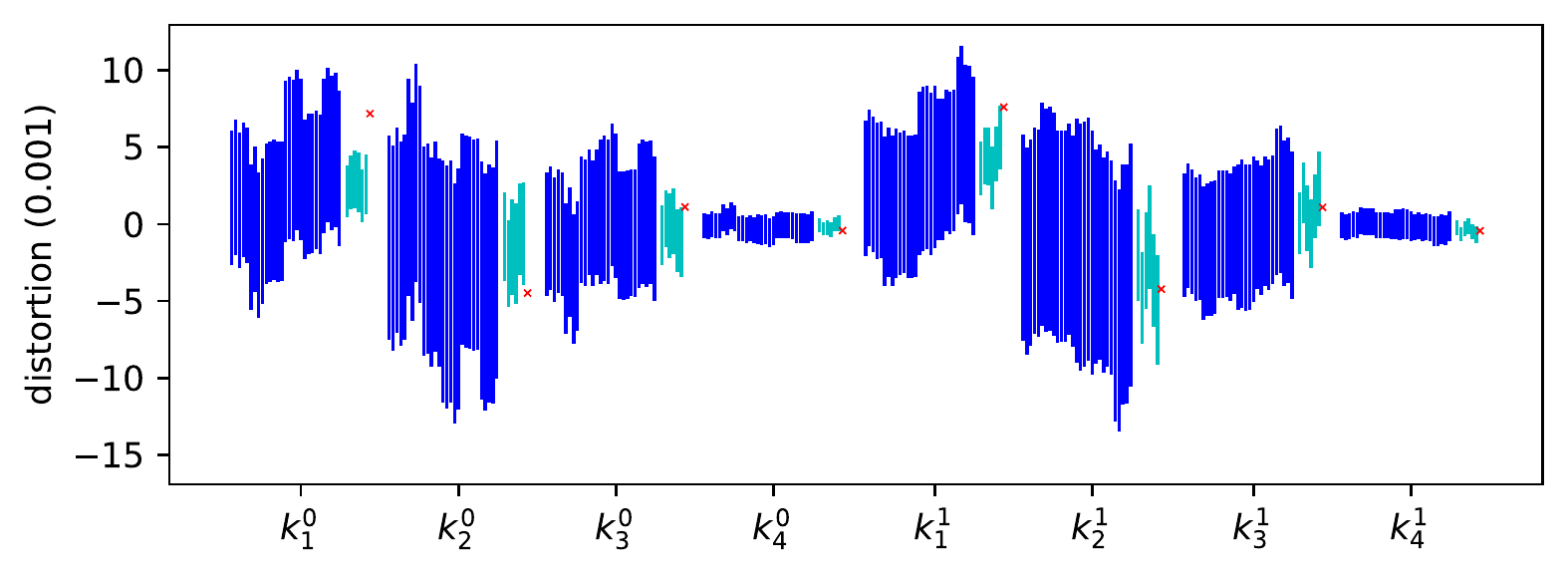}
\includegraphics[width=0.9\columnwidth]{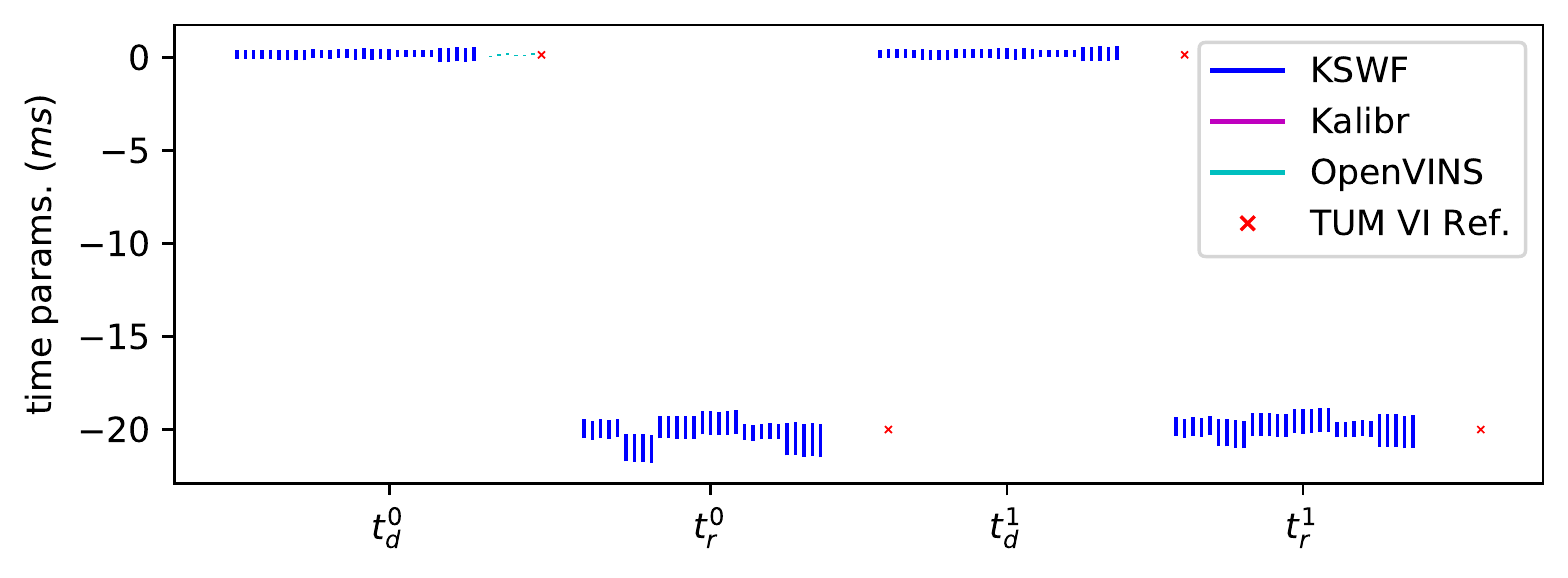}
\caption{Difference of parameter estimates relative to initial values and 3$\sigma$ bounds for the TUM VI benchmark.
Results of KSWF on the six raw room sequences (five attempts for each sequence, 29 times successful) are shown by blue bars.
Result of the camera-IMU calibrator in Kalibr on the calibration sequences of the benchmark is shown by the magenta bars.
The calibrator estimates the IMU intrinsic parameters and camera extrinsic parameters.
Results of OpenVINS on the six calibrated room sequences,
are shown by cyan bars (one run for each sequence).
OpenVINS estimates the camera extrinsic and intrinsic parameters and a camera-IMU time offset $t_d^0$. IMU biases for OpenVINS is omitted since the calibrated sequences are corrected in IMU data.
Red crosses show reference values from the benchmark.
Note that frame readout times were initialized to 20 $ms$.}
\label{fig:std-params}
\end{figure*}

To examine observability of parameters in full self-calibration,
we compared their values estimated by KSWF and OpenVINS, those provided in the benchmark, and those estimated by the Kalibr toolbox.
For KSWF, the parameter estimates are from the 29 convergent runs of KSWF in the previous test on the raw room sequences.
For OpenVINS, we estimated camera-related parameters by running OpenVINS on the six calibrated room sequences.
For reference, the benchmark authors have provided the IMU intrinsic and camera extrinsic parameters.
Moreover, we independently calibrated the stereo camera-IMU system with the provided sequences for calibration by using the Kalibr toolbox \cite{rehderExtendingKalibrCalibrating2016},
resulting in estimates for camera extrinsic parameters and IMU scale, misalignment, and $g$-sensitivity.
For comparison, parameter estimates and variances from OpenVINS, the benchmark, and Kalibr were converted to our notation.
These parameter values with 3$\sigma$ bounds (if available) relative to initial values are visualized in 
Fig.~\ref{fig:std-params}.
Since OpenVINS gives identical results for five runs on the same sequence, only one estimate of a parameter is shown for each sequence.

\input{tables/tumvi-room-rpe.tex}

The plots in Fig.~\ref{fig:std-params} show that the reference values given in the benchmark and those obtained by Kalibr are fairly close.
For parameters estimated by KSWF, we see that almost all parameters were estimated close to the reference values, 
but $\mbf M_a(3,3)$, $f_x^k$, and $f_y^k$, $k=0, 1$, fell short to converge to the reference values with slight deviations.
Since all parameters in full calibration had moved from their initial inaccurate values toward the reference values, this confirms that they are all weakly observable.
For OpenVINS, since the IMU intrinsic errors had been mostly corrected
and only camera parameters were calibrated, we expect it to achieve estimates closer to the reference and to have smaller variances than KSWF which ran on the raw sequences.

To evaluate the efficiency of KSWF, we timed its four components,
feature extraction, keyframe-based feature matching, filtering (including propagation and
update), and marginalization, with the EuRoC benchmark \cite{burriEuRoCMicroAerial2016}, 
on a Thinkpad P53 laptop with 32 GB RAM and a 2.50 GHz 8-core Intel Core i5-9400H processor running Ubuntu 18.04.
For the sake of timing, the feature matching is done with only one thread.
The time costs of KSWF, KSWF with minimal calibration (only calibrating biases and camera extrinsic parameters), SL-KSWF with minimal calibration, 
in monocular and stereo mode, are presented in Table~\ref{tab:euroc-timing}.
The SL-KSWF takes less than half of the time used by KSWF for filtering since the landmarks are excluded from the state vector.
By comparing KSWF and KSWF with minimal calibration, we see that calibrating camera-IMU intrinsic parameters increases the time expense of filtering by about 30\%.
As feature extraction is done in parallel, average processing time for a frame bundle of KSWF 
amounts to the time costs for feature matching, filtering, and marginalization, thereby, 7.8 ms (128 Hz) in the monocular
mode, and 21.7 ms (46 Hz) in the stereo mode, meaning that the proposed method has good efficiency even with full calibration.

\input{tables/euroc-timing.tex}

\section{Conclusion And Future Work}
\label{sec:conclusion}

Regarding VIO with self-calibration, this paper solves two existing problems: 
whether the full set of calibration parameters are observable, 
and the drift issue in conventional structureless monocular filters under degenerate motion.

For the former problem, using Lie derivatives, we prove that camera intrinsics, extrinsics, time offset, camera readout time, and IMU intrinsics of the camera-IMU system are observable
under general motion using observations of opportunistic landmarks.
These observability assertions were confirmed with simulation and real data tests.
Our insight is that automated symbolic deduction greatly simplifies observability analysis of 
a slew of parameter estimation problems and eases result verification.

For the latter problem, we introduce the keyframe concept to both the feature matching frontend and sliding window filtering backend. 
Together with the self-calibration capability, this forms the KSWF framework.
Real world tests validated the strength of keyframe-based filtering, revealing that either keyframe-based matching or 
keeping landmarks in the state vector helps remedy the drift issue of structureless monocular filters in standstill.
Moreover, on datasets captured by inaccurate sensors, KSWF with self-calibration largely outperformed recent VIO methods of minimal calibration, 
indicating the value of self-calibration for low-cost sensors.

We are aware that full self-calibration will encounter unobservable dimensions 
when the camera-IMU system is confined in motion, \eg, on a ground vehicle.
In such situations, it is advisable to fix those unobservable parameters.
To this end, identifying unobservable directions with Lie derivatives in constrained motion is one of the worthwhile directions to look into.
A related problem is VIO initialization in the case of coarse calibration.

\appendices

\section{Relation between Observability and Rank of the Jacobian Matrix}
\label{app:proof}
Since \cite[Corollary 2.1]{maesObservability2019} is presented without proof, here we provide a sketch.
Consider the nonlinear system that is affine in inputs, $u_i$, $i=1,2, \cdots, m$,
\begin{equation}
	\dot{\mbf{x}} = \mbf{f}_0(\mbf x) + \sum_{i = 1}^{m}\mbf f_i(\mbf x) u_i, \quad
	\mbf{y}(t) = \mbf h(\mbf x(t)).
\end{equation}
The problem is to estimate $\mbf x$ given control inputs 
$\mathbf{u} = [u_1, u_2, \cdots, u_m]\tran$ and observations in time interval $[t, t+\Delta t]$.
First, the knowledge of $\mbf y$ in $[t, t+\Delta t]$ is equivalent to the knowledge of $\frac{d^j\mbf h}{dt^j}, j = 0, 1, \cdots$, by the Taylor theorem.
Time derivatives of $\mbf h$ bring about a set of nonlinear equations of $\mbf x$,
\begin{equation}
\mbf Y = \begin{bmatrix}
\mbf y \\ \frac{\partial \mbf y}{\partial t} \\ \vdots \\ \frac{\partial^j \mbf y}{\partial t^j}
\end{bmatrix} = \begin{bmatrix}
\mbf h(\mbf x) \\ \frac{\partial \mbf h}{\partial t}(\mbf x) \\ \vdots \\
\frac{\partial^j \mbf h}{\partial t^j}(\mbf x)
\end{bmatrix} = \mbf G(\mbf x)
\end{equation}
To solve for $\mbf x$, we can linearize $\mbf G(\mbf x)$ at a rough value of $\mbf x$, $\mbf x_0$,
obtaining a linearized system,
\begin{equation}
\label{eq:linearized-sys}
\begin{split}
	\mbf Y - \mbf G(\mbf x_0) \simeq \mbf J\delta \mbf x \\
	\mbf J = \frac{\partial \mbf G}{\partial \mbf x} \bigg|_{\mbf x_0} =
	\begin{bmatrix}
		\mbf{J}_r & \mbf J_n
	\end{bmatrix}, \enskip \delta \mbf x = \begin{bmatrix}
	\delta \mbf x_r \\ \delta x_n
\end{bmatrix}
\end{split}
\end{equation}
where, without loss of generality, we choose a variable in $\mbf x$, $x_n$, and put it behind the remaining variables $\mbf x_r$.

By referring to the definition of weak observability, and basic matrix properties, 
we have the following equivalent statements.

\begin{lemma}
\label{lem:equiv}
$x_n$ is weakly observable. $\Leftrightarrow$
$\delta x_n$ can be solved with the linearized system \eqref{eq:linearized-sys}. $\Leftrightarrow$
The effect of a small change in $x_n$ on the linearized system is unique and cannot be
represented by the combined effect of a small change in $\mbf x_r$. $\Leftrightarrow$
For all real coefficient vector $\mbf c$, and a small number $\epsilon$, $\mbf J_r \mbf c \epsilon + \mbf J_n \epsilon \neq \mathbf{0}$. 
$\Leftrightarrow$ $\mbf J_n$ is independent of $\mbf J_r$. $\Leftrightarrow$
$\mathrm{rank}(\mbf J) = \mathrm{rank}(\mbf J_r) + 1$.
\end{lemma}

Of course, contrapositives of these statements are true, \eg, 
$x_n$ is weakly unobservable. $\Leftrightarrow$
$\mbf J_n$ can be expressed by $\mbf J_r$. $\Leftrightarrow$
$\mathrm{rank}(\mbf J) = \mathrm{rank}(\mbf J_r)$.

\section{When Time Offset Is Unobservable}
\label{app:time-offset}
This section identifies the condition when the time offset between inputs and outputs of a nonlinear system affine in inputs is unobservable.
The system \eqref{eq:affine-input-sys} with inputs $\mbf u = [u_1, u_2, \cdots, u_m]\tran$ is given by
\begin{equation}
\dot{\mbf{x}} = \mbf{f}_0(\mbf x) + \sum_{i = 1}^{m}\mbf f_i(\mbf x) u_i, \quad
\mbf{y}(t) = \mbf h(\mbf x(t + t_d)),
\end{equation}
where $t_d$ is an unknown time offset.
Next we establish a set of nonlinear equations of the unknowns $\mbf x(t + t_d)$ and $t_d$ and analyze the condition to solve for $t_d$ given $\mbf u(t + t_d)$ and $\mbf y(t)$ for $t \in [t, t+\Delta t]$.

First, by the Taylor theorem, we know the values of all time derivatives of $\mbf y$ at $t$ by the Taylor theorem.
On the other hand, each time derivative of $\mbf y$ is a linear combination of products of Lie derivatives (\eg, $\mathcal{L}^0\mbf h$, $\mathcal{L}^1_{\mbf f_i} h$)
and time derivatives of inputs (\eg, 1, $u_i$, $\dot{u}_i$).
This fact can be proved by induction, since
\begin{equation}
	\mbf y(t) = \mbf h(\mbf x(t + t_d)) = \mathcal{L}^0\mbf h
\end{equation}
\begin{equation}
\begin{split}
\frac{\partial \mbf y}{\partial t} &= \frac{\partial \mbf{h}(\mbf x(t+t_d))}{\partial t} = \nabla \mbf h \cdot \dot{\mbf x}(t+t_d) \\
&= \mathcal{L}^1_{\mbf f_0} \mbf h + \sum_{i=1}^{m} \mathcal{L}_{\mbf f_i}^1 \mbf h u_i(t+t_d)
\end{split}
\end{equation}
\begin{equation}
	\begin{split}
	\ddot {\mbf y}(t) &= \ddot {\mbf h}(\mbf x(t+t_d)) \\ 
	&= \nabla (\mathcal{L}^1_{\mbf f_0} \mbf h + \sum_{i=1}^{m} \mathcal{L}_{\mbf f_i}^1 \mbf h u_i) \dot{\mbf x} + \sum_{i=1}^{m} \mathcal{L}_{\mbf f_i}^1 \mbf h \dot{u}_i 
		\\&= \mathcal{L}^2_{\mbf f_0, \mbf f_0} \mbf h + \sum_{i=1}^m \mathcal{L}^2_{\mbf f_0, \mbf f_i} \mbf h u_i + \sum_{i=1}^m \mathcal{L}^2_{\mbf f_i, \mbf f_0} \mbf h u_i + 
		\\&\quad \sum_{j=1}^m \sum_{i=1}^m \mathcal{L}^2_{\mbf f_i,\mbf f_j} \mbf h u_i u_j + \sum_{i=1}^{m} \mathcal{L}_{\mbf f_i}^1 \mbf h \dot{u}_i
	\end{split}
\end{equation}
Therefore, we can write the time derivatives of $\mbf y$ as
\begin{equation}
\mbf Y = \mbf U(\mbf u(t+t_d), \dot{\mbf u}(t+t_d), \cdots) \cdot \mbf H(\mbf x(t + t_d)),
\end{equation}
where $\mbf Y$ is the vector of derivatives of $\mbf y$, $\mbf U$ is a matrix consisting of products and derivatives of $\mbf u$ at $t+t_d$, 
and $\mbf H$ is a vector consisting of all Lie derivatives of $\mbf h$, \ie,
\begin{equation}
\begin{split}
\mbf Y &= \begin{bmatrix} \mbf y(t) & \dot{\mbf y}(t) & \cdots\end{bmatrix}\tran, \\
\mbf H &= \begin{bmatrix}
    \mathcal{L}^0 \mbf h & \mathcal{L}_{\mbf f_i}^1 \mbf h & \mathcal{L}_{\mbf f_i, \mbf f_j}^2 \mbf h & \cdots
\end{bmatrix}\tran,\\
\mbf U &= \begin{bmatrix}
    1 & 0 & 0 & 0 & \cdots \\
    0 & 1 & u_1 & u_2 & \cdots \\
    0 & 0 & \dot{u}_1 & \dot{u_2} & \cdots \\
    \mbf 0 & \mbf 0 & \cdots & \cdots & \cdots
\end{bmatrix}.
\end{split}
\end{equation}

The Jacobian of the nonlinear system is 
\begin{equation}
	\begin{split}
		\mbf J &= \left[\frac{\partial \mbf{UH}}{\partial \mbf x}, \frac{\partial \mbf{UH}}{\partial t_d}\right]	\\
		&= \left[\mbf U \frac{\partial \mbf H}{\partial \mbf x}, \frac{\partial \mbf U}{\partial t_d} \mbf H + \mbf U \frac{\partial \mbf H}{\partial t_d}\right] \\
		&= \left[\mbf U \frac{\partial \mbf H}{\partial \mbf x}, \frac{\partial \mbf U}{\partial t_d} \mbf H + \mbf U \frac{\partial \mbf H}{\partial \mbf x} \frac{\partial \mbf x}{\partial t_d}\right] \\
		&\equiv [\mbf J_{\mbf x}, \mbf J_{t_d}]
	\end{split}.
\end{equation}
By using Lemma \ref{lem:equiv}, we know that when $\mbf J_{t_d} = \mbf c \mbf J_{\mbf x}$, for a coefficient vector $\mbf c$, the time offset is unobservable.
Obviously, we only need to focus on $\frac{\partial \mbf U}{\partial t_d} \mbf H$ which in general is independent of $\mbf U \frac{\partial \mbf H}{\partial \mbf x}$.
However, there are two exceptions: 1) when the control inputs are constant, $\frac{\partial \mbf U}{\partial t_d}$ becomes a zero matrix; 2) when the observations are trivially constant, $\frac{\partial \mbf U}{\partial t_d} \mbf H$ becomes a zero vector.
These two exceptions combine into Theorem~\ref{theorem}.
Because of the general independence between $\mbf J_{\mbf x}$ and $\mbf J_{t_d}$, the two conditions are almost necessary to render time offset weakly unobservable.

\section{Observability Analysis of Frame Readout Time}
\label{app:readout-time}
Following the approach in Appendix~\ref{app:time-offset}, we can show that the frame readout time is generally weakly observable.
To simplify notation, we will work with the line delay $d$ instead of readout time $t_r$ since $d=t_r / h$ where $h$ is the image height.
The VIO system with time offset $t_d$ and line delay $d$ is an instance of the nonlinear system affine in inputs, $\mbf u = [1, u_1, u_2, \cdots, u_m]\tran$,
\begin{equation}
\begin{split}
\dot{\mbf{x}}(t) &= \sum_{i = 0}^{m}\mbf f_i(\mbf {x}(t)) u_i(t)\\
\mbf{y}(t) &\equiv \begin{bmatrix} y_1(t), v(t)\end{bmatrix}\tran =
\mbf h(\mbf x(t + t_d + v(t) \cdot d))
\end{split}
\end{equation}
where $v(t)$ is the row of the image observation $\mbf y(t)$.

We first establish the nonlinear system of the unknowns $\mbf x(t + t_d + v(t)d)$, $t_d$, and $d$ given the time derivatives of $\mbf y$ at $t$, and control inputs $\mbf u(t + t_d + v(t)d)$ for $t \in [t, t+\Delta t]$.
Every time derivative of $\mbf y$ can be expressed by Lie derivatives of $\mbf h$, $d$, and time derivatives of control inputs. 
The first three are given by
\begin{equation}
\begin{split}
\mbf y(t) &= \mathcal{L}^0 \mbf h(\hat{t}), \\
\dot{\mbf y}(t) &= \frac{\partial \mbf h}{\partial \mbf x}\cdot 
\dot{\mbf x}(\hat{t}) \cdot (1 + \frac{\partial v}{\partial t}(t) d) \\
&= \left(\sum_{i=0}^m \mathcal{L}_{f_i}\mbf h u_i\right) (1+\frac{\partial v}{\partial t}(t) d),\\
\ddot{\mbf y}(t) &= \sum_{j=0}^m \sum_{i=0}^m \mathcal{L}_{f_i,f_j}^2\mbf h u_i u_j (1+\frac{\partial v}{\partial t} d)^2 + \\
&\quad \sum_{i=0}^m \mathcal{L}_{f_i}^1\mbf h \dot{u}_i (1+\frac{\partial v}{\partial t} d)^2 + \sum_{i=0}^m \mathcal{L}_{f_i}^1\mbf h u_i \frac{\partial^2 v}{\partial t^2} d
\end{split}
\end{equation}
where we define $\hat{t} \equiv t + t_d + v(t)d$ for brevity.
By induction, we can prove that the j-th order time derivative of $\mbf y(t)$ is a sum of products given by
\begin{equation}
\begin{split}
\mbf y^{(j)} &\equiv \frac{d^j \mbf y(t)}{dt^j} \\
&= \sum_i \mbf L_i(\mbf x) \cdot \mbf C_i(\mbf u, \dot{\mbf u}, \cdots) \cdot \mbf V_i(d, v(t), \dot{v}(t), \cdots),
\end{split}
\end{equation}
where $\mbf L_i(\mbf x), i=0, 1, \cdots$, are functions of only $\mbf x$ including these Lie derivatives of $\mbf h$,
$\mbf C_i$ are functions of control inputs and their time derivatives,
and $\mbf V_i$ are functions of $d, v(t)$, and its time derivatives.
Let's look at the derivatives of $\mbf y^{(j)}$ relative to $t_d$ and $d$,
\begin{equation}
\begin{split}
\frac{\partial \mbf y^{(j)}}{\partial t_d} &= \sum_i \frac{\partial \mbf L_i}{\partial \mbf x} \frac{\partial \mbf x}{\partial t} \mbf C_i \mbf V_i + \\
&\quad \sum_i \mbf L_i \left(\sum_k \frac{\partial \mbf C_i}{\partial u_k} \frac{du_k}{dt} + 
\sum_k \frac{\partial \mbf C_i}{\partial \dot{u}_k} \frac{d\dot{u}_k}{dt} + \cdots 
\right) \\
\frac{\partial \mbf y^{(j)}}{\partial d} &= \frac{\partial \mbf y^{(j)}}{\partial t_d} \cdot v(t) +
\sum_i \mbf L_i \mbf C_i \frac{\partial \mbf V_i}{\partial d}.
\end{split}
\end{equation}
That is, every $\frac{\partial \mbf y^{(j)}}{\partial d}$ involves a unique time derivative of $v(t)$ at least one order higher than time derivatives of $v(t)$ 
shown up in $\frac{\partial \mbf y^{(j)}}{\partial \mbf x}$ and $\frac{\partial \mbf y^{(j)}}{\partial t_d}$.
For the whole nonlinear system, its Jacobian relative to $d$, $\mbf J_d$, is expressed by
\begin{equation}
	\mbf J_{d} = \mbf J_{t_d} v(t) + \mbf E(\mbf x, \mbf u, \dot{\mbf u}, v, v^{(1)}, \cdots)
	[v \enskip v^{(1)} \enskip v^{(2)} \enskip \cdots]
	\tran,
\end{equation}
where $\mbf E$ is the coefficient matrix for the time derivative vector of $v(t)$.
In general, since it is impossible to get $v^{(j)}$ by linear combination of lower order derivatives $v^{(l)}, l=1, 2, \cdots, j-1$, $\mbf J_d$ cannot be expressed with $\mbf J_{d}$ and $\mbf J_{\mbf x}$ by linear combination.
Therefore, by Lemma~\ref{lem:equiv}, line delay is weakly observable in general.
The trivial exception is when the observations $\mbf y(t)$ are constant.

\section{Supplementary Material}
\label{app:video}
A supplementary video showing simulation and real data tests is at \url{https://youtu.be/DeylCUuS-zU}.

\section*{Acknowledgment}
Slogging through 2+ grant proposals every month, J. Huai feels that he owes the Editor a decent bribe who granted the paper five extensions.
Also, J. Huai is thankful to Agostino Martinelli for the initial guidance on Lie derivatives, to practitioner Hongxing Sun and academician Yuanxi Yang for discussions about EKF, to Binliang Wang for help in processing data with the Kalibr toolbox.
Moreover, we appreciate the inspiring comments from anonymous reviewers 
who exhorted us to novelty and contribution.
\ifCLASSOPTIONcaptionsoff
  \newpage
\fi

\bibliographystyle{IEEEtran}
\bibliography{IEEEabrv,bib/zotero}

\end{document}

%% file: custom_commands/latin_abbreviations.tex
\usepackage{xspace}

\makeatletter
\DeclareRobustCommand\onedot{\futurelet\@let@token\@onedot}
\def\@onedot{\ifx\@let@token.\else.\null\fi\xspace}

\def\eg{\emph{e.g}\onedot} 
\def\ie{\emph{i.e}\onedot}

\def\etal{\emph{et al}\onedot}
\makeatother

%% file: tables/compared-methods.tex
\begin{table}[!htb]
	\centering
	\caption{Features of the compared odometry methods.}
	\label{tab:features-odometry-methods}
	\begin{tabular}{@{}l|cccc|cc|c@{}}
		\toprule
		\multicolumn{1}{c|}{Method} &
		\multicolumn{4}{c|}{\begin{tabular}[c]{@{}c@{}}Self-calibrated\\ camera params\end{tabular}} &
		\multicolumn{2}{c|}{\begin{tabular}[c]{@{}c@{}}Self-calibrated\\ IMU params\end{tabular}} &
		\multirow{2}{*}{\begin{tabular}[c]{@{}c@{}}Rolling \\ shutter\\ camera\end{tabular}} \\ \cmidrule(r){1-7}
		&
		$\mathbf{T}_{BC_k}$ &
		intr. &
		$t_d$ &
		$t_r$ &
		\begin{tabular}[c]{@{}c@{}}scale and\\ misalign.\end{tabular} &
		biases &
		\\ \midrule
		KSWF &
		$\checkmark$ &
		$\checkmark$ &
		$\checkmark$ &
		$\checkmark$ &
		$\checkmark$ &
		$\checkmark$ &
		$\checkmark$ \\ \midrule
		SL-KSWF &
		$\checkmark$ &
		$\checkmark$ &
		$\checkmark$ &
		$\checkmark$ &
		$\checkmark$ &
		$\checkmark$ &
		$\checkmark$ \\ \midrule
		\begin{tabular}[c]{@{}l@{}}KSWF cal.\\ camera\end{tabular} &
		$\checkmark$ &
		$\checkmark$ &
		$\checkmark$ &
		$\checkmark$ &
		&
		$\checkmark$ &
		$\checkmark$ \\ \midrule
		\begin{tabular}[c]{@{}l@{}}KSWF\\ cal. IMU\end{tabular} &
		$\checkmark$ &
		&
		&
		&
		$\checkmark$ &
		$\checkmark$ &
		$\checkmark$ \\ \midrule
		\begin{tabular}[c]{@{}l@{}}KSWF\\ min cal.\end{tabular} &
		$\checkmark$ &
		&
		&
		&
		&
		$\checkmark$ &
		$\checkmark$ \\ \midrule
		OpenVINS &
		$\checkmark$ &
		$\checkmark$ &
		$\checkmark$ &
		&
		&
		$\checkmark$ &
		\\ \midrule
		OKVIS &
		$\checkmark$ &
		&
		&
		&
		&
		$\checkmark$ &
		\\ \midrule
		VINS-Mono &
		$\checkmark$ &
		&
		$\checkmark$ &
		&
		&
		$\checkmark$ &
		$\checkmark$ \\ \bottomrule
	\end{tabular}
\end{table}

%% file: tables/init-value-std.tex
\begin{table}
\centering
\caption{Mean values and standard deviations $\sigma$ at each dimension of IMU and camera parameters used to simulate data and initialize estimators. Metric units are used unless noted.}
\begin{tabular}{cccccc}
\toprule
 &
$\mathbf{b}_g (^{\circ}/s)$ &
\begin{tabular}[c]{@{}c@{}}$\mathbf{b}_a$\end{tabular} &
$\mathbf{M}_g$ &
$\mathbf{T}_s$ &
$\mathbf{M}_a$ \\ \midrule
Mean &
$\mathbf{0}$ &
$\mathbf{0}$ &
$\mathbf{I}_3$ &
$\mathbf{0}_{3\times3}$ &
$\mathbf{I}_3$ \\ \midrule
\begin{tabular}[c]{@{}c@{}}$\sigma$\end{tabular} &
0.57 &
0.02 &
0.005 &
0.005 &
0.005 \\ \bottomrule\toprule
 &
$\mathbf{R}_{BC}$ &
\begin{tabular}[c]{@{}c@{}}$\mathbf{t}_{BC}$\\ ($cm$) \end{tabular} &
\begin{tabular}[c]{@{}c@{}}$f_x, f_y$\\ $c_x, c_y$ (px) \end{tabular} &
\begin{tabular}[c]{@{}c@{}}$k_1, k_2$\\ $p_1, p_2$\end{tabular} &
\begin{tabular}[c]{@{}c@{}}$t_d, t_r$\\ $(ms)$\end{tabular} \\ \midrule
Mean &
\begingroup
\setlength\arraycolsep{0.8pt}
\renewcommand*{\arraystretch}{0.4}
$\begin{bmatrix}0 & -1 & 0 \\ 0 & 0 & -1 \\ 1 & 0 & 0\end{bmatrix}$
\endgroup &
$\mathbf{0}$ &
\begin{tabular}[c]{@{}c@{}}350, 360 \\ 378, 238\end{tabular} &
\begin{tabular}[c]{@{}c@{}}0, 0 \\ 0, 0\end{tabular} &
20, 20 \\ \midrule
\begin{tabular}[c]{@{}c@{}}$\sigma$\end{tabular} &
0.57 ($^{\circ}/s$) &
2.0 &
\begin{tabular}[c]{@{}c@{}}2.0, 2.0\\ 2.0, 2.0\end{tabular} &
\begin{tabular}[c]{@{}c@{}}0.01, 0.01\\ 0.01, 0.01\end{tabular} &
5, 5 \\ \bottomrule
\end{tabular}
\label{tab:init_value_std}
\end{table}

%% file: tables/sim-rmse.tex
\begin{table}[!htb]
	\centering
	\caption{RMSE at the end of 5-minute wavy motion for OKVIS, SL-KSWF, and KSWF with varying levels of self-calibration, 
		computed over successful runs of 100 attempts which were deemed successful if the position error at the end is \textless 100 m.}
	\label{tab:sim-rmse}
	\begin{tabular}{@{}l|ccc|ccc@{}}
		\toprule
		\multicolumn{1}{c|}{}                                               & \multicolumn{3}{c|}{Wavy circle} & \multicolumn{3}{c}{TUM VI corridor3} \\ \cmidrule(l){2-7} 
		\multicolumn{1}{c|}{Method} &
		\multicolumn{2}{c|}{RMSE} &
		\multirow{2}{*}{\begin{tabular}[c]{@{}c@{}}Succ.\\ runs\end{tabular}} &
		\multicolumn{2}{c|}{RMSE} &
		\multirow{2}{*}{\begin{tabular}[c]{@{}c@{}}Succ.\\ runs\end{tabular}} \\ \cmidrule(lr){2-3} \cmidrule(lr){5-6}
		&
		\begin{tabular}[c]{@{}c@{}}$\mbf{p}_{WB}$\\ (m)\end{tabular} &
		\multicolumn{1}{c|}{\begin{tabular}[c]{@{}c@{}}$\mbf{R}_{WB}$\\ ($^\circ$)\end{tabular}} &
		&
		\begin{tabular}[c]{@{}c@{}}$\mbf{p}_{WB}$\\ (m)\end{tabular} &
		\multicolumn{1}{c|}{\begin{tabular}[c]{@{}c@{}}$\mbf{R}_{WB}$\\ ($^\circ$)\end{tabular}} &
		\\ \midrule
		\begin{tabular}[c]{@{}l@{}}OKVIS \\calibrated\end{tabular}                                                            & 3.92       & 36.24     & 100     & 6.6         & 21.13       & 100      \\ \midrule
		\begin{tabular}[c]{@{}l@{}}OKVIS \\uncalibrated\end{tabular} & 9.67       & 37.74     & 100     & 13.98       & 28.61       & 100      \\ \midrule
		\begin{tabular}[c]{@{}l@{}}KSWF \\ min cal.\end{tabular}           & 16.62      & 39.37     & 91      & 11.03       & 23.03       & 87       \\ \midrule
		\begin{tabular}[c]{@{}l@{}}KSWF\\ cal. camera\end{tabular}         & 7.82       & 39.17     & 93      & 4.12        & 10          & 98       \\ \midrule
		\begin{tabular}[c]{@{}l@{}}KSWF\\ cal. IMU\end{tabular}            & 14.37      & 39.76     & 100     & 17.75       & 28.04       & 88       \\ \midrule
		\begin{tabular}[c]{@{}l@{}}KSWF\\ full cal.\end{tabular}           & 0.64       & 10.77     & 100     & 1.55        & 4.48        & 100      \\ \midrule
		\begin{tabular}[c]{@{}l@{}}SL-KSWF\\ full cal.\end{tabular}        & 3.73       & 24.81     & 16      & 5.98        & 3.57        & 35       \\ \bottomrule
	\end{tabular}
\end{table}

%% file: tables/tumvi-room-rpe.tex
\begin{table}[!htb]
	\centering
	\caption{Relative pose errors (RTE: relative translation error, RRE: relative rotation error) of the compared methods on the six raw TUM VI room sequences, each with five runs. Also listed are the number of runs (out of 30) with the absolute translation error (ATE) \textgreater 15 m and the number of runs with the absolute rotation error (ARE) \textgreater 15$^\circ$. All methods calibrate IMU biases and camera extrinsic parameters during motion estimation. Note that large RTE values say 100 indicate that there are trajectories of large drift.}
	\label{tab:tumvi-room-rpe}
	\begin{tabular}{@{}l|cccc@{}}
		\toprule
		\multicolumn{1}{c|}{Method} &
		\begin{tabular}[c]{@{}c@{}}RTE\\ (\%)\end{tabular} &
		\begin{tabular}[c]{@{}c@{}}RRE\\ ($^\circ/m$)\end{tabular} &
		\begin{tabular}[c]{@{}c@{}}\#Runs with\\ ATE\textgreater{}15 m\end{tabular} &
		\begin{tabular}[c]{@{}c@{}}\#Runs with \\ ARE\textgreater{}15$^\circ$\end{tabular} \\ \midrule
		KSWF                                                                                          & 0.34  & 0.086 & 0  & 1  \\ \midrule
		SL-KSWF                                                                                       & 924.4 & 0.111 & 1  & 2  \\ \midrule
		\begin{tabular}[c]{@{}l@{}}KSWF cal.\\camera\end{tabular}                & 6969  & 0.499 & 15 & 15 \\ \midrule
		\begin{tabular}[c]{@{}l@{}}KSWF cal.\\IMU\end{tabular}                   & 1421  & 0.200 & 2  & 2  \\ \midrule
		\begin{tabular}[c]{@{}l@{}}KSWF min\\ cal.\end{tabular} & 14771 & 0.502 & 18 & 18 \\ \midrule
		OpenVINS                                                                                      & 20976 & 0.426 & 20 & 20 \\ \midrule
		OKVIS                                                                                         & 9783  & 0.443 & 22 & 22 \\ \midrule
		VINS-mono                                                                                     & 183.5 & 0.698 & 25 & 29 \\ \bottomrule
	\end{tabular}
\end{table}

%% file: tables/euroc-timing.tex
\begin{table}[!htb]
	\centering
	\caption{Time costs of four components of KSWF,  KSWF with minimal calibration, and SL-KSWF with minimal calibration in processing EuRoC sequences in milliseconds. 
	The last column shows the average time to process a frame bundle which is the sum of \textbf{bold} values in each row.}
	\label{tab:euroc-timing}
	\begin{tabular}{@{}lccccc@{}}
		\toprule
		\multicolumn{1}{c}{Method} &
		\begin{tabular}[c]{@{}c@{}}Feature\\ extraction\end{tabular} &
		\begin{tabular}[c]{@{}c@{}}Feature\\ matching\end{tabular} &
		Filtering &
		Marg. &
		\\ \midrule
		mo. KSWF &
		4.0$\pm$1.1 &
		\textbf{2.3}$\pm$1.0 &
		\textbf{4.7}$\pm$2.2 &
		\textbf{0.8}$\pm$1.0 &
		7.8 \\ \midrule
		\begin{tabular}[c]{@{}l@{}}mo. KSWF\\ min cal.\end{tabular} &
		3.7$\pm$0.8 &
		\textbf{2.1}$\pm$0.9 &
		\textbf{3.8}$\pm$1.9 &
		\textbf{0.5}$\pm$0.6 &
		6.5 \\ \midrule
		\begin{tabular}[c]{@{}l@{}}mo. SL-\\ KSWF\\ min cal.\end{tabular} &
		3.5$\pm$0.6 &
		\textbf{2.0}$\pm$0.9 &
		\textbf{1.5}$\pm$0.6 &
		\textbf{0.4}$\pm$0.6 &
		4.0 \\ \midrule
		KSWF &
		11.2$\pm$2.0 &
		\textbf{5.7}$\pm$2.4 &
		\textbf{13.7}$\pm$5.4 &
		\textbf{2.2}$\pm$2.8 &
		21.7 \\ \midrule
		\begin{tabular}[c]{@{}l@{}}KSWF\\ min cal.\end{tabular} &
		9.9$\pm$1.3 &
		\textbf{5.1}$\pm$1.9 &
		\textbf{9.7}$\pm$3.0 &
		\textbf{1.2}$\pm$1.3 &
		16.0 \\ \midrule
		\begin{tabular}[c]{@{}l@{}}SL-KSWF\\ min cal.\end{tabular} &
		9.9$\pm$1.3 &
		\textbf{5.0}$\pm$1.8 &
		\textbf{4.2}$\pm$1.7 &
		\textbf{1.1}$\pm$1.4 &
		10.3 \\ \bottomrule
	\end{tabular}
\end{table}